%% file: acl_latex.tex
\pdfoutput=1

\documentclass[11pt]{article}

\usepackage[preprint]{acl}

\usepackage{fancyhdr}
\usepackage{times}
\usepackage{latexsym}
\usepackage{amsmath}
\usepackage{array} 
\usepackage{booktabs} 
\usepackage{graphicx} 
\usepackage{makecell} 
\usepackage{colortbl}
\usepackage{tabularx} 
\usepackage{caption} 
\usepackage{float}
\usepackage{tcolorbox}
\tcbuselibrary{breakable}

\usepackage[T1]{fontenc}

\usepackage[utf8]{inputenc}

\usepackage{microtype}

\usepackage{inconsolata}

\usepackage{graphicx}

\title{Explicit Learning and the LLM in Machine Translation}

 \author{Malik Marmonier \quad Rachel Bawden \quad Benoît Sagot \\
         Inria, Paris, France\\ \{\texttt{firstname.lastname\}@inria.fr}}

\begin{document}
\maketitle
\begin{abstract}
This study explores an LLM's ability to learn new languages using explanations found in a grammar book—a process we term ``explicit learning.'' To rigorously assess this ability, we design controlled translation experiments between English and constructed languages generated—through specific cryptographic means—from Latin or French. Contrary to previous studies, our results demonstrate that LLMs do possess a measurable capacity for explicit learning. This ability, however, diminishes as the complexity of the linguistic phenomena to be learned increases. Supervised fine-tuning on \textit{ad hoc} chains of thought significantly enhances LLM performance but struggles to generalize to typologically novel or more complex linguistic features. These findings point to the need for more diverse training sets and alternative fine-tuning strategies to further improve explicit learning by LLMs, benefiting low-resource languages typically described in grammar books but lacking extensive corpora.
\end{abstract}

\section{Introduction}
The question as to whether large language models (LLMs) can learn to translate a ``new'' language using grammar books has recently been of interest to a growing number of researchers \citep[inter alia]{tanzer_benchmark_2023, hus_back_2024,aycock_can_2024}, but has not yet received—to the best of our knowledge—a definitive answer. Two main challenges explain this delay:

\begin{figure}[t!]
    \centering
    \includegraphics[width=1\columnwidth]{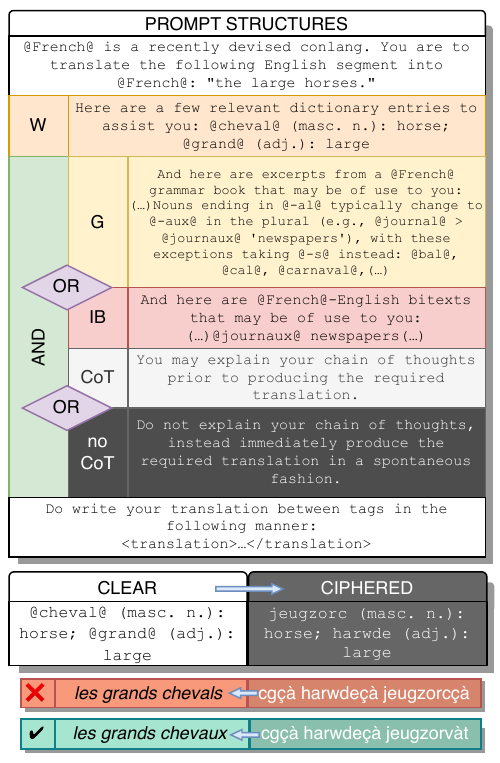} 
    \caption{We evaluate an LLM's ability to learn a ``new'' language using a grammar book--a process we term ``explicit learning.'' We obtain such new languages through special encipherment of French or Latin segments (enclosed in `@' characters for selective encryption) in our English prompts. We study the impact of manually adapted grammar book excerpts on translation performance by varying the information content of our prompts: (\textit{W}) dictionary entries offered by default; (\textit{G}) grammar book excerpts, alternating with (\textit{IB}) incidental bitexts, i.e., the parallel corpus formed by the translated examples occurring in the grammar book excerpts. We further study the impact of chain-of-thought (CoT) reasoning on LLM performance. Decipherment of LLM output facilitates qualitative analysis, and we rely on match accuracy for quantitative evaluation.}
    \label{fig:synoptic}
\end{figure}

\textbf{Opaque prior knowledge of LLMs.} Trained on web-scale datasets rendered opaque by their very size, LLMs are likely to have implicitly learned—though to a degree virtually unknown to the researcher—a wide array of languages, hindering credit assignment: when grammatical explanations are added to the prompt of an LLM tasked with performing a translation task, should correct translations be attributed to these explanations or to the model's prior, latent knowledge of the ``new'' language, whether direct or indirect?\footnote{We note that even extremely low-resource languages absent from the Web are likely to share transparent surface features with high-resource counterparts, due to either contact with vehicular languages or shared phylogeny.} 

\textbf{Incidental bitexts in grammar books.} Grammar books invariably include translated examples that are likely to be processed as few-shot bitexts by LLMs. When prompted with the text of a grammar book, are such models able to learn aspects of a new language by relying on the metalinguistic explanations provided—an operation we choose to call ``explicit learning,'' borrowing a term from second language acquisition studies \citep{ellis_implicit_2009, rebuschat_implicit_2015}? Or do they incidentally learn said aspects from a parallel corpus (formed by the translated examples)—the default approach to neural model training and therefore a trivial task?

Our first contribution addresses the former challenge. We devise cryptographic techniques, robust to the innate cryptanalytic abilities of current LLMs \citep{cao_unnatural_2023,yuan_gpt-4_2024}, which also preserve the regularity of grammatical paradigms. We use these techniques to generate constructed languages (henceforth `conlangs') through encryption of Latin or French segments of text, thus creating languages truly new to LLMs for subsequent experiments.

The second of our contributions tackles the challenge of incidental bitexts within grammar books. We manually craft authentic-style grammar books and dictionaries for our conlangs, preserving their typical translated examples for ecological validity. By carefully keeping track of these incidental bitexts, we design a rigorous, contrastive test set to distinguish learning from such parallel data versus from grammatical explanations. Furthermore, we develop extensive chain-of-thought (CoT) training and validation sets to investigate the possibility of enhancing explicit learning through fine-tuning.

Our third and main contribution lies in our results. To the best of our knowledge, we provide the first clear evidence that LLMs possess a measurable capacity for explicit learning, though this ability diminishes as the complexity of the linguistic phenomena at hand increases. We show that supervised fine-tuning on chains of thought (CoT) significantly enhances this capacity, but that such a fine-tuning regimen fails to generalize to typologically novel or to more complex linguistic phenomena. These findings point to the need to incorporate a broader range of languages into training datasets such as ours and explore alternative fine-tuning strategies to foster universal explicit learning in LLMs.

We publicly release our code and datasets under MIT and CC-BY-SA 4.0 licenses, respectively.\footnote{\url{https://github.com/mmarmonier/SWELLS}}

\section{Related Work}
BIG-bench \citep{srivastava_beyond_2023} comprises a little-known ciphered conlang translation task.\footnote{Conlang Translation Problems: \url{https://github.com/google/BIG-bench/tree/main/bigbench/benchmark_tasks/conlang_translation}. Accessed: 2024-10-10.} It consists of Rosetta Stone problems, either directly borrowed from puzzle sets of the International Linguistic Olympiads\footnote{International Linguistic Olympiads: \url{https://ioling.org/index.html}. Accessed: 2024-10-10.} and the North American Computational Linguistics Olympiad,\footnote{North American Computational Linguistics Olympiad: \url{https://naclo.org/}. Accessed: 2024-10-10.} or inspired by them. Encipherment of these test sets was linguistically informed, rather than algorithmic, and was performed manually by the authors (personal correspondence). \citet{bozhanov_rosetta_2013} describe the \textit{modus operandi} for crafting such Rosetta Stone problems, and prior—cipher-less—machine translation (MT) experiments involving such datasets were conducted by \citet{sahin_puzzling_2020}. Both the latter study and the BIG-bench task exclusively rely on implicit learning, evaluating or training models on bitexts illustrative of the linguistic phenomena of interest, but involving no metalinguistic explanations of them. The challenging character of these tasks, as compared with run-of-the-mill MT which objectively relies on the same training modality, results from the extreme data scarcity and lack of redundancy they enforce, while being theoretically self-contained and solvable. \citet{wei_frequency_2021} offer a compelling analysis of such frequency effects on the implicit assimilation of abstract grammar rules by a language model.

While not directly concerned with explicit learning, \citet{coleman_llm-assisted_2024} and \citet{guo_teaching_2024} offer interesting illustrations of the related use of an LLM's metalinguistic abilities in a cascaded MT pipeline, and \citet{thrush_i_2024} present a novel LLM benchmark based on the processing of a specific subset of metalinguistic discourses: those of the self-referential kind.

A growing number of studies could be construed as dealing with explicit learning in low-resource MT. \citet{tanzer_benchmark_2023} introduced the seminal MTOB (Machine Translation from One Book) LLM benchmark, which involves translating 100 English--Kalamang \citep{visser_grammar_2022} bitexts using word list entries, grammar book excerpts, bitexts or combinations thereof in a RAG setting,\footnote{Retrieval-Augmented Generation; this aspect of the benchmark appears to have been abandoned in its most recent implementations, which provide the full text of the grammar book and dictionary to long-context LLMs \citep{gemini_team_gemini_2024}.} with \citet{hus_back_2024} extending this benchmark to 15 additional languages. Several studies have also touched on this subject tangentially: \citet{zhang_teaching_2024} conducted a small side experiment on the explicit learning of constituent order rules in Zhuang--Chinese MT, finding that the task poses ``significant challenges'' to LLMs; \citet{zhang_hire_2024} used GPT-4 to summarize grammar books in order to meet context-window constraints, and tentatively suggested that these documents improve word order correctness in LLM-based MT; and \citet{sanchez_linguini_2024} reported results from a small-scale side experiment providing grammar books to an LLM tasked with solving Rosetta Stone problems, yielding mixed results.

Work by \citet{aycock_can_2024} is the most relevant to our own. These authors sought to ``disentangl[e] the parallel/non-parallel signal [in the MTOB grammar book],'' and performed numerous experiments to that effect, most notably ``ablating'' all occurrences of English--Kalamang bitexts in the grammar book, thus compiling a small parallel corpus with the extracted data, and crafting an example-free version of that document, although this could possibly confuse LLMs. The authors' conclusion that ``[grammatical] explanations [have] no significant advantage over parallel data'' left open interesting questions regarding the extent to which LLMs can process metalinguistic explanations in translation and whether this ability can be enhanced or is effectively set in stone.

\section{Materials and Techniques}
This section describes the methods and resources we devised to assess explicit learning in LLMs—that is these models' ability to learn aspects of a new language on the basis of grammatical explanations rather than through implicit exposure to the linguistic phenomena—addressing the challenge of their opaque prior knowledge by means of cryptographic conlangs, and that of the incidental bitexts in grammar books through the manual crafting of an \textit{ad hoc} test set.

Our core approach evaluates the LLM's ability to translate segments of increasing complexity--though never longer than a single sentence (see Table~\ref{tab:linguistic_phenomena})--between English and artificial languages, in both directions (eng$\Leftrightarrow$art\footnote{We use the ISO 639-2 code `art' to refer to our conlangs.}), using dictionary entries embedded into prompts and contrastively augmented with either grammar book excerpts or the incidental bitexts extracted from those excerpts.
Section 3.1 presents these linguistic resources, including the grammar books, dictionaries, and structured datasets. Section 3.2 then explains the cryptographic cipher schemes used to encrypt all conlang segments in these datasets, ensuring their novelty to the LLM, as well as enabling renewal of these resources through re-encryption for data augmentation or test-leakage-control purposes.

\subsection{Resources}
The following paragraphs describe the grammar books, dictionaries, and datasets we developed, all crafted as cleartext\footnote{In cryptography, the term cleartext (or plaintext) refers to text in its normal, unencrypted, human-readable form, before it has been converted into a secret code (ciphertext).} templates where `@' characters delimit text segments intended for selective encipherment, thus forming the basis for specific conlang resources (see also Figures~\ref{fig:synoptic} and \ref{fig:cipherEx}).

\textbf{Grammar books.} We adapted content from English Wikipedia pages on French and Latin grammar (see Appendix~\ref{sec:gs_appendix}) to create three distinct grammar book templates: one each for French, Latin, and ``reversed-French.'' While the French/Latin templates received minimal changes, the reversed-French grammar required substantial rewriting (e.g., suffixes becoming prefixes, antecedents becoming head nouns, word order being reversed) due to the specific cipher ultimately used to encrypt this template (Section 3.2).

\textbf{Dictionaries.} We created reversible French-English (158 nouns, 76 adjectives, 6 verbs) and Latin-English (19 nouns, 9 adjectives, 10 verbs) dictionary templates. Lemmas were selected based on linguistic phenomena described in our grammar books that we deemed interesting and assigned to three categories: (A)~Lemmas appearing in the grammar book within bitexts illustrative of a phenomenon of interest, allowing the LLM to rely on direct copying for translation; (B)~Lemmas appearing only in metalinguistic explanations within the grammar book, pointing to rules governing the targeted phenomenon without direct illustration of the expected form; and (C)~The majority of lemmas absent from the grammar book entirely, requiring the LLM to infer the applicable rule for generating or analyzing the targeted phenomenon in translation. Detailed counts per category are in Appendix~\ref{app:dictionary_counts}.

\textbf{Test set.} Using entries from our dictionaries and excerpts from our grammar books embedded into prompts, we created a test set evaluating 14 linguistic phenomena of interest identified during dictionary construction (Table~\ref{tab:linguistic_phenomena}), with 100 different base instances per phenomenon (1,400 in total).\footnote{Of these 1,400 base instances, 400 pertain to the 4 Latin-derived phenomena listed in Table~\ref{tab:linguistic_phenomena} (100 each). The remaining 1,000 instances cover the 10 French-derived phenomena; for each of these, the 100 instances are evenly split between French and reversed-French (50 for each).} As previously outlined, each test instance was further made to vary according to prompt content (\textit{W} for dictionary entries only, \textit{W+G} for dictionary entries and relevant grammar book excerpts, \textit{W+IB} for dictionary entries and incidental bitexts taken from the grammar book), translation direction (eng→art and art→eng) and CoT use (Appendix~\ref{sec:cotNoCot}), resulting in 24 contrastive versions of the test set (33,600 instances in total).

\textbf{Training and validation sets.} We also crafted $\sim$358,000 CoT demonstration templates (Appendix~\ref{sec:pr_appendix}) for French-derived phenomena (Latin phenomena were excluded to test generalization), plus equivalent non-CoT instances. To prevent test set leakage, we produced hundreds of thousands of paraphrased grammar book excerpts which we used in training prompts in lieu of their test set equivalents (Appendix~\ref{sec:gtempl_appendix}). 

Encipherment of the resulting datasets can significantly expand the diversity and quantity of these resources, as the key space of the cipher schemes we will now describe ranges into the hundreds of novemdecillions.

\begin{table*}[h]
\renewcommand{\arraystretch}{1.3}
\centering
\tiny
\rowcolors{2}{gray!15}{white}
\begin{tabular}{>{\centering\arraybackslash}m{0.2cm}|p{7cm}|p{7cm}}
\toprule
\rowcolor{gray!30}
\textbf{ID} & \textbf{Linguistic Phenomena of Interest and Combinations Thereof} & \textbf{Bitext Examples} \\
\midrule
1. & Plural of French common nouns & @les coraux@ $\leftrightarrow$ the corals \\
2m. & Plural and position of French masculine common nouns and adjectives & @les nouveaux jeux@ $\leftrightarrow$ the new games \\
2f. & Plural and position of French feminine common nouns and adjectives & @les jolies maisons@ $\leftrightarrow$ the pretty houses \\
3. & French verbal morphology & @nous choisirions@ $\leftrightarrow$ we would choose \\
4. & French verbal morphology; plural and position of French common nouns & @elle admirait les coraux@ $\leftrightarrow$ she was admiring the corals \\
5m. & French verbal morphology; plural and position of French masculine common nouns and adjectives & @j'ai choisi les voyous généreux@ $\leftrightarrow$ I chose the generous thugs \\
5f. & French verbal morphology; plural and position of French feminine common nouns and adjectives & @il enfouira les fourmis rouges@ $\leftrightarrow$ he will bury the red ants \\
6. & Plural of French common nouns and complex past participle agreement rule & @les croix que nous avons choisies@ $\leftrightarrow$ the crosses that we chose \\
7m. & Plural of French masculine common nouns and adjectives, and complex past participle agreement rule & @les pneus gris que nous avons choisis@ $\leftrightarrow$ the gray tires that we chose \\
7f. & Plural of French feminine common nouns and adjectives, and complex past participle agreement rule & @les croix cassées que nous avons enfouies@ $\leftrightarrow$ the broken crosses that we buried \\
8a. & Latin verbal morphology, nominative and accusative singular and plural of the Latin noun & @equus piratam timet@ $\leftrightarrow$ the horse fears the pirate \\
8g. & Latin verbal morphology, nominative and genitive singular and plural of the Latin noun & @nauta miserebitur piratarum@ $\leftrightarrow$ the sailor will pity the pirates \\
9a. & Latin verbal morphology, nominative and accusative singular and plural of the Latin noun and adjective & @mali piratae sacram effigiem timebunt@ $\leftrightarrow$ the bad pirates will fear the sacred statue \\
9g. & Latin verbal morphology, nominative and genitive singular and plural of the Latin noun and adjective & @novi agricolae obliviscentur parvi bovis@ $\leftrightarrow$ the new farmers will forget the small ox \\
\bottomrule
\end{tabular}
\caption{Linguistic phenomena targeted in our datasets, numbered in increasing order of complexity according to our own \textit{a priori} judgment, with example bitexts.}
\label{tab:linguistic_phenomena}
\end{table*}

\subsection{Cipher schemes}
Following \citet{couturat_histoire_1903}, conlangs are generally classified into two types: \textit{a priori} conlangs, which bear no relation to existing natural languages, such as John Wilkins's Real Character \citep{wilkins_essay_1668}; and \textit{a posteriori} conlangs, such as Esperanto, closely modeled on existing languages from the point of view of their grammar and lexical inventory. In order to maximize the ecological validity of our experiments, we chose to generate radically \textit{a posteriori} conlangs through the masking of French or Latin segments by cryptographic means.

Identifying a suitable ciphering scheme proved nontrivial. Modern advanced ciphers possess the undesirable—for our purpose—property of diffusion \citep{shannon_mathematical_1945}, destructive of the morphological paradigms and semantic regularities in ciphered texts, leaving only older, simpler ciphers as viable options. LLMs, however, exhibit a remarkable ability to see through these simpler schemes. Simple substitution ciphers, such as the Caesar cipher, which replace each cleartext character with a single stand-in symbol, have been shown to be transparent to these advanced models \citep{yuan_gpt-4_2024}, and experiments by \citet{cao_unnatural_2023} incidentally indicate that the same likely applies to transposition ciphers, which preserve cleartext symbols but shuffle their positions.

We devised two cryptographic schemes that meet our task's constraints, preserving grammatical and lexical paradigms, while being opaque to the LLM tested. 

\textbf{First scheme.} Our first scheme fundamentally relies on a substitution of the Polybius-square type, which in our implementation uses 2D-grid coordinates to select vowel-consonant bigrams in a cipher alphabet as substitutes for cleartext characters. Since Polybius-square encipherment doubles the length of the original text strings, dramatically increasing token counts, we chose to apply it only partially, i.e., on a subset of the cleartext alphabet (in proportions varying between 20\% and 75\% for each of the thousands of conlangs we generated), with the remaining characters enciphered through simple substitutions of the Caesar-cipher type. The resulting composite cipher can be formalized in the following way: let the character list of the natural language chosen to be the matrix of the \textit{a posteriori} conlang be defined as:
\[
\Sigma = \{\text{a}, \text{b}, \text{c}, \ldots, \text{z}\}.
\]
Define two sets \(C\) and \(V\) of consonantal and vocalic characters, respectively, to be used in the enciphered conlangs:
\[
C = \{c_1, c_2, \ldots, c_m\}, \quad V = \{v_1, v_2, \ldots, v_n\},
\]
with subsets \(C_k\) and \(V_l\) randomly sampled as:
\[
C_k \sim \binom{C}{k}, \quad V_l \sim \binom{V}{l}.
\]
The set of code points \(G\) is then defined as:
\begin{align*}
G = &\; \big\{(i, j) \mid i \in C_k, \, j \in V_l\big\} \\
    &\cup \big\{i \mid i \in C \setminus C_k\big\} \\
    &\cup \big\{j \mid j \in V \setminus V_l\big\},
\end{align*}
with $C \setminus C_k$ representing the set of consonants not selected to form digrams at the previous step, and $V \setminus V_l$ the equivalent set of vocalic characters. Finally, let substitutions be randomly assigned by the application \(\phi: \Sigma \to G\):
\begin{align*}
    \phi(\text{a}) &= (c_{10}, v_1), \\
    \phi(\text{b}) &= (c_{16}), \\
    &\;\vdots \\
    \phi(\text{z}) &= (c_5, v_1),
\end{align*}
with monograms and digrams in set \(G\) selected as code points with differing probabilities, themselves randomly defined:
\[
\phi(x) = 
\begin{cases} 
  \text{digram}, & \text{with probability } p, \\
  \text{monogram}, & \text{with probability } 1 - p,
\end{cases}
\]
where \(p \sim \text{Uniform}\big(\{2/5, 1/2, 2/3, 3/4\}\big)\).

\textbf{Second scheme.} The second cipher we devised simply consists in the composite first scheme just described, combined with an additional transposition that reverses the order of the cleartext symbols:
\begin{align*}
s'[i] &= s[t - i + 1] \quad \text{for } 1 \leq i \leq t,
\end{align*}
with \(s\) a sentence, and \(t\) the number of characters in \(s\). 

\begin{figure}[h]
    \centering
    \includegraphics[width=0.9\columnwidth]{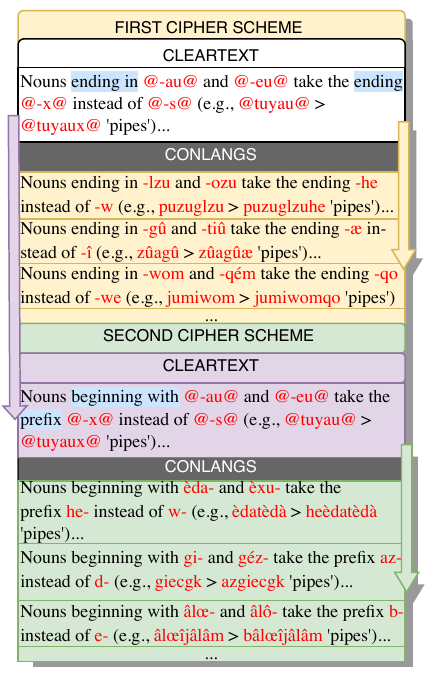} 
    \caption{The two cipher schemes used to generate conlangs. The First Scheme relies on character substitutions, while the Second Scheme adds a transposition (character order reversal), requiring prior manual adaptation of grammar book excerpts (purple arrow).}
    \label{fig:cipherEx}
\end{figure}

Figure~\ref{fig:cipherEx} gives an illustration of these schemes, and Appendix~\ref{sec:encEx} provides additional examples of their concrete aspect.

These ciphering schemes were applied to the cleartext templates (detailed in Section 3.1) to generate our conlang resources: the first scheme (substitutions only) was used for French-derived and Latin-derived resources, while the second scheme (substitutions and transposition) was applied to the reversed-French resource templates. While each template instance in the training and validation sets was enciphered with a new, randomly generated key (i.e., a different state of $\phi$) in order to maximize data diversity, for the test set, each of the 1,400 base instances utilized its own distinct encipherment key (creating 1,400 unique conlangs in their surface forms), but this specific key was kept consistent across all 24 experimental variations of that instance to ensure fair comparison across test conditions.

\section{Experimental Setup}
We created a small curriculum of seven courses, corresponding to the ``French'' partitions detailed in Table~\ref{tab:linguistic_phenomena} (ignoring noun gender distinctions). \texttt{GPT-4o-mini-2024-07-18} was fine-tuned on this curriculum for one epoch per ``course,'' using a single instance per batch and a learning rate multiplier of 1.8.\footnote{These last two hyperparameters were set autonomously by the API.} The resulting checkpoint was further fine-tuned for four epochs on a final training set comprising varied instances (see Table~\ref{tab:sft_curriculum} for instance and token counts). Throughout these training sets, the art$\rightarrow$eng direction was underrepresented at a 1:10 ratio relative to eng$\rightarrow$art, as it was deemed less challenging \textit{a priori}; the same ratio applied to examples of spontaneous translation as opposed to CoT demonstrations; French-derived and reversed-French-dervied conlangs were equally represented. Latin-derived conlangs were excluded from the training data to assess the model's ability to generalize explicit learning to typologically novel linguistic phenomena.

\begin{table}[h]
    \centering
    \small
    \rowcolors{2}{gray!15}{white}
    \begin{tabularx}{0.8\columnwidth}{ >{\centering\arraybackslash}X >{\centering\arraybackslash}X >{\centering\arraybackslash}X }
        \toprule
        \rowcolor{gray!30}
        \textbf{SFT Partitions} & \textbf{Instance Counts} & \textbf{Token Counts} \\
        \midrule
        Course 1 & 276 & 454,454 \\
        Course 2 & 1,292 & 3,856,069 \\
        Course 3 & 754 & 4,626,879 \\
        Course 4 & 370 & 2,612,112 \\
        Course 5 & 286 & 2,567,109 \\
        Course 6 & 326 & 2,777,347 \\
        Course 7 & 376 & 3,111,776 \\
        Final & 1,408 & 10,551,709 \\
        \bottomrule
    \end{tabularx}
    \caption{Instance and token counts in our training sets.}
    \label{tab:sft_curriculum}
\end{table}

Our baseline was \texttt{GPT-4o-mini}.\footnote{\url{https://openai.com/index/gpt-4o-mini-advancing-cost-efficient-intelligence/}, accessed: 2025-02-14} Test set inferences were conducted in December 2024 via OpenAI's batch API with a temperature of 0.05 (see Appendix~\ref{sec:gtemp_appendix}) and \texttt{max\_tokens} values ranging from 100 to 1300, depending on the configuration and instance type evaluated.

\section{Results}
We used accuracy as our primary metric, expressed as percentages. An exact match between the LLM’s lowercase candidate translation\footnote{All candidate translations were lower-cased and stripped of full stops.} and the reference translation was scored as 100\%; deviations were scored as 0\%. In an effort to soften the brittleness of accuracy scoring, we awarded 50\% to translations with incorrect adjective placement as their sole error in partitions involving French-derived adjectives.\footnote{A special reference, \texttt{ad\_order\_error\_reference}, was created during test set generation to facilitate this scoring. French adjectives follow specific placement rules based on the conceptual category to which they belong: some adjectives typically precede the noun (e.g.,~adjectives describing size, age, or beauty), while others typically follow it (e.g.,~adjectives specifying color, nationality, or shape). Incorrect adjective placement refers here specifically to the reversal of this established ordering, and not to random misplacement which we would still score 0\%.} For Latin-derived conlangs (partitions 8- and 9-), accuracy was computed using bag-of-words matching, accommodating Latin’s relatively free word order.

In the tables below, each cell gives the mean accuracy over 100 distinct test instances (each a unique bitext with a unique conlang encipherment). For French-derived phenomena (partitions 1-7f), results are a 50/50 composite of standard and ``reversed'' French instances (see Appendix~\ref{sec:trVsRev} for a breakdown of baseline scores). Critically, the 100 base instances for any partition remain consistent across the experimental conditions (W, W+IB, W+G, and CoT/no-CoT) presented in columns, allowing for direct comparison.

\subsection{ENG$\rightarrow$ART}
This translation direction is the most challenging for the LLM \textit{a priori}, as it requires generating a segment in the conlang—a language unknown, learned only in context within the provided prompt.

\subsubsection{Baseline}
The scores in Table~\ref{tab:baseline_eng_art}--supported by extensive qualitative analysis of LLM output--show that the base model can perform explicit learning in context. This ability, however, declines sharply as the linguistic phenomena involved increase in number and complexity. This observation is evident in the score difference between the 2m and 2f partitions: since French dictionary entries list adjectives in the masculine singular form, converting them to feminine singular prior to generating the expected plural results in a 90\% drop in accuracy compared to the masculine adjective partition, with grammar excerpts of equal length.

\input{table3}

Effect sizes based on Cohen's~\textit{d} \citep{cohen_statistical_1988} indicate that, although generally weak for all but the simplest test set partitions, explicit learning remains evident in aggregate. This is reflected in medium effect sizes of 0.46 and 0.35 for the CoT \textit{W+G} treatment compared to the \textit{W} and \textit{W+IB} treatments, respectively (see Appendix~\ref{sec:cohdEngArtB}). A small effect size (0.14) for CoT over `no CoT' in the \textit{W+G} configuration further suggests that ``reflection'' contributes positively to model performance.

\subsubsection{Fine-tuned checkpoint}

The accuracy scores in Table~\ref{tab:ft_scores_eng_art} show that fine-tuning dramatically enhances the LLM's explicit learning capability, with very large effect sizes \citep{sawilowsky_new_2009} of up to 1.84 and 1.75 for \textit{W+G} over \textit{W} and \textit{W+IB}, respectively (see Appendix~\ref{sec:cohdEngArtF}). These results also highlight the amplified role of CoTs in the performance of the fine-tuned checkpoint, with an effect size of 0.58 for CoT \textit{W+G} over its CoT-free counterpart. These improvements, however, did not extend to the Latin-derived conlangs excluded from the fine-tuning dataset, for which the fine-tuned checkpoint performs on par with the baseline.

\input{table4}

\subsection{ART$\rightarrow$ENG}
This translation direction is easier, \textit{a priori}, for the LLM, as the generated text is in a language well-known to it: English. One would expect dictionary entries (\textit{W}) and incidental bitexts (\textit{W+IB}) to provide sufficient information for the model to guess the expected translations to varying degrees.

\subsubsection{Baseline}
We were surprised to find that the \textit{W+G} treatment achieved significantly higher accuracy scores and effect sizes than the \textit{W+IB} treatment in this translation direction (see Table~\ref{tab:baseline_scores_art_eng} and Appendix~\ref{sec:cohdArtEngB}). This further confirms our conclusion that the LLM does indeed leverage metalinguistic discourses in the grammar book, although, as was already the case in the opposite translation direction, this capability is quickly overwhelmed as linguistic complexity increases.

\input{table5}

We note that in this translation direction, the benefits of CoT for the \textit{W+G} treatment appear negligible (effect size: 0.03).

\subsubsection{Fine-tuned checkpoint}

Despite the underrepresentation of this translation direction in the fine-tuning data, accuracy scores for linguistic phenomena seen during fine-tuning improved dramatically (see Table~\ref{tab:ft_scores_art_eng}). While fine-tuning also enhanced the model's ability to guess English translations from dictionary entries and incidental bitexts, the \textit{W+G} treatment remained surprisingly superior, notably for partitions involving complex verb forms (i.e., 3 and above).

\input{table6}

The measured effect sizes for \textit{W+G} remain substantial as a consequence (see Appendix~\ref{sec:cohdArtEngF}), ranging from 1.09 to 1.47 over \textit{W} and from 0.43 to 0.79 over \textit{W+IB}. We note a negative, though negligible, impact of CoT on the \textit{W+G} treatment (effect size: -0.05), consistent with baseline trends.

Lastly, comparing the accuracy scores of the baseline and fine-tuned checkpoint (see Tables~\ref{tab:baseline_scores_art_eng} and \ref{tab:ft_scores_art_eng}) reveals a significant degradation in performance on the Latin-derived conlangs unseen during fine-tuning, with average scores for partitions 8 and 9 dropping from 10.25\% to 1.33\% after fine-tuning.

\section{Discussion}
A cursory examination of a sample prompt from one of the more challenging partitions of our dataset (see Appendices~\ref{sec:fiveF} and \ref{sec:latin_example}) should convince the reader of the immense complexity of the task at hand. Varied cipher keys and grammar book paraphrases (Appendix~\ref{sec:gtempl_appendix}) ensured distinct conlangs and novel documents for the fine-tuned model at test time. Given the high sensitivity of LLMs to even minor prompt formatting changes \citep[inter alia]{voronov_mind_2024, sclar_quantifying_2024}, the fine-tuned checkpoint’s performance demonstrates a remarkable improvement in its explicit learning capabilities under these challenging conditions.

This improvement, however, did not extend to novel linguistic phenomena in the typologically Latin conlangs. Table~\ref{tab:discussion_results} shows accuracy scores for various checkpoints on the eng$\rightarrow$art, \textit{W+G} with CoT version of our test set compared to the baseline LLM. Checkpoints fine-tuned on each ``course'' of our curriculum, in a cumulative fashion, do not suggest that fine-tuning on simpler linguistic features leads to substantial gains on unseen, more complex phenomena. Interestingly, scores from an additional checkpoint, fine-tuned from scratch on instances representing the phenomena in partitions 5 and 7 exclusively, suggest that explicit learning generalizes more effectively in the opposite direction—from complex to simpler sets of features.

\input{table7}

A surprising pattern in our results is that the best-performing checkpoints for unseen Latin-derived phenomena were trained on the least data (see Appendix~\ref{sec:cohdAblOOD}). This may stem from suboptimal fine-tuning hyperparameters set by OpenAI's API or, more likely, from the inherent limitations of supervised fine-tuning in fostering universal explicit learning in LLMs. Addressing these issues could involve exploring DPO \citep{rafailov_direct_2024} or GRPO \citep{shao_deepseekmath_2024} fine-tuning and incorporating \textit{a posteriori} conlangs that mask a wider range of linguistic types—promising directions we leave for future research.

\section{Conclusion}
This study demonstrates that LLMs do possess a capacity for explicit learning, though this ability diminishes as the complexity and variety of the linguistic phenomena at hand increase. We showed that supervised fine-tuning on demonstrations of chains of thought can significantly enhance this ability for similar or simpler phenomena but generalizes poorly to typologically novel or more complex linguistic features. These findings highlight the need for more diverse training sets and the exploration of alternative fine-tuning methods to further study and improve the explicit learning capabilities of LLMs.

\section*{Limitations}

As is common in related research, our work involved a single LLM. This choice was guided by considerations of the financial and environmental cost of our experiments, which required handling long-context inputs. \texttt{GPT-4o-mini} was selected because it offered a sufficiently large context window\footnote{A constraint likewise invoked by \citet{aycock_can_2024} and \citet{hus_back_2024} to explain their choice of \texttt{Gemini-1.5-Flash-001} and \texttt{GPT-4-turbo}, respectively.} and could be fine-tuned (a rare option in 2024). Its accessibility via an API arguably makes replication easier for researchers without access to large GPU clusters. However, as a closed model, its parameter count and fine-tuning specifics (e.g.,~whether it employs LoRA and which modules are targeted) remain undisclosed. Our positive results and internally consistent experiments remain valid as an existence proof in spite of this observation, and we share our datasets and code to allow others to perform experiments with a broader range of artifacts.

Translation directions were unequally represented in our training set due to the high cost of fine-tuning on long contexts and our decision to prioritize the more challenging eng$\rightarrow$art direction in our experiments. Consequently, our observations on the differing roles of CoT across translation directions may partly reflect this imbalance rather than signal an essential difference in the explicit learning performed by the LLM as a function of translation direction.

While we verified that random variations in ciphers had a negligible impact on accuracy scores at our chosen temperature, one drawback of our method is that the resulting conlangs tend to be oversegmented by tokenizers. This does not compromise the ecological validity of our experiments, as low-resource languages are themselves likely to be oversegmented, and the orthographies of many attested languages (e.g., Polish, Georgian, Tashelhit) feature complex graphemic consonant clusters and polyphthongs similar to those occasionally occurring in our conlangs. However, this oversegmentation could very significantly increase training and inference costs on larger LLMs.

One purpose of the \textit{W+IB} configuration in our test set was to ensure that LLMs would not guess the ciphers used, particularly after fine-tuning. We are satisfied that this is not the case. We also verified that more advanced models, such as \texttt{GPT-4o}, \texttt{GPT-o1-preview}, and \texttt{Gemini 1.5 Pro}, similarly fail to decipher our conlangs, as tested via their Web UI with instances containing C-category vocabulary items. While we are confident that explicit learning of the metalinguistic discourses provided in the grammar book excerpts remains the path of least resistance for LLMs under the high variability of our ciphers, we cannot guarantee that future LLMs or LRMs \citep{valmeekam_llms_2024} will not acquire this capability, particularly through fine-tuning. To address this, scaling our technique to these models might require employing a larger number of conlang matrices during fine-tuning, with less-resourced languages documented in open science grammar books\footnote{\label{ft14}Cf. LangSci Press Catalog: \url{https://langsci-press.org/catalogSearch}, accessed: 2024-10-10.} representing particularly valuable candidates.

Finally and most importantly, our experimental setup simulates perfect information retrieval (IR) from the dictionary, by design.\footnote{The bitexts to be translated were devised based on the dictionary and grammar book, and the lemmas chosen to ensure unambiguous translations, permitting our use of accuracy as the evaluation metric.} While this approach avoids introducing IR-related confounding variables into our experiments, it is admittedly a science-fictional scenario in real-world applications. We emphasize that the goal of our experiments was to study explicit learning in LLMs, with MT chosen as the ideal context for this purpose. However, our findings should not be interpreted to suggest that MT, as a task, is easily achievable through explicit learning based solely on grammar books and real-world dictionary entries.

\section*{Ethics Statement}
We acknowledge the environmental cost associated with long-context inputs and fine-tuning in LLMs. To mitigate this, we limited the scope of our experiments and prioritized efficient use of resources.

Since OpenAI policies potentially restrict the publication of inference examples suitable for qualitative analysis, we relied on contrastive experimental settings and exact match/accuracy scores to ensure our results remain interpretable and meaningful, even in the absence of qualitative illustrations.

Our use of cryptographic conlangs to simulate low-resource contexts might offer an ethical advantage by avoiding direct reliance on oppressed and endangered groups for abstract experimental purposes in low-resource natural language processing (NLP). This approach minimizes the risk of exploitative practices and allows for scalable, repeatable experiments without imposing on real-world communities at proof-of-concept stages. We stress, however, that our work complements—and in no way replaces—the invaluable contributions of researchers working directly with less-resourced languages.

Lastly, languages classified as low-resource by MT researchers are the primary concern of fieldwork linguists \citep{thieberger_introduction_2011}, and grammar books and lexicons, sometimes published as open science resources (cf. Footnote~\ref{ft14}) are therefore readily available for languages otherwise deprived of materials suitable for the development of critically needed NLP applications \citep{bhuiyan_lost_2023, deck_ai_2023, young_digital_2015, kornai_digital_2013, ostler_introduction_2014}. We hope that this study will encourage further development of methodologies and resources that improve NLP models’ ability to utilize such grammar books and lexicographical documents to the benefit of less-resourced languages and linguistic diversity.

\section*{Acknowledgments}
This research was funded by the French \textit{Agence Nationale de la Recherche} (ANR) under the project TraLaLaM (``ANR-23-IAS1-0006''). The last two authors' participation was also partly funded through their chairs in the PRAIRIE institute, now PRAIRIE-PSAI, funded by the French national agency ANR, respectively as part of the “Investissements d’avenir” programme under the reference ANR-19-P3IA-0001 and as part of the ``France 2030'' strategy under the reference ANR-23-IACL-0008.

We thank Marine Carpuat for her insightful remarks at an intermediate stage of this work, and the anonymous reviewers for their time and effort in reviewing this manuscript.

Additionally, we acknowledge the use of a large language model for coding support during the development of this project. We are also grateful to the CLEPS infrastructure at Inria Paris for providing computing resources and support during the exploratory phase of this project.

\bibliography{acl_latex}

\appendix

\input{appendix_ciphers}

\input{appendix_grammar_source}

\input{appendix_grammar_template}

\input{dictionaryBD}

\input{appendix_eng_art_prompts_1}

\input{appendix_eng_art_prompts_2}

\input{appendix_5f}

\input{appendix_2m}

\input{latin_test}

\input{appendix_WIB}

\section{Temperature and Cipher Variability Impact}
\label{sec:gtemp_appendix}

We verified that randomness in cipher-key generation for our conlangs and its effects on tokenization had a negligible impact on LLM performance. In order to do so, we selected 14 instances of varying difficulty from our test set (half for `true' French and half for `reversed' French), and enciphered each with 200 different random keys. At the 0.05 temperature which we used in our final evaluations, we found that the average two-sided confidence interval width was 1.17\% at a 95\% certainty level, with mode 0.0\% for the tested baseline (\texttt{gpt-4o-mini}).

Conducting the same experiment at a 0.7 temperature yielded a slightly larger confidence interval width of 1.47\% at a 95\% certainty level, with a mode of 0.0\%.

\input{appendix_cohensD}

\input{appendix_kallini}

\end{document}

%% file: table3.tex
\begin{table}[h!]
\renewcommand{\arraystretch}{1.5} 
\centering
\large 
\resizebox{1\columnwidth}{!}{ 
\begin{tabularx}{\textwidth}{>{\centering\arraybackslash}X >{\centering\arraybackslash}X >{\centering\arraybackslash}X >{\centering\arraybackslash}X >{\centering\arraybackslash}X >{\centering\arraybackslash}X >{\centering\arraybackslash}X}
\toprule
\multicolumn{7}{c}{\textbf{BASELINE --- ENG$\rightarrow$ART}} \\ 
\midrule
\makecell{\textbf{Test Set} \\ \textbf{Partition}} & 
\makecell{\textbf{W} \\ \textbf{(no CoT)}} & 
\makecell{\textbf{W} \\ \textbf{(CoT)}} & 
\makecell{\textbf{W+IB} \\ \textbf{(no CoT)}} & 
\makecell{\textbf{W+IB} \\ \textbf{(CoT)}} & 
\makecell{\textbf{W+G} \\ \textbf{(no CoT)}} & 
\makecell{\textbf{W+G} \\ \textbf{(CoT)}} \\ 
\midrule

1. & \cellcolor[rgb]{0.050,0.030,0.528}\textcolor{white}{0.0} & \cellcolor[rgb]{0.050,0.030,0.528}\textcolor{white}{0.0} & \cellcolor[rgb]{0.274,0.012,0.623}\textcolor{white}{11.0} & \cellcolor[rgb]{0.338,0.006,0.643}\textcolor{white}{15.0} & \cellcolor[rgb]{0.702,0.174,0.557}\textcolor{white}{41.0} & \cellcolor[rgb]{0.923,0.463,0.333}\textcolor{black}{66.0} \\
2m. & \cellcolor[rgb]{0.050,0.030,0.528}\textcolor{white}{0.0} & \cellcolor[rgb]{0.050,0.030,0.528}\textcolor{white}{0.0} & \cellcolor[rgb]{0.115,0.024,0.555}\textcolor{white}{2.5} & \cellcolor[rgb]{0.115,0.024,0.555}\textcolor{white}{2.5} & \cellcolor[rgb]{0.261,0.013,0.618}\textcolor{white}{10.5} & \cellcolor[rgb]{0.579,0.064,0.635}\textcolor{white}{31.0} \\
2f. & \cellcolor[rgb]{0.050,0.030,0.528}\textcolor{white}{0.0} & \cellcolor[rgb]{0.050,0.030,0.528}\textcolor{white}{0.0} & \cellcolor[rgb]{0.075,0.027,0.538}\textcolor{white}{1.0} & \cellcolor[rgb]{0.050,0.030,0.528}\textcolor{white}{0.0} & \cellcolor[rgb]{0.086,0.026,0.543}\textcolor{white}{1.5} & \cellcolor[rgb]{0.124,0.023,0.559}\textcolor{white}{3.0} \\
3. & \cellcolor[rgb]{0.050,0.030,0.528}\textcolor{white}{0.0} & \cellcolor[rgb]{0.050,0.030,0.528}\textcolor{white}{0.0} & \cellcolor[rgb]{0.241,0.015,0.610}\textcolor{white}{9.0} & \cellcolor[rgb]{0.186,0.019,0.587}\textcolor{white}{6.0} & \cellcolor[rgb]{0.506,0.016,0.656}\textcolor{white}{26.0} & \cellcolor[rgb]{0.418,0.001,0.658}\textcolor{white}{20.0} \\
4. & \cellcolor[rgb]{0.050,0.030,0.528}\textcolor{white}{0.0} & \cellcolor[rgb]{0.050,0.030,0.528}\textcolor{white}{0.0} & \cellcolor[rgb]{0.050,0.030,0.528}\textcolor{white}{0.0} & \cellcolor[rgb]{0.050,0.030,0.528}\textcolor{white}{0.0} & \cellcolor[rgb]{0.050,0.030,0.528}\textcolor{white}{0.0} & \cellcolor[rgb]{0.221,0.016,0.602}\textcolor{white}{8.0} \\
5m. & \cellcolor[rgb]{0.050,0.030,0.528}\textcolor{white}{0.0} & \cellcolor[rgb]{0.050,0.030,0.528}\textcolor{white}{0.0} & \cellcolor[rgb]{0.050,0.030,0.528}\textcolor{white}{0.0} & \cellcolor[rgb]{0.050,0.030,0.528}\textcolor{white}{0.0} & \cellcolor[rgb]{0.050,0.030,0.528}\textcolor{white}{0.0} & \cellcolor[rgb]{0.106,0.024,0.551}\textcolor{white}{2.0} \\
5f. & \cellcolor[rgb]{0.050,0.030,0.528}\textcolor{white}{0.0} & \cellcolor[rgb]{0.050,0.030,0.528}\textcolor{white}{0.0} & \cellcolor[rgb]{0.050,0.030,0.528}\textcolor{white}{0.0} & \cellcolor[rgb]{0.050,0.030,0.528}\textcolor{white}{0.0} & \cellcolor[rgb]{0.050,0.030,0.528}\textcolor{white}{0.0} & \cellcolor[rgb]{0.050,0.030,0.528}\textcolor{white}{0.0} \\
6. & \cellcolor[rgb]{0.050,0.030,0.528}\textcolor{white}{0.0} & \cellcolor[rgb]{0.050,0.030,0.528}\textcolor{white}{0.0} & \cellcolor[rgb]{0.050,0.030,0.528}\textcolor{white}{0.0} & \cellcolor[rgb]{0.050,0.030,0.528}\textcolor{white}{0.0} & \cellcolor[rgb]{0.050,0.030,0.528}\textcolor{white}{0.0} & \cellcolor[rgb]{0.050,0.030,0.528}\textcolor{white}{0.0} \\
7m. & \cellcolor[rgb]{0.050,0.030,0.528}\textcolor{white}{0.0} & \cellcolor[rgb]{0.050,0.030,0.528}\textcolor{white}{0.0} & \cellcolor[rgb]{0.050,0.030,0.528}\textcolor{white}{0.0} & \cellcolor[rgb]{0.050,0.030,0.528}\textcolor{white}{0.0} & \cellcolor[rgb]{0.050,0.030,0.528}\textcolor{white}{0.0} & \cellcolor[rgb]{0.050,0.030,0.528}\textcolor{white}{0.0} \\
7f. & \cellcolor[rgb]{0.050,0.030,0.528}\textcolor{white}{0.0} & \cellcolor[rgb]{0.050,0.030,0.528}\textcolor{white}{0.0} & \cellcolor[rgb]{0.050,0.030,0.528}\textcolor{white}{0.0} & \cellcolor[rgb]{0.050,0.030,0.528}\textcolor{white}{0.0} & \cellcolor[rgb]{0.050,0.030,0.528}\textcolor{white}{0.0} & \cellcolor[rgb]{0.050,0.030,0.528}\textcolor{white}{0.0} \\
8a. & \cellcolor[rgb]{0.050,0.030,0.528}\textcolor{white}{0.0} & \cellcolor[rgb]{0.050,0.030,0.528}\textcolor{white}{0.0} & \cellcolor[rgb]{0.075,0.027,0.538}\textcolor{white}{1.0} & \cellcolor[rgb]{0.075,0.027,0.538}\textcolor{white}{1.0} & \cellcolor[rgb]{0.075,0.027,0.538}\textcolor{white}{1.0} & \cellcolor[rgb]{0.124,0.023,0.559}\textcolor{white}{3.0} \\
8g. & \cellcolor[rgb]{0.050,0.030,0.528}\textcolor{white}{0.0} & \cellcolor[rgb]{0.050,0.030,0.528}\textcolor{white}{0.0} & \cellcolor[rgb]{0.050,0.030,0.528}\textcolor{white}{0.0} & \cellcolor[rgb]{0.050,0.030,0.528}\textcolor{white}{0.0} & \cellcolor[rgb]{0.075,0.027,0.538}\textcolor{white}{1.0} & \cellcolor[rgb]{0.050,0.030,0.528}\textcolor{white}{0.0} \\
9a. & \cellcolor[rgb]{0.050,0.030,0.528}\textcolor{white}{0.0} & \cellcolor[rgb]{0.050,0.030,0.528}\textcolor{white}{0.0} & \cellcolor[rgb]{0.050,0.030,0.528}\textcolor{white}{0.0} & \cellcolor[rgb]{0.050,0.030,0.528}\textcolor{white}{0.0} & \cellcolor[rgb]{0.050,0.030,0.528}\textcolor{white}{0.0} & \cellcolor[rgb]{0.050,0.030,0.528}\textcolor{white}{0.0} \\
9g. & \cellcolor[rgb]{0.050,0.030,0.528}\textcolor{white}{0.0} & \cellcolor[rgb]{0.050,0.030,0.528}\textcolor{white}{0.0} & \cellcolor[rgb]{0.050,0.030,0.528}\textcolor{white}{0.0} & \cellcolor[rgb]{0.050,0.030,0.528}\textcolor{white}{0.0} & \cellcolor[rgb]{0.075,0.027,0.538}\textcolor{white}{1.0} & \cellcolor[rgb]{0.050,0.030,0.528}\textcolor{white}{0.0} \\

\midrule
\textbf{Average} & \cellcolor[rgb]{0.050,0.030,0.528}\textcolor{white}{0.0} & \cellcolor[rgb]{0.050,0.030,0.528}\textcolor{white}{0.0} & \cellcolor[rgb]{0.096,0.025,0.547}\textcolor{white}{1.8} & \cellcolor[rgb]{0.096,0.025,0.547}\textcolor{white}{1.8} & \cellcolor[rgb]{0.186,0.019,0.587}\textcolor{white}{5.9} & \cellcolor[rgb]{0.248,0.014,0.613}\textcolor{white}{9.5} \\
\bottomrule
\end{tabularx}
}
\caption{Baseline accuracy for eng$\rightarrow$art.}
\label{tab:baseline_eng_art}
\end{table}

%% file: table4.tex
\begin{table}[h!]
\renewcommand{\arraystretch}{1.5} 
\centering
\large 
\resizebox{1\columnwidth}{!}{ 
\begin{tabularx}{\textwidth}{>{\centering\arraybackslash}X >{\centering\arraybackslash}X >{\centering\arraybackslash}X >{\centering\arraybackslash}X >{\centering\arraybackslash}X >{\centering\arraybackslash}X >{\centering\arraybackslash}X}
\toprule
\multicolumn{7}{c}{\textbf{FINETUNED --- ENG$\rightarrow$ART}} \\ 
\midrule
\makecell{\textbf{Test Set} \\ \textbf{Partition}} & 
\makecell{\textbf{W} \\ \textbf{(no CoT)}} & 
\makecell{\textbf{W} \\ \textbf{(CoT)}} & 
\makecell{\textbf{W+IB} \\ \textbf{(no CoT)}} & 
\makecell{\textbf{W+IB} \\ \textbf{(CoT)}} & 
\makecell{\textbf{W+G} \\ \textbf{(no CoT)}} & 
\makecell{\textbf{W+G} \\ \textbf{(CoT)}} \\ 
\midrule
1. & \cellcolor[rgb]{0.050,0.030,0.528}\textcolor{white}{0.0} & \cellcolor[rgb]{0.050,0.030,0.528}\textcolor{white}{0.0} & \cellcolor[rgb]{0.319,0.008,0.638}\textcolor{white}{14.0} & \cellcolor[rgb]{0.200,0.018,0.593}\textcolor{white}{7.0} & \cellcolor[rgb]{0.978,0.861,0.143}\textcolor{black}{93.0} & \cellcolor[rgb]{0.952,0.942,0.153}\textcolor{black}{98.0} \\
2m. & \cellcolor[rgb]{0.050,0.030,0.528}\textcolor{white}{0.0} & \cellcolor[rgb]{0.050,0.030,0.528}\textcolor{white}{0.0} & \cellcolor[rgb]{0.115,0.024,0.555}\textcolor{white}{2.5} & \cellcolor[rgb]{0.164,0.020,0.577}\textcolor{white}{5.0} & \cellcolor[rgb]{0.951,0.523,0.292}\textcolor{black}{70.5} & \cellcolor[rgb]{0.968,0.895,0.147}\textcolor{black}{95.0} \\
2f. & \cellcolor[rgb]{0.050,0.030,0.528}\textcolor{white}{0.0} & \cellcolor[rgb]{0.050,0.030,0.528}\textcolor{white}{0.0} & \cellcolor[rgb]{0.050,0.030,0.528}\textcolor{white}{0.0} & \cellcolor[rgb]{0.164,0.020,0.577}\textcolor{white}{5.0} & \cellcolor[rgb]{0.890,0.406,0.373}\textcolor{black}{61.0} & \cellcolor[rgb]{0.987,0.822,0.144}\textcolor{black}{91.0} \\
3. & \cellcolor[rgb]{0.050,0.030,0.528}\textcolor{white}{0.0} & \cellcolor[rgb]{0.050,0.030,0.528}\textcolor{white}{0.0} & \cellcolor[rgb]{0.164,0.020,0.577}\textcolor{white}{5.0} & \cellcolor[rgb]{0.124,0.023,0.559}\textcolor{white}{3.0} & \cellcolor[rgb]{0.931,0.478,0.323}\textcolor{black}{67.0} & \cellcolor[rgb]{0.976,0.597,0.245}\textcolor{black}{76.0} \\
4. & \cellcolor[rgb]{0.050,0.030,0.528}\textcolor{white}{0.0} & \cellcolor[rgb]{0.050,0.030,0.528}\textcolor{white}{0.0} & \cellcolor[rgb]{0.050,0.030,0.528}\textcolor{white}{0.0} & \cellcolor[rgb]{0.050,0.030,0.528}\textcolor{white}{0.0} & \cellcolor[rgb]{0.905,0.430,0.356}\textcolor{black}{63.0} & \cellcolor[rgb]{0.987,0.822,0.144}\textcolor{black}{91.0} \\
5m. & \cellcolor[rgb]{0.050,0.030,0.528}\textcolor{white}{0.0} & \cellcolor[rgb]{0.050,0.030,0.528}\textcolor{white}{0.0} & \cellcolor[rgb]{0.050,0.030,0.528}\textcolor{white}{0.0} & \cellcolor[rgb]{0.050,0.030,0.528}\textcolor{white}{0.0} & \cellcolor[rgb]{0.646,0.121,0.599}\textcolor{white}{36.0} & \cellcolor[rgb]{0.994,0.747,0.164}\textcolor{black}{86.0} \\
5f. & \cellcolor[rgb]{0.050,0.030,0.528}\textcolor{white}{0.0} & \cellcolor[rgb]{0.050,0.030,0.528}\textcolor{white}{0.0} & \cellcolor[rgb]{0.050,0.030,0.528}\textcolor{white}{0.0} & \cellcolor[rgb]{0.050,0.030,0.528}\textcolor{white}{0.0} & \cellcolor[rgb]{0.656,0.130,0.592}\textcolor{white}{37.0} & \cellcolor[rgb]{0.988,0.652,0.211}\textcolor{black}{80.0} \\
6. & \cellcolor[rgb]{0.050,0.030,0.528}\textcolor{white}{0.0} & \cellcolor[rgb]{0.050,0.030,0.528}\textcolor{white}{0.0} & \cellcolor[rgb]{0.050,0.030,0.528}\textcolor{white}{0.0} & \cellcolor[rgb]{0.050,0.030,0.528}\textcolor{white}{0.0} & \cellcolor[rgb]{0.460,0.004,0.660}\textcolor{white}{23.0} & \cellcolor[rgb]{0.989,0.810,0.145}\textcolor{black}{90.0} \\
7m. & \cellcolor[rgb]{0.050,0.030,0.528}\textcolor{white}{0.0} & \cellcolor[rgb]{0.050,0.030,0.528}\textcolor{white}{0.0} & \cellcolor[rgb]{0.050,0.030,0.528}\textcolor{white}{0.0} & \cellcolor[rgb]{0.050,0.030,0.528}\textcolor{white}{0.0} & \cellcolor[rgb]{0.350,0.004,0.646}\textcolor{white}{16.0} & \cellcolor[rgb]{0.989,0.810,0.145}\textcolor{black}{90.0} \\
7f. & \cellcolor[rgb]{0.050,0.030,0.528}\textcolor{white}{0.0} & \cellcolor[rgb]{0.050,0.030,0.528}\textcolor{white}{0.0} & \cellcolor[rgb]{0.050,0.030,0.528}\textcolor{white}{0.0} & \cellcolor[rgb]{0.050,0.030,0.528}\textcolor{white}{0.0} & \cellcolor[rgb]{0.406,0.001,0.657}\textcolor{white}{19.5} & \cellcolor[rgb]{0.988,0.652,0.211}\textcolor{black}{80.0} \\
8a. & \cellcolor[rgb]{0.050,0.030,0.528}\textcolor{white}{0.0} & \cellcolor[rgb]{0.050,0.030,0.528}\textcolor{white}{0.0} & \cellcolor[rgb]{0.050,0.030,0.528}\textcolor{white}{0.0} & \cellcolor[rgb]{0.075,0.027,0.538}\textcolor{white}{1.0} & \cellcolor[rgb]{0.106,0.024,0.551}\textcolor{white}{2.0} & \cellcolor[rgb]{0.124,0.023,0.559}\textcolor{white}{3.0} \\
8g. & \cellcolor[rgb]{0.050,0.030,0.528}\textcolor{white}{0.0} & \cellcolor[rgb]{0.050,0.030,0.528}\textcolor{white}{0.0} & \cellcolor[rgb]{0.050,0.030,0.528}\textcolor{white}{0.0} & \cellcolor[rgb]{0.075,0.027,0.538}\textcolor{white}{1.0} & \cellcolor[rgb]{0.075,0.027,0.538}\textcolor{white}{1.0} & \cellcolor[rgb]{0.050,0.030,0.528}\textcolor{white}{0.0} \\
9a. & \cellcolor[rgb]{0.050,0.030,0.528}\textcolor{white}{0.0} & \cellcolor[rgb]{0.050,0.030,0.528}\textcolor{white}{0.0} & \cellcolor[rgb]{0.050,0.030,0.528}\textcolor{white}{0.0} & \cellcolor[rgb]{0.050,0.030,0.528}\textcolor{white}{0.0} & \cellcolor[rgb]{0.050,0.030,0.528}\textcolor{white}{0.0} & \cellcolor[rgb]{0.050,0.030,0.528}\textcolor{white}{0.0} \\
9g. & \cellcolor[rgb]{0.050,0.030,0.528}\textcolor{white}{0.0} & \cellcolor[rgb]{0.050,0.030,0.528}\textcolor{white}{0.0} & \cellcolor[rgb]{0.050,0.030,0.528}\textcolor{white}{0.0} & \cellcolor[rgb]{0.050,0.030,0.528}\textcolor{white}{0.0} & \cellcolor[rgb]{0.075,0.027,0.538}\textcolor{white}{1.0} & \cellcolor[rgb]{0.050,0.030,0.528}\textcolor{white}{0.0} \\
\midrule
\textbf{Average} & \cellcolor[rgb]{0.050,0.030,0.528}\textcolor{white}{0.0} & \cellcolor[rgb]{0.050,0.030,0.528}\textcolor{white}{0.0} & \cellcolor[rgb]{0.096,0.025,0.547}\textcolor{white}{1.7} & \cellcolor[rgb]{0.086,0.026,0.543}\textcolor{white}{1.4} & \cellcolor[rgb]{0.631,0.108,0.608}\textcolor{white}{34.9} & \cellcolor[rgb]{0.905,0.430,0.356}\textcolor{black}{62.9} \\
\bottomrule
\end{tabularx}
}
\caption{Fine-tuned checkpoint accuracy for eng$\rightarrow$art.}
\label{tab:ft_scores_eng_art}
\end{table}

%% file: table5.tex
\begin{table}[h!]
\renewcommand{\arraystretch}{1.5} 
\centering
\large 
\resizebox{1\columnwidth}{!}{ 
\begin{tabularx}{\textwidth}{>{\centering\arraybackslash}X >{\centering\arraybackslash}X >{\centering\arraybackslash}X >{\centering\arraybackslash}X >{\centering\arraybackslash}X >{\centering\arraybackslash}X >{\centering\arraybackslash}X}
\toprule
\multicolumn{7}{c}{\textbf{BASELINE --- ART$\rightarrow$ENG}} \\ 
\midrule
\makecell{\textbf{Test Set} \\ \textbf{Partition}} & 
\makecell{\textbf{W} \\ \textbf{(no CoT)}} & 
\makecell{\textbf{W} \\ \textbf{(CoT)}} & 
\makecell{\textbf{W+IB} \\ \textbf{(no CoT)}} & 
\makecell{\textbf{W+IB} \\ \textbf{(CoT)}} & 
\makecell{\textbf{W+G} \\ \textbf{(no CoT)}} & 
\makecell{\textbf{W+G} \\ \textbf{(CoT)}} \\
\midrule

1. & \cellcolor[rgb]{0.050,0.030,0.528}\textcolor{white}{0.0} & \cellcolor[rgb]{0.075,0.027,0.538}\textcolor{white}{1.0} & \cellcolor[rgb]{0.693,0.165,0.565}\textcolor{white}{40.0} & \cellcolor[rgb]{0.938,0.493,0.313}\textcolor{black}{68.0} & \cellcolor[rgb]{0.957,0.928,0.152}\textcolor{black}{97.0} & \cellcolor[rgb]{0.940,0.975,0.131}\textcolor{black}{100.0} \\
2m & \cellcolor[rgb]{0.050,0.030,0.528}\textcolor{white}{0.0} & \cellcolor[rgb]{0.124,0.023,0.559}\textcolor{white}{3.0} & \cellcolor[rgb]{0.430,0.001,0.659}\textcolor{white}{21.0} & \cellcolor[rgb]{0.875,0.383,0.390}\textcolor{black}{59.0} & \cellcolor[rgb]{0.994,0.747,0.164}\textcolor{black}{86.0} & \cellcolor[rgb]{0.964,0.908,0.149}\textcolor{black}{96.0} \\
2f & \cellcolor[rgb]{0.050,0.030,0.528}\textcolor{white}{0.0} & \cellcolor[rgb]{0.124,0.023,0.559}\textcolor{white}{3.0} & \cellcolor[rgb]{0.776,0.254,0.491}\textcolor{white}{48.0} & \cellcolor[rgb]{0.959,0.543,0.279}\textcolor{black}{72.0} & \cellcolor[rgb]{0.994,0.747,0.164}\textcolor{black}{86.0} & \cellcolor[rgb]{0.974,0.875,0.144}\textcolor{black}{94.0} \\
3. & \cellcolor[rgb]{0.050,0.030,0.528}\textcolor{white}{0.0} & \cellcolor[rgb]{0.050,0.030,0.528}\textcolor{white}{0.0} & \cellcolor[rgb]{0.646,0.121,0.599}\textcolor{white}{36.0} & \cellcolor[rgb]{0.460,0.004,0.660}\textcolor{white}{23.0} & \cellcolor[rgb]{0.621,0.099,0.614}\textcolor{white}{34.0} & \cellcolor[rgb]{0.605,0.086,0.623}\textcolor{white}{33.0} \\
4. & \cellcolor[rgb]{0.050,0.030,0.528}\textcolor{white}{0.0} & \cellcolor[rgb]{0.050,0.030,0.528}\textcolor{white}{0.0} & \cellcolor[rgb]{0.164,0.020,0.577}\textcolor{white}{5.0} & \cellcolor[rgb]{0.274,0.012,0.623}\textcolor{white}{11.0} & \cellcolor[rgb]{0.306,0.009,0.634}\textcolor{white}{13.0} & \cellcolor[rgb]{0.524,0.025,0.653}\textcolor{white}{27.0} \\
5m. & \cellcolor[rgb]{0.050,0.030,0.528}\textcolor{white}{0.0} & \cellcolor[rgb]{0.050,0.030,0.528}\textcolor{white}{0.0} & \cellcolor[rgb]{0.106,0.024,0.551}\textcolor{white}{2.0} & \cellcolor[rgb]{0.200,0.018,0.593}\textcolor{white}{7.0} & \cellcolor[rgb]{0.338,0.006,0.643}\textcolor{white}{15.0} & \cellcolor[rgb]{0.477,0.007,0.660}\textcolor{white}{24.0} \\
5f. & \cellcolor[rgb]{0.050,0.030,0.528}\textcolor{white}{0.0} & \cellcolor[rgb]{0.050,0.030,0.528}\textcolor{white}{0.0} & \cellcolor[rgb]{0.164,0.020,0.577}\textcolor{white}{5.0} & \cellcolor[rgb]{0.287,0.011,0.627}\textcolor{white}{12.0} & \cellcolor[rgb]{0.287,0.011,0.627}\textcolor{white}{12.0} & \cellcolor[rgb]{0.350,0.004,0.646}\textcolor{white}{16.0} \\
6 & \cellcolor[rgb]{0.050,0.030,0.528}\textcolor{white}{0.0} & \cellcolor[rgb]{0.050,0.030,0.528}\textcolor{white}{0.0} & \cellcolor[rgb]{0.106,0.024,0.551}\textcolor{white}{2.0} & \cellcolor[rgb]{0.050,0.030,0.528}\textcolor{white}{0.0} & \cellcolor[rgb]{0.506,0.016,0.656}\textcolor{white}{26.0} & \cellcolor[rgb]{0.149,0.021,0.571}\textcolor{white}{4.0} \\
7m. & \cellcolor[rgb]{0.050,0.030,0.528}\textcolor{white}{0.0} & \cellcolor[rgb]{0.050,0.030,0.528}\textcolor{white}{0.0} & \cellcolor[rgb]{0.106,0.024,0.551}\textcolor{white}{2.0} & \cellcolor[rgb]{0.106,0.024,0.551}\textcolor{white}{2.0} & \cellcolor[rgb]{0.319,0.008,0.638}\textcolor{white}{14.0} & \cellcolor[rgb]{0.149,0.021,0.571}\textcolor{white}{4.0} \\
7f. & \cellcolor[rgb]{0.050,0.030,0.528}\textcolor{white}{0.0} & \cellcolor[rgb]{0.050,0.030,0.528}\textcolor{white}{0.0} & \cellcolor[rgb]{0.106,0.024,0.551}\textcolor{white}{2.0} & \cellcolor[rgb]{0.106,0.024,0.551}\textcolor{white}{2.0} & \cellcolor[rgb]{0.287,0.011,0.627}\textcolor{white}{12.0} & \cellcolor[rgb]{0.164,0.020,0.577}\textcolor{white}{5.0} \\
8a & \cellcolor[rgb]{0.164,0.020,0.577}\textcolor{white}{5.0} & \cellcolor[rgb]{0.221,0.016,0.602}\textcolor{white}{8.0} & \cellcolor[rgb]{0.200,0.018,0.593}\textcolor{white}{7.0} & \cellcolor[rgb]{0.387,0.001,0.654}\textcolor{white}{18.0} & \cellcolor[rgb]{0.186,0.019,0.587}\textcolor{white}{6.0} & \cellcolor[rgb]{0.350,0.004,0.646}\textcolor{white}{16.0} \\
8g & \cellcolor[rgb]{0.287,0.011,0.627}\textcolor{white}{12.0} & \cellcolor[rgb]{0.221,0.016,0.602}\textcolor{white}{8.0} & \cellcolor[rgb]{0.319,0.008,0.638}\textcolor{white}{14.0} & \cellcolor[rgb]{0.387,0.001,0.654}\textcolor{white}{18.0} & \cellcolor[rgb]{0.338,0.006,0.643}\textcolor{white}{15.0} & \cellcolor[rgb]{0.241,0.015,0.610}\textcolor{white}{9.0} \\
9a & \cellcolor[rgb]{0.106,0.024,0.551}\textcolor{white}{2.0} & \cellcolor[rgb]{0.200,0.018,0.593}\textcolor{white}{7.0} & \cellcolor[rgb]{0.124,0.023,0.559}\textcolor{white}{3.0} & \cellcolor[rgb]{0.369,0.003,0.651}\textcolor{white}{17.0} & \cellcolor[rgb]{0.221,0.016,0.602}\textcolor{white}{8.0} & \cellcolor[rgb]{0.350,0.004,0.646}\textcolor{white}{16.0} \\
9g & \cellcolor[rgb]{0.186,0.019,0.587}\textcolor{white}{6.0} & \cellcolor[rgb]{0.149,0.021,0.571}\textcolor{white}{4.0} & \cellcolor[rgb]{0.319,0.008,0.638}\textcolor{white}{14.0} & \cellcolor[rgb]{0.255,0.014,0.615}\textcolor{white}{10.0} & \cellcolor[rgb]{0.306,0.009,0.634}\textcolor{white}{13.0} & \cellcolor[rgb]{0.255,0.014,0.615}\textcolor{white}{10.0} \\
\midrule
\textbf{Average} & \cellcolor[rgb]{0.096,0.025,0.547}\textcolor{white}{1.8} & \cellcolor[rgb]{0.115,0.024,0.555}\textcolor{white}{2.4} & \cellcolor[rgb]{0.325,0.007,0.640}\textcolor{white}{14.4} & \cellcolor[rgb]{0.460,0.004,0.660}\textcolor{white}{22.8} & \cellcolor[rgb]{0.579,0.064,0.635}\textcolor{white}{31.2} & \cellcolor[rgb]{0.595,0.077,0.628}\textcolor{white}{32.4} \\
\bottomrule
\end{tabularx}
}
\caption{Baseline accuracy for art$\rightarrow$eng.}
\label{tab:baseline_scores_art_eng}
\end{table}

%% file: table6.tex
\begin{table}[h!]
\renewcommand{\arraystretch}{1.5} 
\centering
\large 
\resizebox{1\columnwidth}{!}{ 
\begin{tabularx}{\textwidth}{>{\centering\arraybackslash}X >{\centering\arraybackslash}X >{\centering\arraybackslash}X >{\centering\arraybackslash}X >{\centering\arraybackslash}X >{\centering\arraybackslash}X >{\centering\arraybackslash}X}
\toprule
\multicolumn{7}{c}{\textbf{FINETUNED ---   ART$\rightarrow$ENG}} \\ 
\midrule
\makecell{\textbf{Test Set} \\ \textbf{Partition}} & 
\makecell{\textbf{W} \\ \textbf{(no CoT)}} & 
\makecell{\textbf{W} \\ \textbf{(CoT)}} & 
\makecell{\textbf{W+IB} \\ \textbf{(no CoT)}} & 
\makecell{\textbf{W+IB} \\ \textbf{(CoT)}} & 
\makecell{\textbf{W+G} \\ \textbf{(no CoT)}} & 
\makecell{\textbf{W+G} \\ \textbf{(CoT)}} \\ 
\midrule

1. & \cellcolor[rgb]{0.991,0.670,0.202}\textcolor{black}{81.0} & \cellcolor[rgb]{0.350,0.004,0.646}\textcolor{white}{16.0} & \cellcolor[rgb]{0.940,0.975,0.131}\textcolor{black}{100.0} & \cellcolor[rgb]{0.995,0.729,0.172}\textcolor{black}{85.0} & \cellcolor[rgb]{0.940,0.975,0.131}\textcolor{black}{100.0} & \cellcolor[rgb]{0.940,0.975,0.131}\textcolor{black}{100.0} \\
2m & \cellcolor[rgb]{0.798,0.280,0.470}\textcolor{white}{50.0} & \cellcolor[rgb]{0.430,0.001,0.659}\textcolor{white}{21.0} & \cellcolor[rgb]{0.944,0.962,0.147}\textcolor{black}{99.0} & \cellcolor[rgb]{0.957,0.928,0.152}\textcolor{black}{97.0} & \cellcolor[rgb]{0.940,0.975,0.131}\textcolor{black}{100.0} & \cellcolor[rgb]{0.940,0.975,0.131}\textcolor{black}{100.0} \\
2f & \cellcolor[rgb]{0.881,0.393,0.383}\textcolor{black}{60.0} & \cellcolor[rgb]{0.646,0.121,0.599}\textcolor{white}{36.0} & \cellcolor[rgb]{0.940,0.975,0.131}\textcolor{black}{100.0} & \cellcolor[rgb]{0.957,0.928,0.152}\textcolor{black}{97.0} & \cellcolor[rgb]{0.940,0.975,0.131}\textcolor{black}{100.0} & \cellcolor[rgb]{0.940,0.975,0.131}\textcolor{black}{100.0} \\
3. & \cellcolor[rgb]{0.050,0.030,0.528}\textcolor{white}{0.0} & \cellcolor[rgb]{0.106,0.024,0.551}\textcolor{white}{2.0} & \cellcolor[rgb]{0.679,0.152,0.575}\textcolor{white}{39.0} & \cellcolor[rgb]{0.387,0.001,0.654}\textcolor{white}{18.0} & \cellcolor[rgb]{0.991,0.670,0.202}\textcolor{black}{81.0} & \cellcolor[rgb]{0.994,0.717,0.177}\textcolor{black}{84.0} \\
4. & \cellcolor[rgb]{0.075,0.027,0.538}\textcolor{white}{1.0} & \cellcolor[rgb]{0.149,0.021,0.571}\textcolor{white}{4.0} & \cellcolor[rgb]{0.646,0.121,0.599}\textcolor{white}{36.0} & \cellcolor[rgb]{0.369,0.003,0.651}\textcolor{white}{17.0} & \cellcolor[rgb]{0.973,0.586,0.252}\textcolor{black}{75.0} & \cellcolor[rgb]{0.905,0.430,0.356}\textcolor{black}{63.0} \\
5m. & \cellcolor[rgb]{0.124,0.023,0.559}\textcolor{white}{3.0} & \cellcolor[rgb]{0.274,0.012,0.623}\textcolor{white}{11.0} & \cellcolor[rgb]{0.524,0.025,0.653}\textcolor{white}{27.0} & \cellcolor[rgb]{0.350,0.004,0.646}\textcolor{white}{16.0} & \cellcolor[rgb]{0.963,0.554,0.272}\textcolor{black}{73.0} & \cellcolor[rgb]{0.953,0.528,0.289}\textcolor{black}{71.0} \\
5f. & \cellcolor[rgb]{0.075,0.027,0.538}\textcolor{white}{1.0} & \cellcolor[rgb]{0.186,0.019,0.587}\textcolor{white}{6.0} & \cellcolor[rgb]{0.430,0.001,0.659}\textcolor{white}{21.0} & \cellcolor[rgb]{0.338,0.006,0.643}\textcolor{white}{15.0} & \cellcolor[rgb]{0.905,0.430,0.356}\textcolor{black}{63.0} & \cellcolor[rgb]{0.910,0.439,0.350}\textcolor{black}{64.0} \\
6 & \cellcolor[rgb]{0.221,0.016,0.602}\textcolor{white}{8.0} & \cellcolor[rgb]{0.050,0.030,0.528}\textcolor{white}{0.0} & \cellcolor[rgb]{0.857,0.356,0.410}\textcolor{black}{57.0} & \cellcolor[rgb]{0.241,0.015,0.610}\textcolor{white}{9.0} & \cellcolor[rgb]{0.944,0.962,0.147}\textcolor{black}{99.0} & \cellcolor[rgb]{0.978,0.861,0.143}\textcolor{black}{93.0} \\
7m. & \cellcolor[rgb]{0.448,0.002,0.660}\textcolor{white}{22.0} & \cellcolor[rgb]{0.050,0.030,0.528}\textcolor{white}{0.0} & \cellcolor[rgb]{0.833,0.325,0.434}\textcolor{white}{54.0} & \cellcolor[rgb]{0.186,0.019,0.587}\textcolor{white}{6.0} & \cellcolor[rgb]{0.952,0.942,0.153}\textcolor{black}{98.0} & \cellcolor[rgb]{0.993,0.778,0.153}\textcolor{black}{88.0} \\
7f. & \cellcolor[rgb]{0.221,0.016,0.602}\textcolor{white}{8.0} & \cellcolor[rgb]{0.050,0.030,0.528}\textcolor{white}{0.0} & \cellcolor[rgb]{0.890,0.406,0.373}\textcolor{black}{61.0} & \cellcolor[rgb]{0.241,0.015,0.610}\textcolor{white}{9.0} & \cellcolor[rgb]{0.957,0.928,0.152}\textcolor{black}{97.0} & \cellcolor[rgb]{0.987,0.822,0.144}\textcolor{black}{91.0} \\
8a & \cellcolor[rgb]{0.075,0.027,0.538}\textcolor{white}{1.0} & \cellcolor[rgb]{0.075,0.027,0.538}\textcolor{white}{1.0} & \cellcolor[rgb]{0.075,0.027,0.538}\textcolor{white}{1.0} & \cellcolor[rgb]{0.200,0.018,0.593}\textcolor{white}{7.0} & \cellcolor[rgb]{0.164,0.020,0.577}\textcolor{white}{5.0} & \cellcolor[rgb]{0.149,0.021,0.571}\textcolor{white}{4.0} \\
8g & \cellcolor[rgb]{0.075,0.027,0.538}\textcolor{white}{1.0} & \cellcolor[rgb]{0.106,0.024,0.551}\textcolor{white}{2.0} & \cellcolor[rgb]{0.149,0.021,0.571}\textcolor{white}{4.0} & \cellcolor[rgb]{0.149,0.021,0.571}\textcolor{white}{4.0} & \cellcolor[rgb]{0.050,0.030,0.528}\textcolor{white}{0.0} & \cellcolor[rgb]{0.075,0.027,0.538}\textcolor{white}{1.0} \\
9a & \cellcolor[rgb]{0.050,0.030,0.528}\textcolor{white}{0.0} & \cellcolor[rgb]{0.050,0.030,0.528}\textcolor{white}{0.0} & \cellcolor[rgb]{0.050,0.030,0.528}\textcolor{white}{0.0} & \cellcolor[rgb]{0.050,0.030,0.528}\textcolor{white}{0.0} & \cellcolor[rgb]{0.050,0.030,0.528}\textcolor{white}{0.0} & \cellcolor[rgb]{0.050,0.030,0.528}\textcolor{white}{0.0} \\
9g & \cellcolor[rgb]{0.050,0.030,0.528}\textcolor{white}{0.0} & \cellcolor[rgb]{0.050,0.030,0.528}\textcolor{white}{0.0} & \cellcolor[rgb]{0.050,0.030,0.528}\textcolor{white}{0.0} & \cellcolor[rgb]{0.050,0.030,0.528}\textcolor{white}{0.0} & \cellcolor[rgb]{0.050,0.030,0.528}\textcolor{white}{0.0} & \cellcolor[rgb]{0.075,0.027,0.538}\textcolor{white}{1.0} \\
\midrule
\textbf{Average} & \cellcolor[rgb]{0.369,0.003,0.651}\textcolor{white}{16.9} & \cellcolor[rgb]{0.207,0.017,0.596}\textcolor{white}{7.1} & \cellcolor[rgb]{0.723,0.196,0.539}\textcolor{white}{42.8} & \cellcolor[rgb]{0.524,0.025,0.653}\textcolor{white}{27.1} & \cellcolor[rgb]{0.907,0.435,0.353}\textcolor{black}{63.6} & \cellcolor[rgb]{0.893,0.411,0.370}\textcolor{black}{61.4} \\

\bottomrule
\end{tabularx}
}
\caption{Fine-tuned checkpoint accuracy for art$\rightarrow$eng.}
\label{tab:ft_scores_art_eng}
\end{table}

%% file: table7.tex
\begin{table*}[h!]
\centering
\small 
\resizebox{0.9\textwidth}{!}{ 
\begin{tabularx}{\textwidth}{>{\centering\arraybackslash}X>{\centering\arraybackslash}X>{\centering\arraybackslash}X>{\centering\arraybackslash}X>{\centering\arraybackslash}X>{\centering\arraybackslash}X>{\centering\arraybackslash}X>{\centering\arraybackslash}X>{\centering\arraybackslash}X>{\centering\arraybackslash}X>{\centering\arraybackslash}X}
\midrule 
\multicolumn{11}{c}{\textbf{VARIOUS CHECKPOINTS --- W + G --- ENG$\rightarrow$ART}} \\ 
\midrule 
\scalebox{0.8}{\makecell{\textbf{Test Set} \\ \textbf{Partition}}} & 
\scalebox{0.8}{\textbf{Baseline}} & 
\scalebox{0.5}{\makecell{\textbf{Fine-tuned on} \\ \textbf{Course 1}}} & 
\scalebox{0.5}{\makecell{\textbf{FT on Courses} \\ \textbf{Up to 2}}} & 
\scalebox{0.5}{\makecell{\textbf{FT on Courses} \\ \textbf{Up to 3}}} & 
\scalebox{0.5}{\makecell{\textbf{FT on Courses} \\ \textbf{Up to 4}}} & 
\scalebox{0.5}{\makecell{\textbf{FT on Courses} \\ \textbf{Up to 5}}} & 
\scalebox{0.5}{\makecell{\textbf{FT on Courses} \\ \textbf{Up to 6}}} & 
\scalebox{0.5}{\makecell{\textbf{FT on Courses} \\ \textbf{Up to 7}}} & 
\scalebox{0.5}{\makecell{\textbf{FT on 5} \\ \textbf{\& 7 Only}}} &
\scalebox{0.8}{\textbf{Full FT}} \\
\midrule 

1. & \cellcolor[rgb]{0.923,0.463,0.333}\textcolor{black}{66.0} & \cellcolor[rgb]{0.993,0.699,0.186}\textcolor{black}{83.0} & \cellcolor[rgb]{0.952,0.942,0.153}\textcolor{black}{98.0} & \cellcolor[rgb]{0.952,0.942,0.153}\textcolor{black}{98.0} & \cellcolor[rgb]{0.952,0.942,0.153}\textcolor{black}{98.0} & \cellcolor[rgb]{0.964,0.908,0.149}\textcolor{black}{96.0} & \cellcolor[rgb]{0.952,0.942,0.153}\textcolor{black}{98.0} & \cellcolor[rgb]{0.957,0.928,0.152}\textcolor{black}{97.0} & \cellcolor[rgb]{0.983,0.842,0.142}\textcolor{black}{92.0} & \cellcolor[rgb]{0.952,0.942,0.153}\textcolor{black}{98.0} \\
2m & \cellcolor[rgb]{0.579,0.064,0.635}\textcolor{white}{31.0} & \cellcolor[rgb]{0.584,0.069,0.633}\textcolor{white}{31.5} & \cellcolor[rgb]{0.983,0.842,0.142}\textcolor{black}{92.0} & \cellcolor[rgb]{0.976,0.597,0.245}\textcolor{black}{76.0} & \cellcolor[rgb]{0.993,0.699,0.186}\textcolor{black}{83.0} & \cellcolor[rgb]{0.983,0.842,0.142}\textcolor{black}{92.0} & \cellcolor[rgb]{0.974,0.875,0.144}\textcolor{black}{94.0} & \cellcolor[rgb]{0.991,0.791,0.149}\textcolor{black}{89.0} & \cellcolor[rgb]{0.993,0.699,0.186}\textcolor{black}{83.0} & \cellcolor[rgb]{0.968,0.895,0.147}\textcolor{black}{95.0} \\
2f & \cellcolor[rgb]{0.124,0.023,0.559}\textcolor{white}{3.0} & \cellcolor[rgb]{0.214,0.017,0.599}\textcolor{white}{7.5} & \cellcolor[rgb]{0.931,0.478,0.323}\textcolor{black}{67.0} & \cellcolor[rgb]{0.943,0.503,0.306}\textcolor{black}{69.0} & \cellcolor[rgb]{0.748,0.223,0.517}\textcolor{white}{45.0} & \cellcolor[rgb]{0.988,0.652,0.211}\textcolor{black}{80.0} & \cellcolor[rgb]{0.938,0.493,0.313}\textcolor{black}{68.0} & \cellcolor[rgb]{0.994,0.747,0.164}\textcolor{black}{86.0} & \cellcolor[rgb]{0.976,0.597,0.245}\textcolor{black}{76.0} & \cellcolor[rgb]{0.987,0.822,0.144}\textcolor{black}{91.0} \\
3. & \cellcolor[rgb]{0.418,0.001,0.658}\textcolor{white}{20.0} & \cellcolor[rgb]{0.524,0.025,0.653}\textcolor{white}{27.0} & \cellcolor[rgb]{0.306,0.009,0.634}\textcolor{white}{13.0} & \cellcolor[rgb]{0.910,0.439,0.350}\textcolor{black}{64.0} & \cellcolor[rgb]{0.918,0.454,0.340}\textcolor{black}{65.0} & \cellcolor[rgb]{0.896,0.416,0.366}\textcolor{black}{62.0} & \cellcolor[rgb]{0.953,0.528,0.289}\textcolor{black}{71.0} & \cellcolor[rgb]{0.953,0.528,0.289}\textcolor{black}{71.0} & \cellcolor[rgb]{0.631,0.108,0.608}\textcolor{white}{35.0} & \cellcolor[rgb]{0.976,0.597,0.245}\textcolor{black}{76.0} \\
4. & \cellcolor[rgb]{0.221,0.016,0.602}\textcolor{white}{8.0} & \cellcolor[rgb]{0.200,0.018,0.593}\textcolor{white}{7.0} & \cellcolor[rgb]{0.200,0.018,0.593}\textcolor{white}{7.0} & \cellcolor[rgb]{0.506,0.016,0.656}\textcolor{white}{26.0} & \cellcolor[rgb]{0.949,0.518,0.296}\textcolor{black}{70.0} & \cellcolor[rgb]{0.949,0.518,0.296}\textcolor{black}{70.0} & \cellcolor[rgb]{0.963,0.554,0.272}\textcolor{black}{73.0} & \cellcolor[rgb]{0.973,0.586,0.252}\textcolor{black}{75.0} & \cellcolor[rgb]{0.850,0.347,0.417}\textcolor{black}{56.0} & \cellcolor[rgb]{0.987,0.822,0.144}\textcolor{black}{91.0} \\
5m. & \cellcolor[rgb]{0.106,0.024,0.551}\textcolor{white}{2.0} & \cellcolor[rgb]{0.086,0.026,0.543}\textcolor{white}{1.5} & \cellcolor[rgb]{0.115,0.024,0.555}\textcolor{white}{2.5} & \cellcolor[rgb]{0.241,0.015,0.610}\textcolor{white}{9.0} & \cellcolor[rgb]{0.670,0.143,0.582}\textcolor{white}{38.0} & \cellcolor[rgb]{0.857,0.356,0.410}\textcolor{black}{57.0} & \cellcolor[rgb]{0.890,0.406,0.373}\textcolor{black}{61.0} & \cellcolor[rgb]{0.963,0.554,0.272}\textcolor{black}{73.0} & \cellcolor[rgb]{0.748,0.223,0.517}\textcolor{white}{45.0} & \cellcolor[rgb]{0.994,0.747,0.164}\textcolor{black}{86.0} \\
5f. & \cellcolor[rgb]{0.050,0.030,0.528}\textcolor{white}{0.0} & \cellcolor[rgb]{0.050,0.030,0.528}\textcolor{white}{0.0} & \cellcolor[rgb]{0.050,0.030,0.528}\textcolor{white}{0.0} & \cellcolor[rgb]{0.132,0.022,0.563}\textcolor{white}{3.5} & \cellcolor[rgb]{0.274,0.012,0.623}\textcolor{white}{11.0} & \cellcolor[rgb]{0.772,0.249,0.495}\textcolor{white}{47.5} & \cellcolor[rgb]{0.679,0.152,0.575}\textcolor{white}{39.0} & \cellcolor[rgb]{0.875,0.383,0.390}\textcolor{black}{59.0} & \cellcolor[rgb]{0.728,0.201,0.535}\textcolor{white}{43.0} & \cellcolor[rgb]{0.988,0.652,0.211}\textcolor{black}{80.0} \\
6 & \cellcolor[rgb]{0.050,0.030,0.528}\textcolor{white}{0.0} & \cellcolor[rgb]{0.050,0.030,0.528}\textcolor{white}{0.0} & \cellcolor[rgb]{0.050,0.030,0.528}\textcolor{white}{0.0} & \cellcolor[rgb]{0.050,0.030,0.528}\textcolor{white}{0.0} & \cellcolor[rgb]{0.050,0.030,0.528}\textcolor{white}{0.0} & \cellcolor[rgb]{0.050,0.030,0.528}\textcolor{white}{0.0} & \cellcolor[rgb]{0.776,0.254,0.491}\textcolor{white}{48.0} & \cellcolor[rgb]{0.981,0.613,0.235}\textcolor{black}{77.0} & \cellcolor[rgb]{0.621,0.099,0.614}\textcolor{white}{34.0} & \cellcolor[rgb]{0.989,0.810,0.145}\textcolor{black}{90.0} \\
7m. & \cellcolor[rgb]{0.050,0.030,0.528}\textcolor{white}{0.0} & \cellcolor[rgb]{0.050,0.030,0.528}\textcolor{white}{0.0} & \cellcolor[rgb]{0.050,0.030,0.528}\textcolor{white}{0.0} & \cellcolor[rgb]{0.050,0.030,0.528}\textcolor{white}{0.0} & \cellcolor[rgb]{0.050,0.030,0.528}\textcolor{white}{0.0} & \cellcolor[rgb]{0.050,0.030,0.528}\textcolor{white}{0.0} & \cellcolor[rgb]{0.241,0.015,0.610}\textcolor{white}{9.0} & \cellcolor[rgb]{0.992,0.681,0.195}\textcolor{black}{82.0} & \cellcolor[rgb]{0.656,0.130,0.592}\textcolor{white}{37.0} & \cellcolor[rgb]{0.989,0.810,0.145}\textcolor{black}{90.0} \\
7f. & \cellcolor[rgb]{0.050,0.030,0.528}\textcolor{white}{0.0} & \cellcolor[rgb]{0.050,0.030,0.528}\textcolor{white}{0.0} & \cellcolor[rgb]{0.050,0.030,0.528}\textcolor{white}{0.0} & \cellcolor[rgb]{0.050,0.030,0.528}\textcolor{white}{0.0} & \cellcolor[rgb]{0.050,0.030,0.528}\textcolor{white}{0.0} & \cellcolor[rgb]{0.050,0.030,0.528}\textcolor{white}{0.0} & \cellcolor[rgb]{0.193,0.018,0.590}\textcolor{white}{6.5} & \cellcolor[rgb]{0.931,0.478,0.323}\textcolor{black}{67.0} & \cellcolor[rgb]{0.656,0.130,0.592}\textcolor{white}{37.0} & \cellcolor[rgb]{0.988,0.652,0.211}\textcolor{black}{80.0} \\
\midrule
8a & \cellcolor[rgb]{0.124,0.023,0.559}\textcolor{white}{3.0} & \cellcolor[rgb]{0.350,0.004,0.646}\textcolor{white}{16.0} & \cellcolor[rgb]{0.221,0.016,0.602}\textcolor{white}{8.0} & \cellcolor[rgb]{0.186,0.019,0.587}\textcolor{white}{6.0} & \cellcolor[rgb]{0.149,0.021,0.571}\textcolor{white}{4.0} & \cellcolor[rgb]{0.287,0.011,0.627}\textcolor{white}{12.0} & \cellcolor[rgb]{0.255,0.014,0.615}\textcolor{white}{10.0} & \cellcolor[rgb]{0.255,0.014,0.615}\textcolor{white}{10.0} & \cellcolor[rgb]{0.399,0.001,0.656}\textcolor{white}{19.0} & \cellcolor[rgb]{0.124,0.023,0.559}\textcolor{white}{3.0} \\
8g & \cellcolor[rgb]{0.050,0.030,0.528}\textcolor{white}{0.0} & \cellcolor[rgb]{0.075,0.027,0.538}\textcolor{white}{1.0} & \cellcolor[rgb]{0.106,0.024,0.551}\textcolor{white}{2.0} & \cellcolor[rgb]{0.050,0.030,0.528}\textcolor{white}{0.0} & \cellcolor[rgb]{0.050,0.030,0.528}\textcolor{white}{0.0} & \cellcolor[rgb]{0.050,0.030,0.528}\textcolor{white}{0.0} & \cellcolor[rgb]{0.050,0.030,0.528}\textcolor{white}{0.0} & \cellcolor[rgb]{0.050,0.030,0.528}\textcolor{white}{0.0} & \cellcolor[rgb]{0.050,0.030,0.528}\textcolor{white}{0.0} & \cellcolor[rgb]{0.050,0.030,0.528}\textcolor{white}{0.0} \\
9a & \cellcolor[rgb]{0.050,0.030,0.528}\textcolor{white}{0.0} & \cellcolor[rgb]{0.050,0.030,0.528}\textcolor{white}{0.0} & \cellcolor[rgb]{0.075,0.027,0.538}\textcolor{white}{1.0} & \cellcolor[rgb]{0.050,0.030,0.528}\textcolor{white}{0.0} & \cellcolor[rgb]{0.050,0.030,0.528}\textcolor{white}{0.0} & \cellcolor[rgb]{0.050,0.030,0.528}\textcolor{white}{0.0} & \cellcolor[rgb]{0.075,0.027,0.538}\textcolor{white}{1.0} & \cellcolor[rgb]{0.050,0.030,0.528}\textcolor{white}{0.0} & \cellcolor[rgb]{0.050,0.030,0.528}\textcolor{white}{0.0} & \cellcolor[rgb]{0.050,0.030,0.528}\textcolor{white}{0.0}\\
9g & \cellcolor[rgb]{0.050,0.030,0.528}\textcolor{white}{0.0} & \cellcolor[rgb]{0.050,0.030,0.528}\textcolor{white}{0.0} & \cellcolor[rgb]{0.050,0.030,0.528}\textcolor{white}{0.0} & \cellcolor[rgb]{0.050,0.030,0.528}\textcolor{white}{0.0} & \cellcolor[rgb]{0.050,0.030,0.528}\textcolor{white}{0.0} & \cellcolor[rgb]{0.050,0.030,0.528}\textcolor{white}{0.0} & \cellcolor[rgb]{0.050,0.030,0.528}\textcolor{white}{0.0} & \cellcolor[rgb]{0.050,0.030,0.528}\textcolor{white}{0.0} & \cellcolor[rgb]{0.050,0.030,0.528}\textcolor{white}{0.0} & \cellcolor[rgb]{0.050,0.030,0.528}\textcolor{white}{0.0} \\
\midrule
\textbf{Avg.} & \cellcolor[rgb]{0.248,0.014,0.613}\textcolor{white}{9.5} & \cellcolor[rgb]{0.300,0.010,0.632}\textcolor{white}{12.5} & \cellcolor[rgb]{0.430,0.001,0.659}\textcolor{white}{20.8} & \cellcolor[rgb]{0.495,0.012,0.658}\textcolor{white}{25.1} & \cellcolor[rgb]{0.557,0.047,0.643}\textcolor{white}{29.6} & \cellcolor[rgb]{0.656,0.130,0.592}\textcolor{white}{36.9} & \cellcolor[rgb]{0.706,0.178,0.554}\textcolor{white}{41.3} & \cellcolor[rgb]{0.850,0.347,0.417}\textcolor{black}{56.1} & \cellcolor[rgb]{0.693,0.165,0.565}\textcolor{white}{39.9} & \cellcolor[rgb]{0.905,0.430,0.356}\textcolor{black}{62.9} \\

\bottomrule 
\end{tabularx}
}
\caption{Accuracy scores for various checkpoints in the eng$\rightarrow$art direction, for the \textit{W+G} treatment with CoT.}\label{tab:discussion_results}
\end{table*}

%% file: appendix_ciphers.tex
\section{Encipherment Examples}
\label{sec:encEx}
This appendix provides illustrations of the encipherment methods used in our experiments. The encipherments involve two kinds of technique: substitutions and transpositions. These methods were designed to obscure the original linguistic structure while preserving all morphological and syntactic structures, enabling controlled evaluations of explicit learning on the part of an LLM.

\subsection{Substitutions}
Following are a few random examples of substitutions applied to the clear text segment ``Nouns ending in @-au@ and @-eu@ take the ending @-x@ instead of @-s@ (e.g., @tuyau@ > @tuyaux@ 'pipes', @jeu@ > @jeux@ 'games'), except for the nouns @landau@, @sarrau@, @pneu@, and @bleu@ which form their plural by adding @-s@ (e.g., @pneu@ > @pneus@ 'tires').'':

\small
\begin{tcolorbox}[colback=gray!5!white, colframe=gray!75!black, breakable]
Nouns ending in -yhn and -œsn take the ending -œz instead of -f (e.g., àknîyhn > àknîyhnœz 'pipes', écœsn > écœsnœz 'games'), except for the nouns çyhâiyhn, fyhàzàzyhn, êzâœsn, and tçœsn which form their plural by adding -f (e.g., êzâœsn > êzâœsnf 'tires').
\end{tcolorbox}
\begin{tcolorbox}[colback=gray!5!white, colframe=gray!75!black, breakable]
Nouns ending in -gû and -tiû take the ending -æ instead of -î (e.g., zûagû > zûagûæ 'pipes', wœtiû > wœtiûæ 'games'), except for the nouns tègbeøgû, îghhgû, rybetiû, and otètiû which form their plural by adding -î (e.g., rybetiû > rybetiûî 'tires').
\end{tcolorbox}
\begin{tcolorbox}[colback=gray!5!white, colframe=gray!75!black, breakable]
Nouns ending in -oxl and -œkl take the ending -ig instead of -d (e.g., œglagoxl > œglagoxlig 'pipes', éœkl > éœklig 'games'), except for the nouns êoxœpnoxl, doxototoxl, þœpœkl, and çêœkl which form their plural by adding -d (e.g., þœpœkl > þœpœkld 'tires').
\end{tcolorbox}
\begin{tcolorbox}[colback=gray!5!white, colframe=gray!75!black, breakable]
Nouns ending in -ld and -bod take the ending -ka instead of -pa (e.g., èdnld > èdnldka `pipes', sobod > sobodka `games'), except for the nouns vltomald, palzizild, citobod, and yvbod which form their plural by adding -pa (e.g., citobod > citobodpa `tires').
\end{tcolorbox}
\begin{tcolorbox}[colback=gray!5!white, colframe=gray!75!black, breakable]
Nouns ending in -tém and -pém take the ending -z instead of -yj (e.g., èhémftém > èhémftémz `pipes', kpém > kpémz `games'), except for the nouns otèçustém, yjtnntém, cèçpém, and yqopém which form their plural by adding -yj (e.g., cèçpém > cèçpémyj `tires').
\end{tcolorbox}
\begin{tcolorbox}[colback=gray!5!white, colframe=gray!75!black, breakable]
Nouns ending in -ébt and -ot take the ending -éc instead of -h (e.g., uxtilébt > uxtilébtéc `pipes', abot > abotéc `games'), except for the nouns jébidamébt, hébvvébt, ézidot, and rjot which form their plural by adding -h (e.g., ézidot > ézidoth `tires').
\end{tcolorbox}
\begin{tcolorbox}[colback=gray!5!white, colframe=gray!75!black, breakable]
Nouns ending in -xèf and -evèf take the ending -b instead of -ef (e.g., ucèfezxèf > ucèfezxèfb `pipes', uvevèf > uvevèfb `games'), except for the nouns dxètéxèf, efxititxèf, kètevèf, and hdevèf which form their plural by adding -ef (e.g., kètevèf > kètevèfef `tires').
\end{tcolorbox}
\begin{tcolorbox}[colback=gray!5!white, colframe=gray!75!black, breakable]
Nouns ending in -âwà and -æà take the ending -t instead of -q (e.g., aàokâwà > aàokâwàt `pipes', oræà > oræàt `games'), except for the nouns îâwñûâwà, qâwèzèzâwà, ezñæà, and nîæà which form their plural by adding -q (e.g., ezñæà > ezñæàq `tires').
\end{tcolorbox}
\begin{tcolorbox}[colback=gray!5!white, colframe=gray!75!black, breakable]
Nouns ending in -wom and -qém take the ending -qo instead of -we (e.g., jumiwom > jumiwomqo `pipes', kqém > kqémqo `games'), except for the nouns vuwoléfwom, wewonunuwom, wéléqém, and syvuqém which form their plural by adding -we (e.g., wéléqém > wéléqémwe `tires').
\end{tcolorbox}
\begin{tcolorbox}[colback=gray!5!white, colframe=gray!75!black, breakable]
Nouns ending in -lzu and -ozu take the ending -he instead of -w (e.g., puzuglzu > puzuglzuhe `pipes', xaozu > xaozuhe `games'), except for the nouns xelbèlzu, wléélzu, kbozu, and ruxeozu which form their plural by adding -w (e.g., kbozu > kbozuw `tires').
\end{tcolorbox}
\normalsize

\subsection{Transpositions}
The $s'$ transposition applied alone to the clear text ``The quick, brown fox.'' would give:

\small
\begin{tcolorbox}[colback=gray!5!white, colframe=gray!75!black, breakable]
xof nworb, kciuq eht.
\end{tcolorbox}
\normalsize
Our implementation of the transposition affects punctuations in smart ways but does not preserve case.

While the final version of our experiments exclusively relied on transposition $s'$--as it produced a more manageable conlang type--we also explored another transposition, $w'$:
\begin{align*}
w'[i] &= w[m - i + 1] \quad \text{for } 1 \leq i \leq m,
\end{align*}
with \(w\) a word, \(m\) the number of characters in \(w\).

The $w'$ transposition applied alone to the clear text ``The quick, brown fox.'' would give:
\small
\begin{tcolorbox}[colback=gray!5!white, colframe=gray!75!black, breakable]
eht kciuq, nworb xof.
\end{tcolorbox}
\normalsize

\subsection{Substitutions + Transposition}
Following are a few random examples of substitutions combined with the $s'$ transposition applied to the clear text segment ``Nouns beginning with @-au@ and @-eu@ take the prefix @-x@ instead of @-s@ (e.g., @tuyau@ > @tuyaux@ `pipes', @jeu@ > @jeux@ `games'), except for the nouns @landau@, @sarrau@, @pneu@, and @bleu@ which form their plural by prefixing the morpheme @-s@ (e.g., @pneu@ > @pneus@ `tires').'':

\small

\begin{tcolorbox}[colback=gray!5!white, colframe=gray!75!black, breakable]
Nouns beginning with gi- and géz- take the prefix az- instead of d- (e.g., giecgk > azgiecgk `pipes', gézav > azgézav `games'), except for the nouns gievrix, giyyid, gézrab, and gézxéc which form their plural by prefixing the morpheme d- (e.g., gézrab > dgézrab `tires').
\end{tcolorbox}
\begin{tcolorbox}[colback=gray!5!white, colframe=gray!75!black, breakable]
Nouns beginning with cuà- and cuþ- take the prefix fû- instead of â- (e.g., cuàgculu > fûcuàgculu `pipes', cuþð > fûcuþð 'games'), except for the nouns cuàahàq, cuàîîàâ, cuþhjy, and cuþqd which form their plural by prefixing the morpheme â- (e.g., cuþhjy > âcuþhjy `tires').
\end{tcolorbox}
\begin{tcolorbox}[colback=gray!5!white, colframe=gray!75!black, breakable]
Nouns beginning with âlœ- and âlô- take the prefix b- instead of e- (e.g., âlœîjâlâm > bâlœîjâlâm `pipes', âlôon > bâlôon 'games'), except for the nouns âlœulanœk, âlœyyœe, âlôanoz, and âlôkê which form their plural by prefixing the morpheme e- (e.g., âlôanoz > eâlôanoz 'tires').
\end{tcolorbox}
\begin{tcolorbox}[colback=gray!5!white, colframe=gray!75!black, breakable]
Nouns beginning with eoj- and eèb- take the prefix èw- instead of s- (e.g., eojiçep > èweojiçep `pipes', eèbg > èweèbg `games'), except for the nouns eojàtubojuz, eojèhèhojs, eèbubon, and eèbuzx which form their plural by prefixing the morpheme s- (e.g., eèbubon > seèbubon `tires').
\end{tcolorbox}
\begin{tcolorbox}[colback=gray!5!white, colframe=gray!75!black, breakable]
Nouns beginning with vqa- and vn- take the prefix o- instead of c- (e.g., vqatyvzy > ovqatyvzy `pipes', vnza > ovnza `games'), except for the nouns vqaubàqame, vqazezeqac, vnbài, and vnmek which form their plural by prefixing the morpheme c- (e.g., vnbài > cvnbài `tires').
\end{tcolorbox}
\begin{tcolorbox}[colback=gray!5!white, colframe=gray!75!black, breakable]
Nouns beginning with èda- and èxu- take the prefix he- instead of w- (e.g., èdatèdà > heèdatèdà 'pipes', èxuhà > heèxuhà 'games'), except for the nouns èdabardade, èdappdaw, èxuro, and èxudem which form their plural by prefixing the morpheme w- (e.g., èxuro > wèxuro `tires').
\end{tcolorbox}
\begin{tcolorbox}[colback=gray!5!white, colframe=gray!75!black, breakable]
Nouns beginning with jte- and jb- take the prefix mi- instead of xe- (e.g., jtenajma > mijtenajma `pipes', jbl > mijbl 'games'), except for the nouns jtepçatehé, jtesstexe, jbçaz, and jbhéfè which form their plural by prefixing the morpheme xe- (e.g., jbçaz > xejbçaz `tires').
\end{tcolorbox}
\begin{tcolorbox}[colback=gray!5!white, colframe=gray!75!black, breakable]
Nouns beginning with éax- and éd- take the prefix on- instead of p- (e.g., éaxjéc > onéaxjéc `pipes', édoq > onédoq `games'), except for the nouns éaxemaxyg, éaxararaxp, édmç, and édygh which form their plural by prefixing the morpheme p- (e.g., édmç > pédmç `tires').
\end{tcolorbox}
\begin{tcolorbox}[colback=gray!5!white, colframe=gray!75!black, breakable]
Nouns beginning with èvà- and èl- take the prefix jo- instead of cy- (e.g., èvàvoèç > joèvàvoèç `pipes', èls > joèls `games'), except for the nouns èvànuvàbi, èvàririvàcy, èlubà, and èlbiqi which form their plural by prefixing the morpheme cy- (e.g., èlubà > cyèlubà `tires').
\end{tcolorbox}
\begin{tcolorbox}[colback=gray!5!white, colframe=gray!75!black, breakable]
Nouns beginning with uþ- and uet- take the prefix û- instead of w- (e.g., uþytun > ûuþytun `pipes', uetj > ûuetj `games'), except for the nouns uþðfþs, uþbbþw, uetfyh, and uetsep which form their plural by prefixing the morpheme w- (e.g., uetfyh > wuetfyh `tires').
\end{tcolorbox}
\normalsize

%% file: appendix_grammar_source.tex
\section{Grammar Book Sources}
\label{sec:gs_appendix}

List of URLs of Wikipedia pages used to create the grammar book templates:

\begin{itemize}
    \item \url{https://en.wikipedia.org/wiki/French_grammar},
    \item \url{https://en.wikipedia.org/wiki/French_verbs},
    \item \url{https://en.wikipedia.org/wiki/French_conjugation},
    \item \url{https://en.wikipedia.org/wiki/French_verb_morphology},
    \item \url{https://en.wikipedia.org/wiki/French_articles_and_determiners},
    \item \url{https://en.wikipedia.org/wiki/French_adverbs},
    \item \url{https://en.wikipedia.org/wiki/French_pronouns},
    \item \url{https://en.wikipedia.org/wiki/Latin_grammar},
    \item \url{https://en.wikipedia.org/wiki/Latin_declension},
    \item \url{https://en.wikipedia.org/wiki/Latin_syntax},
    \item \url{https://en.wikipedia.org/wiki/Latin_tenses},
    \item \url{https://en.wikipedia.org/wiki/Latin_conjugation},
    \item \url{https://en.wikipedia.org/wiki/Latin_word_order}.
\end{itemize}

(Accessed: September to October 2024)

%% file: appendix_grammar_template.tex
\section{Grammar Book Variations}
\label{sec:gtempl_appendix}

To ensure that the grammar book excerpts shown at test time would be novel to the fine-tuned LLM, we created an average of five paraphrases for each of the 90+ paragraphs comprised in these excerpts. In addition to paraphrasing, other strategies were employed to diversify the grammar books used in our fine-tuning sets:

\begin{itemize} \item Example lists were converted into shufflable lists, with their order randomized for each instance.
\item The order of the main sections in the grammar excerpts (e.g., NOUNS, ADJECTIVES, PRONOUNS, VERBS) was randomized, as was the order of subsections within each section (e.g., Gender, Number, and Case under NOUNS). Additionally, most paragraphs within subsections (except the first) were generally presented in random order.
\item Section markers and symbols used in morphological tables (e.g., -, \#, \textasciitilde, §) were randomly selected from a predefined list. The presentation of morphological tables also varied, using randomly selected templates. Such formatting variations have been shown to markedly impact LLM performance \citep{voronov_mind_2024, sclar_quantifying_2024}. \end{itemize}

The resulting documents exhibit significant variation in form, particularly in the most complex partition of the fine-tuning dataset, where longer grammar book excerpts allow for greater diversity. Below are two examples of the NOUN section from the grammar excerpts, along with its test set version, all presented in their unenciphered, clear-text template form.

\small

\begin{tcolorbox}[colback=gray!5!white, colframe=gray!75!black, breakable, title=First Example of Grammar Excerpt Variations]
§§§§§				
NOUNS
§§§§§\\

* Gender\\
All nouns in @French@ are designated with a grammatical gender: masculine or feminine.\\

* Case\\
In @French@, nouns are not inflected for other grammatical distinctions. (Case and person inflections apply only to personal pronouns.)\\

* Number\\
@French@ nouns, like those in English, inflect to indicate number.\\

The plural is most often derived by attaching the morpheme @-s@ to the singular, as in @maison@ to @maisons@ (`houses').\\

Nouns with singular endings of @-s@, @-x@, or @-z@ do not undergo changes when forming the plural (e.g., @la croix@ `the cross' > @les croix@ `the crosses').\\

The plural of most nouns ending in @-ail@ is formed by changing to @-aux@. Examples of this pattern include @émail@, @corail@, @travail@, @soupirail@, @vitrail@, and @bail@ (e.g., @vitrail@ becomes @vitraux@ `stained glass windows'). A few exceptions, however, add @-s@ to form their plural: @attirail@, @rail@, @détail@, @chandail@, @gouvernail@, and @portail@ (e.g., @un rail@ > @des rails@ `rails').\\

For nouns ending in @-au@ and @-eu@, the plural is typically formed by adding @-x@ rather than @-s@ (e.g., @tuyau@ > @tuyaux@ `pipes', @jeu@ > @jeux@ `games'). Exceptions to this rule include @landau@, @sarrau@, @bleu@ and @pneu@, which use @-s@ to create their plural forms (e.g., @pneu@ > @pneus@ `tires').\\

For seven nouns ending in @-ou@, the plural is formed with @-x@: @pou@, @joujou@, @hibou@, @bijou@, @genou@, @chou@, @caillou@ (e.g., @genou@ becomes @genoux@ `knees'). Others, like @bisou@ or @trou@, take the regular plural suffix @-s@ (e.g., @trou@ becomes @trous@ `holes').\\

Nouns ending in @-al@\\
Typically, nouns ending in @-al@ pluralize to @-aux@ (e.g., @journal@ > @journaux@ 'newspapers'). But there are eight exceptions, which add @-s@ for the plural: @récital@, @cal@, @régal@, @serval@, @carnaval@, @chacal@, @bal@, @festival@ (e.g., @le festival@ > @les festivals@).\\

\end{tcolorbox}

\begin{tcolorbox}[colback=gray!5!white, colframe=gray!75!black, breakable, title=Second Example of Grammar Excerpt Variations]
:::::
NOUNS
:::::\\

§ Number\\
As in English, @French@ nouns inflect for number.\\

The plural is generally formed from the singular by appending the morpheme @-s@ (e.g., @maison@ becomes @maisons@, meaning `houses').\\

(Nouns ending in @-ail@)\\
Most nouns ending in @-ail@ form their plural by changing to @-aux@. Examples include @émail@, @bail@, @soupirail@, @corail@, @vitrail@, and @travail@ (e.g., @vitrail@ > @vitraux@ `stained glass windows'). However, some exceptions take @-s@ instead: @gouvernail@, @détail@, @portail@, @rail@, @attirail@, and @chandail@ (e.g., @un rail@ > @des rails@ `rails').\\

There are seven exceptions among nouns ending in @-ou@, which form their plural with @-x@: @chou@, @genou@, @joujou@, @pou@, @bijou@, @caillou@, @hibou@ (e.g., @genou@ > @genoux@ `knees'). Nouns such as @bisou@ and @trou@ form their plural regularly by adding @-s@ (e.g., @trou@ > @trous@ `holes').\\

Nouns ending in @-au@ and @-eu@ form their plural with the ending @-x@ instead of @-s@ (e.g., @jeu@ becomes @jeux@ `games', @tuyau@ becomes @tuyaux@ `pipes'). However, exceptions include the nouns @sarrau@, @landau@, @pneu@, and @bleu@, which take @-s@ to form the plural (e.g., @pneu@ becomes @pneus@ `tires').\\

Nouns ending in @-al@\\
For nouns ending in @-al@, the plural is usually @-aux@ (e.g., @journal@ becomes @journaux@ `newspapers'). However, these eight nouns are exceptions and take @-s@: @bal@, @festival@, @serval@, @régal@, @cal@, @récital@, @carnaval@, @chacal@ (e.g., @le festival@ > @les festivals@).\\

Nouns with singular endings of @-s@, @-x@, or @-z@ do not undergo changes when forming the plural (e.g., @la croix@ `the cross' > @les croix@ 'the crosses').\\

§ Case\\
Nouns in @French@ retain a single form regardless of grammatical categories. (Personal pronouns, on the other hand, vary by case and person.)\\

§ Gender\\
All nouns in @French@ are designated with a grammatical gender: masculine or feminine.\\

\end{tcolorbox}

\begin{tcolorbox}[colback=gray!5!white, colframe=gray!75!black, breakable, title=Equivalent Test Set Grammar Excerpt]
-----
NOUNS-----\\

* Gender\\
Every @French@ noun has a grammatical gender, either masculine or feminine.\\

* Number\\
As in English, @French@ nouns inflect for number.\\

The plural is usually formed from the singular by adding the morpheme @-s@ (e.g., @maison@ > @maisons@ `houses'). \\

Nouns ending in @-au@ and @-eu@ take the ending @-x@ instead of @-s@ (e.g., @tuyau@ > @tuyaux@ `pipes', @jeu@ > @jeux@ `games'), except for the nouns @landau@, @sarrau@, @pneu@, and @bleu@ which form their plural by adding @-s@ (e.g., @pneu@ > @pneus@ `tires').\\

Seven nouns ending in @-ou@ form their plural with @-x@. These are @genou@, @caillou@, @hibou@, @bijou@, @pou@, @chou@, and @joujou@ (e.g., @genou@ > @genoux@ `knees'). All other nouns ending in @-ou@, such as @bisou@ or @trou@, follow the regular plural pattern and take @-s@ (e.g., @trou@ > @trous@ `holes').\\

Nouns that end in @-s@, @-x@ or @-z@ in the singular do not change in the plural (e.g., @la croix@ `the cross' > @les croix@ `the crosses').\\

Nouns that end in @-al@ typically change to @-aux@ in the plural (e.g., @journal@ > @journaux@ `newspapers'). However, eight nouns ending in @-al@ are exceptions to this rule and form their plural by adding @-s@ instead. These are @bal@, @cal@, @carnaval@, @chacal@, @festival@, @récital@, @serval@ and @régal@ (e.g., @le festival@ > @les festivals@).\\

Similarly, most nouns ending in @-ail@ change to @-aux@ in the plural. Examples of this pattern include @bail@, @corail@, @émail@, @soupirail@, @travail@, and @vitrail@ (e.g., @vitrail@ > @vitraux@ `stained glass windows'). However, some @-ail@ nouns do not follow this pattern and form their plural with @-s@, notably @rail@, @attirail@, @chandail@, @détail@, @gouvernail@ and @portail@ (e.g., @un rail@ > @des rails@ `rails').\\

* Case\\
Nouns in @French@ are not inflected for any other grammatical categories. (However, personal pronouns are inflected for case and person.)\\
\end{tcolorbox}
\normalsize

%% file: dictionaryBD.tex
\section{Dictionary Lemma Counts and Categorization}
\label{app:dictionary_counts}

This appendix provides a detailed breakdown of the lemma counts in the French-English and Latin-English dictionary templates crafted for our experiments. As described in Section~3.1, 
lemmas were chosen based on linguistic phenomena in our grammar books and divided into three categories according to their presence and role within these grammar book excerpts:

\begin{itemize}
    \item \textbf{Category A:} Lemmas that appear within illustrative bitexts in the grammar book. This allows the LLM to potentially rely on direct copying or very close adaptation for translation.
    \item \textbf{Category B:} Lemmas that appear \textit{only} within metalinguistic explanations in the grammar book (i.e.,~not in any illustrative bitexts). These point to rules governing a linguistic phenomenon but do not provide a direct example of the expected translated form. For the French--English and Latin--English dictionaries, this category was primarily populated with nouns as per our design focus for this category.
    \item \textbf{Category C:} Lemmas that are entirely absent from the grammar book. For these, the LLM must infer the applicable rule from the general grammatical explanations to generate or analyze the targeted phenomenon in translation.
\end{itemize}

The distribution of lemmas across these categories for the French--English dictionary is detailed in Table~\ref{tab:french_dict_dist_app} and for the Latin--English dictionary in Table~\ref{tab:latin_dict_dist_app}.

\begin{table}[h!]
\centering\small
\resizebox{\columnwidth}{!}{%
\begin{tabular}{lrrrr}
\toprule
\textbf{Part of Speech} & \textbf{Cat. A} & \textbf{Cat. B} & \textbf{Cat. C} & \textbf{Total} \\
\midrule
Nouns                   & 11             & 26             & 121            & 158                 \\
Adjectives              & 10             & 0              & 66             & 76                  \\
Verbs                   & 2              & 0              & 4              & 6                   \\
\midrule
\textbf{Grand Total}    & \textbf{23}    & \textbf{26}    & \textbf{191}   & \textbf{240}        \\
\bottomrule
\end{tabular}%
}
\caption{French--English Dictionary Lemma Distribution}
\label{tab:french_dict_dist_app}
\end{table}

\begin{table}[h!]
\centering\small
\resizebox{\columnwidth}{!}{%
\begin{tabular}{lrrrr}
\toprule
\textbf{Part of Speech} & \textbf{Cat. A} & \textbf{Cat. B} & \textbf{Cat. C} & \textbf{Total} \\
\midrule
Nouns                   & 8              & 4              & 7              & 19                 \\
Adjectives              & 2              & 0              & 7              & 9                  \\
Verbs                   & 4              & 0              & 6              & 10                 \\
\midrule
\textbf{Grand Total}    & \textbf{14}    & \textbf{4}     & \textbf{20}    & \textbf{38}        \\
\bottomrule
\end{tabular}%
}
\caption{Latin--English Dictionary Lemma Distribution}
\label{tab:latin_dict_dist_app}
\end{table}

%% file: appendix_eng_art_prompts_1.tex
\section{Prompt and CoT Demonstration Examples from our Datasets}
\label{sec:pr_appendix}
The following are examples of prompts and completion demonstrations taken from our datasets.

\subsection{ENG$\rightarrow$ART: Example Instance from Partition 1 (Plural of French Common Nouns) for `True' French}

The following are a prompt template (i.e., unenciphered)  for an instance from Partition 1 of our fine-tuning dataset addressing the pluralization of French common nouns, its clear-text CoT demonstration completion, and conlang versions of these two texts (i.e., enciphered). `True' French serves as the conlang matrix for these instances, meaning the encipherment was performed exclusively through substitutions, without transpositions.

\small

\begin{tcolorbox}[colback=gray!5!white, colframe=gray!75!black, breakable, title=Clear Text Prompt]
@French@ is a recently devised conlang. You are to translate the following English segment into @French@ with the help of a few dictionary entries and excerpts from a grammar book.\\\\Here is the text segment you must translate:\\the confessions\\\\Here are a few dictionary entries that may be of use to you; note that each entry follows the format: English lemma : @French@ equivalent (grammatical gender and/or part of speech).\\confession : @aveu@ (masc. n.)\\\\And here are relevant excerpts from a grammar book: \\\\Beginning of @French@ Grammar Book Excerpts\\\\ARTICLES AND DETERMINERS\\\\In @French@, articles and determiners are almost always required with common nouns, much more so than in English. These words agree in gender (masculine or feminine) and number (singular or plural) with the noun they modify, though most have a single plural form for both genders.\\\\Although articles are a subclass of determiners, they are usually treated as a distinct category, which is the approach taken here as well.\\\\§ Articles\\@French@ articles include three categories: the definite article (parallel to English 'the'), the indefinite article (equivalent to 'a/an'), and the partitive article, used similarly to 'some' in English.\\\\§§ Definite article\\In @French@, the definite article, much like the English 'the', is used to refer to a specific noun. Unlike English, the @French@ article varies depending on the noun's gender (masculine or feminine) and number (singular or plural).\\\\The definite article always precedes its noun.\\\\For singular nouns:\\\\The definite article for masculine singular nouns is @le@. For instance, "@le chat@" (the cat) uses @le@ to show that @chat@ (cat) is singular and masculine.\\\\For feminine singular nouns, the definite article is @la@. An example is "@la maison@" (the house), where @la@ identifies @maison@ (house) as singular and feminine.\\\\For plural nouns, the article is always @les@, regardless of gender:\\\\- Examples include "@les chats@" (the cats) and "@les maisons@" (the houses).\\\\To sum up, masculine singular nouns use @le@, feminine singular nouns use @la@, and all plural nouns use @les@, reflecting gender and number in @French@.\\\\In front of @i@, @è@, @u@, @e@, some occurrences of @h@, @é@, @ê@, @a@, @y@, and @o@, the masculine singular article @le@ is elided to @l'@.\\\\NOUNS\\\\§ Gender\\All nouns in @French@ are designated with a grammatical gender: masculine or feminine.\\\\§ Number\\In @French@, nouns inflect for number in a manner similar to English.\\\\The singular is made plural by the addition of @-s@ (e.g., the singular @maison@ becomes @maisons@, meaning 'houses').\\\\Seven nouns with the ending @-ou@ use @-x@ for their plural form: @genou@, @chou@, @caillou@, @bijou@, @joujou@, @pou@, @hibou@ (e.g., @genou@ becomes @genoux@ 'knees'). Regular nouns like @bisou@ or @trou@ follow the typical pattern and add @-s@ (e.g., @trou@ becomes @trous@ 'holes').\\\\The plural of most nouns ending in @-ail@ is formed by changing to @-aux@. Examples of this pattern include @travail@, @corail@, @émail@, @bail@, @soupirail@, and @vitrail@ (e.g., @vitrail@ becomes @vitraux@ 'stained glass windows'). A few exceptions, however, add @-s@ to form their plural: @chandail@, @attirail@, @rail@, @gouvernail@, @portail@, and @détail@ (e.g., @un rail@ > @des rails@ 'rails').\\\\Nouns with singular endings of @-s@, @-x@, or @-z@ do not undergo changes when forming the plural (e.g., @la croix@ 'the cross' > @les croix@ 'the crosses').\\\\Nouns with endings @-au@ and @-eu@ generally form their plural with @-x@ instead of @-s@ (e.g., @tuyau@ changes to @tuyaux@ 'pipes', @jeu@ to @jeux@ 'games'). Exceptions to this are @landau@, @sarrau@, @pneu@, and @bleu@, which form their plural by adding @-s@ (e.g., @pneu@ becomes @pneus@ 'tires').\\\\Nouns that end in @-al@ generally form their plural by changing to @-aux@ (e.g., @journal@ > @journaux@ 'newspapers'). However, the following eight exceptions form their plural by adding @-s@: @récital@, @bal@, @régal@, @carnaval@, @chacal@, @cal@, @festival@ and @serval@ (e.g., @le festival@ > @les festivals@).\\\\§ Case\\@French@ nouns are not modified for additional grammatical categories. (Although personal pronouns do reflect case and person.)\\\\End of @French@ Grammar Book Excerpts\\\\\\A reminder that the English sentence you must translate into @French@ is:\\the confessions\\\\You may explain your chain of thoughts prior to producing the required translation. IMPORTANT: Do write your translation between tags in the following manner: <translation>your translation here</translation>.
\end{tcolorbox}

\begin{tcolorbox}[colback=gray!5!white, colframe=gray!75!black, breakable, title=Clear Text CoT Demonstration]
To translate the phrase "the confessions" into @French@, we need to follow a few steps based on the provided dictionary entries and grammar rules.\\\\1. **Identify the noun and its properties**:\\   - The English word "confessions" appears to correspond to the @French@ noun "@aveu@," which is a masculine noun according to the dictionary entry.\\\\2. **Determine the Number**:\\   - "confessions" is the plural form of "confession." According to the grammar rules, the plural of "@aveu@" is formed by adding the morpheme @-x@. Therefore, "@aveu@" should become "@aveux@" in the plural.\\\\3. **Determine the article to use**:\\   - The definite article in @French@ for plural nouns (regardless of gender) is "@les@."\\\\4. **Combine the article with the plural noun**:\\   - According to the grammar book, we should place the definite article "@les@" before the plural noun "@aveux@."\\\\Putting it all together, the translation for the confessions in @French@ is:\\\\<translation>@les aveux@</translation>
\end{tcolorbox}

\begin{tcolorbox}[colback=gray!5!white, colframe=gray!75!black, breakable, title=Ciphered Conlang Prompt]

Biruméolébu is a recently devised conlang. You are to translate the following English segment into Biruméolébu with the help of a few dictionary entries and excerpts from a grammar book.\\ \\ Here is the text segment you must translate:\\ the confessions\\ \\ Here are a few dictionary entries that may be of use to you; note that each entry follows the format: English lemma : Biruméolébu equivalent (grammatical gender and/or part of speech).\\ confession : dusémép (masc. n.)\\ \\ And here are relevant excerpts from a grammar book: \\ \\ Beginning of Biruméolébu Grammar Book Excerpts\\ \\ ARTICLES AND DETERMINERS\\ \\ In Biruméolébu, articles and determiners are almost always required with common nouns, much more so than in English. These words agree in gender (masculine or feminine) and number (singular or plural) with the noun they modify, though most have a single plural form for both genders.\\ \\ Although articles are a subclass of determiners, they are usually treated as a distinct category, which is the approach taken here as well.\\ \\ § Articles\\ Biruméolébu articles include three categories: the definite article (parallel to English 'the'), the indefinite article (equivalent to 'a/an'), and the partitive article, used similarly to 'some' in English.\\ \\ §§ Definite article\\ In Biruméolébu, the definite article, much like the English 'the', is used to refer to a specific noun. Unlike English, the Biruméolébu article varies depending on the noun's gender (masculine or feminine) and number (singular or plural).\\ \\ The definite article always precedes its noun.\\ \\ For singular nouns:\\ \\ The definite article for masculine singular nouns is nmé. For instance, "nmé lébudumà" (the cat) uses nmé to show that lébudumà (cat) is singular and masculine.\\ \\ For feminine singular nouns, the definite article is ndu. An example is "ndu zudurèvko" (the house), where ndu identifies zudurèvko (house) as singular and feminine.\\ \\ For plural nouns, the article is always nmév, regardless of gender:\\ \\ - Examples include "nmév lébudumàv" (the cats) and "nmév zudurèvkov" (the houses).\\ \\ To sum up, masculine singular nouns use nmé, feminine singular nouns use ndu, and all plural nouns use nmév, reflecting gender and number in Biruméolébu.\\ \\ In front of rè, sè, p, mé, some occurrences of bu, h, ê, du, w, and k, the masculine singular article nmé is elided to n'.\\ \\NOUNS\\ \\ § Gender\\ All nouns in Biruméolébu are designated with a grammatical gender: masculine or feminine.\\ \\ § Number\\ In Biruméolébu, nouns inflect for number in a manner similar to English.\\ \\ The singular is made plural by the addition of -v (e.g., the singular zudurèvko becomes zudurèvkov, meaning 'houses').\\ \\ Seven nouns with the ending -kp use -ri for their plural form: suméokp, lébukp, lédurènnkp, murèfkp, fkpfkp, xkp, burèmukp (e.g., suméokp becomes suméokpri 'knees'). Regular nouns like murèvkp or màrukp follow the typical pattern and add -v (e.g., màrukp becomes màrukpv 'holes').\\ \\ The plural of most nouns ending in -durèn is formed by changing to -dupri. Examples of this pattern include màrudusédurèn, lékrudurèn, hzudurèn, mudurèn, vkpxrèrudurèn, and sérèmàrudurèn (e.g., sérèmàrudurèn becomes sérèmàrudupri 'stained glass windows'). A few exceptions, however, add -v to form their plural: lébuduojdurèn, dumàmàrèrudurèn, rudurèn, sukpséméruodurèn, xkrumàdurèn, and jhmàdurèn (e.g., po rudurèn > jmév rudurènv 'rails').\\ \\ Nouns with singular endings of -v, -ri, or -c do not undergo changes when forming the plural (e.g., ndu lérukrèri 'the cross' > nmév lérukrèri 'the crosses').\\ \\ Nouns with endings -dup and -mép generally form their plural with -ri instead of -v (e.g., màpwdup changes to màpwdupri 'pipes', fmép to fmépri 'games'). Exceptions to this are nduojdup, vdururudup, xomép, and munmép, which form their plural by adding -v (e.g., xomép becomes xomépv 'tires').\\ \\ Nouns that end in -dun generally form their plural by changing to -dupri (e.g., fkpruodun > fkpruodupri 'newspapers'). However, the following eight exceptions form their plural by adding -v: ruhlérèmàdun, mudun, ruhsudun, léduruodusédun, lébudulédun, lédun, ymévmàrèsédun and vmérusédun (e.g., nmé ymévmàrèsédun > nmév ymévmàrèsédunv).\\ \\ § Case\\ Biruméolébu nouns are not modified for additional grammatical categories. (Although personal pronouns do reflect case and person.)\\ \\ End of Biruméolébu Grammar Book Excerpts\\ \\ \\ A reminder that the English sentence you must translate into Biruméolébu is:\\ the confessions\\ \\ You may explain your chain of thoughts prior to producing the required translation. IMPORTANT: Do write your translation between tags in the following manner: <translation>your translation here</translation>.

\end{tcolorbox}

\begin{tcolorbox}[colback=gray!5!white, colframe=gray!75!black, breakable, title=Ciphered CoT Demonstration]
To translate the phrase "the confessions" into Biruméolébu, we need to follow a few steps based on the provided dictionary entries and grammar rules.\\\\1. **Identify the noun and its properties**:\\   - The English word "confessions" appears to correspond to the Biruméolébu noun "dusémép," which is a masculine noun according to the dictionary entry.\\\\2. **Determine the Number**:\\   - "confessions" is the plural form of "confession." According to the grammar rules, the plural of "dusémép" is formed by adding the morpheme -ri. Therefore, "dusémép" should become "dusémépri" in the plural.\\\\3. **Determine the article to use**:\\   - The definite article in Biruméolébu for plural nouns (regardless of gender) is "nmév."\\\\4. **Combine the article with the plural noun**:\\   - According to the grammar book, we should place the definite article "nmév" before the plural noun "dusémépri."\\\\Putting it all together, the translation for the confessions in Biruméolébu is:\\\\<translation>nmév dusémépri</translation>
\end{tcolorbox}

\normalsize

%% file: appendix_eng_art_prompts_2.tex
\subsection{ENG$\rightarrow$ART: Example Instance from Partition 1 (Plural of French Common Nouns) for `Reversed' French (Enciphered with Additional Transpositions)}

The following are a prompt template (i.e., unenciphered) for an instance from Partition 1 of our fine-tuning dataset addressing the pluralization of French common nouns, its clear-text CoT demonstration completion, and conlang versions of these two texts (i.e., randomly enciphered). `Reversed' French serves as the conlang matrix for these instances, meaning the encipherment involved both substitutions and the additional $s'$ transposition. Consequently, the grammar excerpts describe a language significantly different from `True' French.

\small
\begin{tcolorbox}[colback=gray!5!white, colframe=gray!75!black, breakable, title=Clear Text Prompt]
@French@ is a recently devised conlang. You are to translate the following English segment into @French@ with the help of a few dictionary entries and excerpts from a grammar book.\\ \\ Here is the text segment you must translate:\\ the metals\\ \\ Here are a few dictionary entries that may be of use to you; note that each entry follows the format: English lemma : @French@ equivalent (grammatical gender and/or part of speech).\\ metal : @métal@ (masc. n.)\\ \\ And here are relevant excerpts from a grammar book: \\ \\ Beginning of @French@ Grammar Book Excerpts\\ \\ARTICLES AND DETERMINERS\\ \\ In @French@, determiners, including articles, are required with nearly every common noun, unlike in English. They must match the noun in gender (masculine or feminine) and number (singular or plural), but most have a unified plural form for both genders.\\ \\ While articles are formally a subset of determiners, they are conventionally treated as separate, and this treatment is adopted here.\\ \\+ Articles\\ @French@ uses three types of articles: the definite article, which often corresponds to English 'the'; the indefinite article, similar to English 'a/an'; and the partitive article, roughly equivalent to 'some' in English.\\ \\ ++ Definite article\\ In @French@, definite articles, like the English 'the', indicate specific nouns. However, they adapt to reflect the gender (masculine or feminine) and number (singular or plural) of the noun they modify.\\ \\ The definite article always comes after the noun it determines.\\ \\ For singular nouns:\\ \\ The article @le@ is used with masculine singular nouns. For instance, "@le chat@" (the cat) demonstrates that the noun @chat@ (cat) is singular and masculine.\\ \\ For feminine singular nouns, the article @la@ applies. An example is "@la maison@" (the house), where @la@ shows that the noun @maison@ (house) is singular and feminine.\\ \\ For plural nouns, regardless of their gender:\\ \\ - The article used is @les@ for masculine and feminine plural nouns alike. (e.g., "@les chats@" means the cats; "@les maisons@" means the houses.)\\ \\ To summarize, @le@ corresponds to masculine singular nouns, @la@ to feminine singular nouns, and @les@ to all plural nouns, marking both grammatical gender and number in @French@.\\ \\ Whenever the singular masculine article @le@ is preceded by @é@, @è@, @e@, @ê@, some occurrences of @h@, @i@, @a@, @o@, @u@, or @y@, it changes to @l'@.\\ \\NOUNS\\ \\ + Case\\ @French@ nouns are not modified for additional grammatical categories. (Although personal pronouns do reflect case and person.)\\ \\ + Number\\ Unlike English, @French@ plural forms are created using prefixation. \\ \\ This involves adding the morpheme @-s@ as a prefix to the singular noun (for example, @maison@ > @maisons@ 'houses').\\ \\ There are seven nouns beginning with @-ou@ that take the prefix @-x@ in their plural form. These nouns are: @chou@, @pou@, @hibou@, @joujou@, @caillou@, @bijou@, and @genou@ (e.g., @genou@ --> @genoux@ 'knees'). All other nouns starting with @-ou@, such as @bisou@ and @trou@, follow the regular plural pattern by prefixing @-s@ (e.g., @trou@ --> @trous@ 'holes').\\ \\ Nouns starting with @-al@ generally pluralize to @-aux@ (e.g., @journal@ -> @journaux@ 'newspapers'). A few exceptions, however, take @-s@ as their plural marker: @récital@, @festival@, @régal@, @bal@, @serval@, @carnaval@, @chacal@, and @cal@ (e.g., @le festival@ -> @les festivals@).\\ \\ Singular nouns starting with @-z@, @-x@, or @-s@ retain their form in the plural (e.g., @la croix@ 'the cross' > @les croix@ 'the crosses').\\ \\ The plural of nouns beginning with @-au@ and @-eu@ is typically formed by prefixing @-x@ instead of @-s@ (e.g., @tuyau@ > @tuyaux@ 'pipes', @jeu@ > @jeux@ 'games'). However, the nouns @landau@, @sarrau@, @pneu@, and @bleu@ are exceptions and prefix @-s@ (e.g., @pneu@ > @pneus@ 'tires').\\ \\ Nouns that start with @-ail@ generally form their plural with @-aux@. Examples are: @bail@, @émail@, @corail@, @soupirail@, @travail@, and @vitrail@ (e.g., @un vitrail@ becomes @des vitraux@ 'stained-glass windows'). However, some exceptions take @-s@ as their plural marker: @gouvernail@, @portail@, @attirail@, @détail@, @rail@, and @chandail@ (e.g., @le rail@ becomes @les rails@ 'the rails').\\ \\ + Gender\\ @French@ nouns are always associated with a grammatical gender, either masculine or feminine.\\ \\ End of @French@ Grammar Book Excerpts\\ \\ \\ A reminder that the English sentence you must translate into @French@ is:\\ the metals\\ \\ You may explain your chain of thoughts prior to producing the required translation. IMPORTANT: Do write your translation between tags in the following manner: <translation>your translation here</translation>.

\end{tcolorbox}

\begin{tcolorbox}[colback=gray!5!white, colframe=gray!75!black, breakable, title=Clear Text CoT Demonstration]
To translate the phrase "the metals" into @French@, we need to follow a few steps based on the provided dictionary entries and grammar rules.\\ \\ 1. **Identify the noun and its properties**:\\    - The English word "metals" appears to correspond to the @French@ noun "@métal@," which is a masculine noun according to the dictionary entry.\\ \\ 2. **Determine the Number**:\\    - "metals" is the plural form of "metal." According to the grammar rules, the plural of "@métal@" is formed by replacing the singular morpheme @-al@ with the morpheme @-aux@. Therefore, "@métal@" should become "@métaux@" in the plural.\\ \\ 3. **Determine the article to use**:\\    - The definite article in @French@ for plural nouns (regardless of gender) is "@les@."\\ \\ 4. **Combine the article with the plural noun**:\\    - According to the grammar book, we should place the definite article "@les@" after the plural noun "@métaux@."\\ \\ Putting it all together, the translation for the metals in @French@ is:\\ \\ <translation>@les métaux@</translation>
\end{tcolorbox}

\begin{tcolorbox}[colback=gray!5!white, colframe=gray!75!black, breakable, title=Ciphered Conlang Prompt]

Tedufebéuc is a recently devised conlang. You are to translate the following English segment into Tedufebéuc with the help of a few dictionary entries and excerpts from a grammar book.\\ \\ Here is the text segment you must translate:\\ the metals\\ \\ Here are a few dictionary entries that may be of use to you; note that each entry follows the format: English lemma : Tedufebéuc equivalent (grammatical gender and/or part of speech).\\ metal : verekha (masc. n.)\\ \\ And here are relevant excerpts from a grammar book: \\ \\ Beginning of Tedufebéuc Grammar Book Excerpts\\ \\ARTICLES AND DETERMINERS\\ \\ In Tedufebéuc, determiners, including articles, are required with nearly every common noun, unlike in English. They must match the noun in gender (masculine or feminine) and number (singular or plural), but most have a unified plural form for both genders.\\ \\ While articles are formally a subset of determiners, they are conventionally treated as separate, and this treatment is adopted here.\\ \\ + Articles\\ Tedufebéuc uses three types of articles: the definite article, which often corresponds to English 'the'; the indefinite article, similar to English 'a/an'; and the partitive article, roughly equivalent to 'some' in English.\\ \\ ++ Definite article\\ In Tedufebéuc, definite articles, like the English 'the', indicate specific nouns. However, they adapt to reflect the gender (masculine or feminine) and number (singular or plural) of the noun they modify.\\ \\ The definite article always comes after the noun it determines.\\ \\ For singular nouns:\\ \\ The article ebv is used with masculine singular nouns. For instance, "ekerted ebv" (the cat) demonstrates that the noun ekerted (cat) is singular and masculine.\\ \\ For feminine singular nouns, the article erv applies. An example is "ufzyciera erv" (the house), where erv shows that the noun ufzyciera (house) is singular and feminine.\\ \\ For plural nouns, regardless of their gender:\\ \\ - The article used is ycebv for masculine and feminine plural nouns alike. (e.g., "ycekerted ycebv" means the cats; "ycufzyciera ycebv" means the houses.)\\ \\ To summarize, ebv corresponds to masculine singular nouns, erv to feminine singular nouns, and ycebv to all plural nouns, marking both grammatical gender and number in Tedufebéuc.\\ \\ Whenever the singular masculine article ebv is preceded by h, p, eb, ê, some occurrences of t, i, er, z, è, or ç, it changes to 'v.\\ \\NOUNS\\ \\ + Case\\ Tedufebéuc nouns are not modified for additional grammatical categories. (Although personal pronouns do reflect case and person.)\\ \\ + Number\\ Unlike English, Tedufebéuc plural forms are created using prefixation. \\ \\ This involves adding the morpheme yc- as a prefix to the singular noun (for example, ufzyciera > ycufzyciera 'houses').\\ \\ There are seven nouns beginning with èz- that take the prefix àf- in their plural form. These nouns are: èzted, èzs, èzlit, èzonèzon, èzvviered, èzonil, and èzufebàk (e.g., èzufebàk --> àfèzufebàk 'knees'). All other nouns starting with èz-, such as èzycil and èzéek, follow the regular plural pattern by prefixing yc- (e.g., èzéek --> ycèzéek 'holes').\\ \\ Nouns starting with ver- generally pluralize to àfèer- (e.g., veruféèzon -> àfèeruféèzon 'newspapers'). A few exceptions, however, take yc- as their plural marker: verekiedhé, verqiekycebuc, veràkhé, verl, verqéebyc, verqeruféered, verederted, and vered (e.g., verqiekycebuc ebv -> ycverqiekycebuc ycebv).\\ \\ Singular nouns starting with ob-, àf-, or yc- retain their form in the plural (e.g., àfizéed erv 'the cross' > àfizéed ycebv 'the crosses').\\ \\ The plural of nouns beginning with èer- and èeb- is typically formed by prefixing àf- instead of yc- (e.g., èerçèek > àfèerçèek 'pipes', èebon > àfèebon 'games'). However, the nouns èerguferv, èerééeryc, èebufs, and èebvl are exceptions and prefix yc- (e.g., èebufs > ycèebufs 'tires').\\ \\ Nouns that start with vier- generally form their plural with àfèer-. Examples are: vierl, vierah, vierézed, vieréisèzyc, vierqeréek, and vieréekiq (e.g., vieréekiq ufè becomes àfèeréekiq ycebg 'stained-glass windows'). However, some exceptions take yc- as their plural marker: vieruféebqèzàk, vierekézs, vieréiekeker, vierekhg, vieré, and vierguferted (e.g., vieré ebv becomes ycvieré ycebv 'the rails').\\ \\ + Gender\\ Tedufebéuc nouns are always associated with a grammatical gender, either masculine or feminine.\\ \\ End of Tedufebéuc Grammar Book Excerpts\\ \\ \\ A reminder that the English sentence you must translate into Tedufebéuc is:\\ the metals\\ \\ You may explain your chain of thoughts prior to producing the required translation. IMPORTANT: Do write your translation between tags in the following manner: <translation>your translation here</translation>.

\end{tcolorbox}

\begin{tcolorbox}[colback=gray!5!white, colframe=gray!75!black, breakable, title=Ciphered CoT Demonstration]

To translate the phrase "the metals" into Tedufebéuc, we need to follow a few steps based on the provided dictionary entries and grammar rules.\\ \\ 1. **Identify the noun and its properties**:\\    - The English word "metals" appears to correspond to the Tedufebéuc noun "verekha," which is a masculine noun according to the dictionary entry.\\ \\ 2. **Determine the Number**:\\    - "metals" is the plural form of "metal." According to the grammar rules, the plural of "verekha" is formed by replacing the singular morpheme ver- with the morpheme àfèer-. Therefore, "verekha" should become "àfèerekha" in the plural.\\ \\ 3. **Determine the article to use**:\\    - The definite article in Tedufebéuc for plural nouns (regardless of gender) is "ycebv."\\ \\ 4. **Combine the article with the plural noun**:\\    - According to the grammar book, we should place the definite article "ycebv" after the plural noun "àfèerekha."\\ \\ Putting it all together, the translation for the metals in Tedufebéuc is:\\ \\ <translation>àfèerekha ycebv</translation>

\end{tcolorbox}
\normalsize

%% file: appendix_5f.tex
\subsection{ENG$\rightarrow$ART: Example Instance from Parition 5F (French verbal morphology, and plural \& position of French feminine common nouns and adjectives)}
\label{sec:fiveF}
The following example is a conlang prompt (i.e., ciphered) taken from the 5f partition, and is more representative of our dataset overall, where prompts tend to be rather long, owing to the complexity and variety of linguistic phenomena addressed in all but the first three partitions (1, 2m, 2f). It is followed by its CoT demonstration.

\small

\begin{tcolorbox}[colback=gray!5!white, colframe=gray!75!black, breakable, title=Ciphered Prompt for the 5f partition in the art$\rightarrow$eng translation direction]

Cosàcévgze is a recently devised conlang. You are to translate the following English segment into Cosàcévgze with the help of a few dictionary entries and excerpts from a grammar book.\\ \\ Here is the text segment you must translate:\\ we were admiring the playful mice\\ \\ Here are a few dictionary entries that may be of use to you; note that each entry follows the format: English lemma : Cosàcévgze equivalent (grammatical gender and/or part of speech).\\ admire (v.) : hèfwésàcésà (1st group v.) \\ mouse : tezàkesàwéte (fem. n.) \\ playful : cévpàzàkeqo (adj.)\\ \\ And here are relevant excerpts from a grammar book: \\ \\ Beginning of Cosàcévgze Grammar Book Excerpts\\ \\           \\ WORD ORDER\\           \\ \\ In Cosàcévgze, sentences are generally constructed using subject-verb-object (SVO) order.\\ \\ For example: pà’hèzàsàcé lh fketewénkecé ('I love music').\\ \\       \\ NOUNS\\       \\ \\ + Number\\ In Cosàcévgze, nouns inflect for number in a manner similar to English.\\ \\ The singular is made plural by the addition of -te (e.g., the singular fhwétezàv becomes fhwétezàvte, meaning 'houses').\\ \\ Nouns with no change\\ Nouns that end in -te, -b, or -x keep the same spelling in the plural as in the singular (e.g., lh gsàzàwéb 'the cross' remains lcéte gsàzàwéb 'the crosses').\\ \\ For seven nouns ending in -zàke, the plural is formed with -b: acévzàke, zewéjezàke, tozàke, jewépàzàke, gzezàke, pàzàkepàzàke, ghwéllzàke (e.g., acévzàke becomes acévzàkeb 'knees'). Others, like jewétezàke or ysàzàke, take the regular plural suffix -te (e.g., ysàzàke becomes ysàzàkete 'holes').\\ \\ Nouns that end in -hl generally form their plural by changing to -hkeb (e.g., pàzàkesàvhl > pàzàkesàvhkeb 'newspapers'). However, the following eight exceptions form their plural by adding -te: sàqogwéyhl, jehl, sàqoahl, gzehghl, tecésàuhl, ghl, ghsàvhuhl and dcéteywéuhl (e.g., lcé dcéteywéuhl > lcéte dcéteywéuhlte).\\ \\ Nouns ending in -hke and -céke form their plural with the ending -b instead of -te (e.g., pàcéke becomes pàcékeb 'games', ykewohke becomes ykewohkeb 'pipes'). However, exceptions include the nouns tehsàsàhke, lhvèhke, tovcéke, and jelcéke, which take -te to form the plural (e.g., tovcéke becomes tovcékete 'tires').\\ \\ Typically, nouns ending in -hwél pluralize with -hkeb. Examples following this rule are tezàketowésàhwél, ysàhuhwél, qofhwél, uwéysàhwél, jehwél, and gzàsàhwél (e.g., uwéysàhwél becomes uwéysàhkeb 'stained glass windows'). Yet, there are exceptions, which instead take -te: sàhwél, azàkeucésàvhwél, hyywésàhwél, tozàsàyhwél, èqoyhwél, and gzehvèhwél (e.g., kev sàhwél > ècéte sàhwélte 'rails').\\ \\ + Case\\ Nouns in Cosàcévgze retain a single form regardless of grammatical categories. (Personal pronouns, on the other hand, vary by case and person.)\\ \\ + Gender\\ In Cosàcévgze, every noun has a grammatical gender, categorized as either masculine or feminine.\\ \\           \\ ADJECTIVES\\           \\ \\ Cosàcévgze adjectives must agree in gender and number with the noun they modify, resulting in four standard forms: masculine singular, feminine singular, masculine plural, and feminine plural. Certain adjectives, such as jecéhke and vzàkeucéhke, also use a fifth form before nouns beginning with qo, wé, cé, qà, ke, ê, wo, h, some occurrences of ze, or zà. Examples include: kev jecéhke pàhsàèwév (a beautiful garden), kev jecél zezàffcé (a handsome man), kevcé jecéllcé dcéffcé (a beautiful woman), lcéte jecéhkeb cévdhvyte (the beautiful children), and lcéte jecéllcéte fhwétezàvte (the beautiful houses).\\ \\ In dictionaries, adjectives are listed in their masculine singular form, which is their base. Most feminine forms are derived by adding -cé to the masculine. Examples: lcévy   > lcévycé, tocésàtehv   > tocésàtehvcé, asàhvè   > asàhvècé.\\ \\ Occasionally, small changes occur when forming feminine adjectives. Adjectives with masculine forms ending in -zàte, -hte, -wél, -cév, -cél, or -zàv double their final consonant before appending -cé. Examples are: hvgwécév -> hvgwécévvcé, jezàv -> jezàvvcé, acévywél -> acévywéllcé, asàzàte -> asàzàtetecé, gsàkecél -> gsàkecéllcé, and jehte -> jehtetecé.\\ \\ Furthermore, specific changes are applied to form feminine adjectives:\\ \\ - -céd changes to -qàucé (e.g., jesàcéd > jesàqàucé);\\ - -césà changes to -qàsàcé (e.g., qoysàhvacésà > qoysàhvaqàsàcé);\\ - -zàke has a special form -zàl (becoming -zàllcé in feminine) that appears before wo, some occurrences of ze, cé, h, ê, qà, ke, qo, zà, or wé (e.g., dzàke/dzàl > dzàllcé, fzàke/fzàl > fzàllcé);\\ - -céy changes to -qàycé (e.g., wévnkewécéy > wévnkewéqàycé);\\ - -cékesà or -cékeb change to -céketecé (e.g., tosàzàfcéyycékesà > tosàzàfcéyycéketecé, dkesàwécékeb > dkesàwécéketecé), with the exception of the irregular uwécékeb (old) whose feminine form is uwécéwéllcé;\\ - -a changes to -akecé (e.g., lzàva > lzàvakecé);\\ - -g changes to -gzecé in the feminine (e.g., jelhvg > jelhvgzecé);\\ - -wéd changes to -wéucé (e.g., hgywéd > hgywéucé).\\ \\ Adjectives with a basic form ending in -cé do not change in the feminine. For instance: kev zezàffcé sàwégzecé becomes kevcé dcéffcé sàwégzecé ('a rich man' becomes 'a rich woman').\\ \\ Plurals in Cosàcévgze are usually created by adding -te to the singular form for both masculine and feminine adjectives (e.g., pàzàlwé -> pàzàlwéte, pàzàlwécé -> pàzàlwécéte).\\ \\ All Cosàcévgze adjectives ending in -céhke take the ending -b in the masculine plural (e.g., vzàkeucéhke -> vzàkeucéhkeb). Adjectives ending in -hl normally change to -hkeb in the masculine plural (e.g., zewéucésàvhl -> zewéucésàvhkeb, gcévysàhl -> gcévysàhkeb), with exceptions: dhyhl -> dhyhlte, vhuhl -> vhuhlte. If the basic form ends in -te, -b, or -x, the masculine plural does not change (e.g., jehte -> jehte, aqovqosàcékeb -> aqovqosàcékeb, èzàkeb -> èzàkeb). The feminine plural consistently adheres to the general rule: gcévysàhlcé -> gcévysàhlcéte.\\ \\ Adjective placement in Cosàcévgze often involves placing the adjective after the noun, as with colors (e.g., lcé uwév sàzàkeacé, 'the red wine'). However, short, common adjectives related to beauty, age, goodness, or size (BAGS) precede the noun. Some examples are: vzàkeucéhke (new), zehkey (high), tocéywéy (small), pàzàlwé (pretty), jesàcéd (brief), lzàva (long), fhkeuhwéte (bad), pàcékevcé (young), dhkeb (false), asàhvè (big/tall), jecéhke (beautiful), fcéwéllcékesà (best), jezàv (good), asàzàte (fat/large), and uwécékeb (old); e.g., kevcé jecéllcé dcéffcé ('a beautiful woman').\\ \\                                 \\ ARTICLES AND DETERMINERS\\                                 \\ \\ In Cosàcévgze, determiners, including articles, are required with nearly every common noun, unlike in English. They must match the noun in gender (masculine or feminine) and number (singular or plural), but most have a unified plural form for both genders.\\ \\ While articles are formally a subset of determiners, they are conventionally treated as separate, and this treatment is adopted here.\\ \\   Articles\\ Three types of articles exist in Cosàcévgze: the definite article, which corresponds to English 'the'; the indefinite article, akin to 'a/an'; and the partitive article, comparable to 'some' in English.\\ \\    Definite article\\ Definite articles in Cosàcévgze are comparable to the English word 'the' and are used to point to specific nouns. However, they differ by being inflected according to the noun's gender (masculine or feminine) and number (singular or plural).\\ \\ The definite article is always positioned before the noun.\\ \\ For singular nouns:\\ \\ The article lcé is used with masculine singular nouns. For instance, "lcé gzehy" (the cat) demonstrates that the noun gzehy (cat) is singular and masculine.\\ \\ For feminine singular nouns, the article lh applies. An example is "lh fhwétezàv" (the house), where lh shows that the noun fhwétezàv (house) is singular and feminine.\\ \\ For plural nouns, the article is always lcéte, regardless of gender:\\ \\ - Examples include "lcéte gzehyte" (the cats) and "lcéte fhwétezàvte" (the houses).\\ \\ To sum up, masculine singular nouns use lcé, feminine singular nouns use lh, and all plural nouns use lcéte, reflecting gender and number in Cosàcévgze.\\ \\ Before some occurrences of ze, qo, qà, h, ê, wo, zà, ke, cé, and wé, the singular masculine article lcé becomes l'.\\ \\ +++++\\ VERBS\\ +++++\\ \\ Verbs in Cosàcévgze are conjugated to indicate several grammatical features:\\ \\   Mood: indicative, imperative, subjunctive, or conditional\\   Tense: past, present, or future (not all tenses combine with all moods)\\   Aspect: perfective or imperfective\\   Voice: active, passive, or reflexive\\   Nonfinite forms: participles, gerunds, infinitives\\ \\ Finite mood verbs (indicative, imperative, subjunctive, conditional) also conjugate to agree with their subjects in person (first, second, or third) and number (singular or plural). As in English, the subject must be expressed (except in the imperative mood), meaning Cosàcévgze is not a null-subject or pro-drop language.\\ \\ Auxiliary verbs combine with past participles of main verbs to form compound tenses, such as the compound past. Most main verbs use the auxiliary huzàwésà ('to have'), while reflexive and certain intransitive verbs use forms of êysàcé ('to be'). The participle agrees with the subject when the auxiliary is êysàcé, and with a preceding direct object (if any) when the auxiliary is huzàwésà. Forms of êysàcé are also used to create the passive voice by combining with past participles of transitive verbs.\\ \\   Morphology\\ \\ In Cosàcévgze, verbs are conjugated to reflect mood, tense, and agreement with the subject in both person and number. The set of all inflected forms of a verb is known as its conjugation, a term derived from Latin grammar.\\ \\    Stems and Endings\\ \\ Simple (one-word) verb forms in Cosàcévgze are made up of a stem (or root), which identifies the verb, and an ending (inflection), which conveys the verb's mood, tense, and subject's person and number. Some second-conjugation verbs add a suffix -wétete- between the stem and the ending.\\ \\ Examples:\\ - In tohsàlhwécévy, tohsàl- is the stem for tohsàlcésà (to speak), and -hwécévy marks third-person plural imperfect indicative.\\ - In dwévwétetezàvte, the stem dwév- is from dwévwésà (to finish), followed by -wétete-, with -zàvte indicating first-person plural present indicative or imperative.\\ \\ Verb conjugations combine with subject pronouns to indicate the subject of the verb.\\ \\    Fixed Stem Principle\\ \\ Stems in the first and second conjugations typically remain unchanged:\\ - tohsàlcésà: pàcé tohsàlcésàhwéte, yke tohsàlhte, nke'wélte tohsàlhtetecévy, nkecé vzàkete tohsàlwézàvte, tohsàlcéx...\\ - dwévwésà: pàcé dwévwésàhwéte, uzàkete dwévîycéte, nke'wélte dwévwétetecévy, dwévwéte, nkecé vzàkete dwévwétetewézàvte...\\ \\ In third-conjugation verbs, stems often vary, even within a single tense:\\ - uzàkelzàwésà: pàcé ucékeb, yke ucékeb, wél ucékey, vzàkete uzàkelzàvte, uzàkete uzàkelcéx, wélte ucékelcévy.\\ \\ Generally, however, nothing is removed from the stem.\\ \\    Endings\\ \\ Endings indicate:\\ - Mood and tense for all verbs\\ - Person and number for finite verbs\\ - Gender and number for past participles\\ \\ While most verbs adhere to predictable ending patterns, highly common verbs like huzàwésà, êysàcé, hllcésà, and dhwésàcé are considered irregular.\\ \\   Formation of Simple Tenses (Active)\\ \\ Simple tenses are formed without auxiliaries. More details are provided in the following section.\\ \\    Infinitive\\ \\ The infinitive form of a verb serves as its base name in Cosàcévgze. Based on their infinitives, verbs fall into three groups:\\ \\ - First group: Verbs ending in -césà. The exception is hllcésà, which is irregular and belongs to the third group.\\   Examples are tosàzàfcévcésà, gzehvycésà, hwéfcésà, jehlhwocésà, sàcéahsàècésà, dcésàfcésà, cévuzàwocésà, tohwocésà, tohtetecésà, and fhvacésà.\\   Example: tohsàlcésà: stem tohsàl-, ending -césà.\\ \\ - Second group: Verbs ending in -wésà with present participles ending in -wétetehvy.\\   Examples are zehïsà, tehlwésà, tozàlwésà, èqoakecésàtowésà, wévucéteywésà, jeqovwésà, dlcékesàwésà, asàhvèwésà, sàzàkeawésà, sàkeawésà, and gzàftohywésà.\\   Example: dwévwésà: stem dwév-, ending -wésà.\\ \\ - Third group: All other verbs, including irregular verbs like hllcésà.\\ \\ Note: Auxiliary verbs such as êysàcé and huzàwésà are traditionally classified separately from the third conjugation.\\ \\    Future\\ \\ The future tense endings in Cosàcévgze correspond to the present indicative forms of the verb huzàwésà. These regular endings are: -hwé, -hte, -h, -zàvte, -céx, -zàvy.\\ \\ The future stem is formed by taking the verb's infinitive and appending the appropriate ending. Examples include:\\ \\ - pàcé gzezàwétewésàhwé ('I will choose')\\ - yke sàcéahsàècésàhte ('you will watch')\\ - céllcé tezàsàywésàh ('she will exit')\\ - vzàkete ysàhuhwéllcésàzàvte ('we will work')\\ - uzàkete sàzàkeawésàcéx ('you will blush')\\ - wélte tohsàywésàzàvy ('they will leave')\\ \\ Many third-group verbs feature irregular future stems, although their endings remain consistent.\\ \\ Example:\\ \\ --- 1st Group: sàcéahsàècésà (to watch)\\ \\ **Singular Forms**\\ - **1st person:** pàcé sàcéahsàècésàhwé\\ - **2nd person:** yke sàcéahsàècésàhte\\ - **3rd person:** wél/céllcé/zàv sàcéahsàècésàh\\ \\ **Plural Forms**\\ - **1st person:** vzàkete sàcéahsàècésàzàvte\\ - **2nd person:** uzàkete sàcéahsàècésàcéx\\ - **3rd person:** wélte/céllcéte sàcéahsàècésàzàvy\\ \\     \\ \\ --- 2nd Group: gzezàwétewésà (to choose)\\ \\ **Singular Forms**\\ - **1st person:** pàcé gzezàwétewésàhwé\\ - **2nd person:** yke gzezàwétewésàhte\\ - **3rd person:** wél/céllcé/zàv gzezàwétewésàh\\ \\ **Plural Forms**\\ - **1st person:** vzàkete gzezàwétewésàzàvte\\ - **2nd person:** uzàkete gzezàwétewésàcéx\\ - **3rd person:** wélte/céllcéte gzezàwétewésàzàvy\\ \\     \\ \\ --- 3rd Group: ècétegcévèsàcé (to go down)\\ \\ **Singular Forms**\\ - **1st person:** pàcé ècétegcévèsàhwé\\ - **2nd person:** yke ècétegcévèsàhte\\ - **3rd person:** wél/céllcé/zàv ècétegcévèsàh\\ \\ **Plural Forms**\\ - **1st person:** vzàkete ècétegcévèsàzàvte\\ - **2nd person:** uzàkete ècétegcévèsàcéx\\ - **3rd person:** wélte/céllcéte ècétegcévèsàzàvy\\ \\     \\ \\ **Note:** The following verbs have a double sà in the future tense: cévuzàwocésà (pà'cévucésàsàhwé), hgnkeqosàwésà and gzàvnkeqosàwésà (pà'hgnkecésàsàhwé, pàcé gzàvnkecésàsàhwé), tozàkeuzàwésà (pàcé tozàkesàsàhwé), uzàwésà (pàcé ucésàsàhwé), sàcévuzàwocésà (pàcé sàcévucésàsàhwé), gzàkesàwésà (pàcé gzàkesàsàhwé), fzàkesàwésà (pàcé fzàkesàsàhwé), gzezàwésà and qogzezàwésà (wél gzecésàsàh, wél qogzecésàsàh).\\ \\    Past participle\\ \\ Past participles in Cosàcévgze, unlike present participles and gerundives, are inflected for gender and number by adding -cé and -te, similar to adjectives. For example: "kev dsàkewéy gzàvdwéy", "kevcé tozàwésàcé gzàvdwéycé", "ècéte dsàkewéyte gzàvdwéyte", and "ècéte tozàwésàcéte gzàvdwéycéte".\\ \\ The masculine singular form of a past participle usually ends in -qo for first-group verbs and -wé for second-group verbs. Third-group verbs, however, display a variety of endings, such as:\\ - -ke: (cévycévèsàcé cévycévèke, jezàwésàcé jeke, lwésàcé lke, etc.; tehuzàwésà teke, uzàwésà uke, tozàkeuzàwésà toke)\\ - -wéte: (fcéyysàcé fwéte, tosàcévèsàcé tosàwéte, etc.)\\ - -kete: (wévglkesàcé wévglkete, sàcéglkesàcé sàcéglkete; these are the only examples)\\ - -wéy: (fhkeèwésàcé fhkeèwéy, èwésàcé èwéy, etc.)\\ - -y: (verbs ending in -wévèsàcé, e.g., tocéwévèsàcé tocéwévy)\\ - -césày: (zàkeusàwésà zàkeucésày, gzàkeusàwésà gzàkeucésày, zàddsàwésà zàddcésày, tezàkeddsàwésà tezàkeddcésày)\\ - -céke: (huzàwésà céke).\\ \\    Compound Tenses\\ \\ Cosàcévgze compound tenses are formed by using an auxiliary verb (huzàwésà or êysàcé) alongside the past participle of the primary verb. They are applied to express completed actions or events related to a particular point in time. There is a range of compound tenses in Cosàcévgze, each adding a unique layer of detail to the chronology of actions.\\ \\    Compound Past\\ \\ The tohteteqo gzàftozàteqo, or compound past, is one of the most commonly used tenses in Cosàcévgze. It is employed to narrate completed past actions and corresponds to the English simple past tense. To form this tense, the auxiliary verb (huzàwésà or êysàcé) in the present tense is paired with the past participle of the main verb. For example: "pà'hwé fhvaqo" ('I ate') or "wél cétey hsàsàwéuqo" ('He arrived'). huzàwésà is generally used as the auxiliary verb, except for certain verbs (especially those denoting movement or state change) and all reflexive verbs, which use êysàcé. If uncertain, huzàwésà is the safer option.\\ \\ 1st Group: sàcéahsàècésà (to watch)\\ Singular:\\ - 1st person: pà'hwé sàcéahsàèqo\\ - 2nd person: yke hte sàcéahsàèqo\\ - 3rd person: wél/céllcé/zàv h sàcéahsàèqo\\ Plural:\\ - 1st person: vzàkete huzàvte sàcéahsàèqo\\ - 2nd person: uzàkete hucéx sàcéahsàèqo\\ - 3rd person: wélte/céllcéte zàvy sàcéahsàèqo\\ \\ 2nd Group: gzezàwétewésà (to choose)\\ Singular:\\ - 1st person: pà'hwé gzezàwétewé\\ - 2nd person: yke hte gzezàwétewé\\ - 3rd person: wél/céllcé/zàv h gzezàwétewé\\ Plural:\\ - 1st person: vzàkete huzàvte gzezàwétewé\\ - 2nd person: uzàkete hucéx gzezàwétewé\\ - 3rd person: wélte/céllcéte zàvy gzezàwétewé\\ \\ 3rd Group: ècétegcévèsàcé (to go down)\\ Singular:\\ - 1st person: pàcé tekewéte ècétegcévèke(cé)\\ - 2nd person: yke céte ècétegcévèke(cé)\\ - 3rd person: wél/céllcé/zàv cétey ècétegcévèke(cé)\\ Plural:\\ - 1st person: vzàkete tezàffcéte ècétegcévèke(cé)te\\ - 2nd person: uzàkete êycéte ècétegcévèke(cé)(te)\\ - 3rd person: wélte/céllcéte tezàvy ècétegcévèke(cé)te\\ \\ \\    Past participle agreement\\ \\ Cosàcévgze past participles are used in three ways: as adjectives, in passive voice constructions, and in compound tense formations. When acting as an adjective, they conform to standard adjective agreement rules. In the passive voice, they must always agree with the subject of the sentence. In compound tenses like the compound past, agreement rules are complex, involving the interplay between their descriptive role (requiring agreement) and their role in tense formation (which does not typically require agreement).\\ \\ When the auxiliary is êysàcé and the verb is reflexive, agreement rules align with those for verbs using huzàwésà. The reflexive pronoun corresponds to either the direct or indirect object of the verb.\\ \\ If the direct object precedes the past participle, agreement is required:\\ - (direct object pronoun) pà'hwé dhwéy kevcé yhsàycé. lcéte cévdhvyte tecé lh tezàvy tohsàyhaqocé. ('I made a pie. The children shared it.')\\ - (nkecé relative) lh yhsàycé nkecé lcéte cévdhvyte tecé tezàvy tohsàyhaqocé ('the pie that the children shared')\\ - (wh-question) nkecéllcé yhsàycé tecé tezàvy-wélte tohsàyhaqocé ? ('Which pie did they share?')\\ \\ The reflexive pronoun itself may act as the direct object, necessitating agreement with the subject. This includes inherently reflexive verbs:\\ - (ordinary reflexive) céllcéte tecé tezàvy tekewéuwécéte. vzàkete vzàkete tezàffcéte tehlkeqote. ('They (fem.) followed each other. We greeted each other.')\\ - (inherently reflexive) wélte tecé tezàvy fzànkeqote ècé fzàwé. vzàkete vzàkete tezàffcéte tezàkeucévkete ècé l'qouqovcéfcévy. ('They made fun of me. We remembered the event.')No agreement occurs if there is no direct object, or the direct object follows the past participle. In these cases, the reflexive pronoun serves as the indirect object:\\ - (no direct object) céllcéte tecé tezàvy tekeggqoèqo. vzàkete vzàkete tezàffcéte tohsàlqo. ('They (fem.) succeeded one another. We spoke with each other.')\\ - (direct object after verb) céllcéte tecé tezàvy tozàteqo ècéte nkecéteywézàvte. ('They (fem.) asked each other some questions.')\\ \\ \\ \\    Agreement occurs when the auxiliary is êysàcé and the verb is non-reflexive. The past participle agrees with the subject. Example:\\ \\ céllcéte tezàvy hsàsàwéuqocéte. ('They (fem.) arrived.')\\ \\ The auxiliary verb is huzàwésà.\\ \\ If there is no direct object or the direct object appears after the past participle, the past participle does not agree (it takes the default masculine singular form):\\ - (intransitive) céllcéte zàvy èzàsàfwé. ('They (fem.) slept.')\\ - (direct object after verb) glhwésàcé h uke ècékeb jehlcéwévcéte. ('glhwésàcé saw two whales.')\\ \\ If there is a direct object and it appears before the past participle, the participle must agree with it. Examples include:\\ - (pronoun before the auxiliary) wél wo huhwéy ècékeb jehlcéwévcéte. glhwésàcé lcéte h ukecéte. ('There were two whales. glhwésàcé saw them.')\\ - (clause-initial wh-question element) nkecéllcéte jehlcéwévcéte glhwésàcé h-y-céllcé ukecéte ? ('Which whales did glhwésàcé see?')\\ - (relative clause introduced by nkecé) lcéte ècékeb jehlcéwévcéte nkecé glhwésàcé h ukecéte ('The two whales that glhwésàcé saw.')\\ \\ This rule is one of the most challenging in Cosàcévgze.\\ \\    Imperfect Indicative\\ \\ The imperfect indicative expresses ongoing or continuous actions in the past. For example, "pàcé sàcéahsàèhwéte" means "I was watching."\\ \\ The stem is invariant for each verb and is derived as follows:\\ \\ - **First group (-césà verbs):** Remove the -césà from the infinitive, e.g., sàcéahsàècésà -> sàcéahsàè-.\\ - **Second group (-wésà verbs):** Remove -wésà and add -wétete, e.g., gzezàwétewésà -> gzezàwétewétete-.\\ - **Third group:** Use the stem of the first-person plural present indicative.\\ \\ The endings are the same for all verbs: -hwéte, -hwéte, -hwéy, -wézàvte, -wécéx, -hwécévy.\\ \\ Example:\\ \\ 1st Group: sàcéahsàècésà (to watch)\\ In the singular:\\ - 1st person: pàcé sàcéahsàèhwéte\\ - 2nd person: yke sàcéahsàèhwéte\\ - 3rd person: wél/céllcé/zàv sàcéahsàèhwéy\\ In the plural:\\ - 1st person: vzàkete sàcéahsàèwézàvte\\ - 2nd person: uzàkete sàcéahsàèwécéx\\ - 3rd person: wélte/céllcéte sàcéahsàèhwécévy\\ \\ 2nd Group: gzezàwétewésà (to choose)\\ In the singular:\\ - 1st person: pàcé gzezàwétewétetehwéte\\ - 2nd person: yke gzezàwétewétetehwéte\\ - 3rd person: wél/céllcé/zàv gzezàwétewétetehwéy\\ In the plural:\\ - 1st person: vzàkete gzezàwétewétetewézàvte\\ - 2nd person: uzàkete gzezàwétewétetewécéx\\ - 3rd person: wélte/céllcéte gzezàwétewétetehwécévy\\ \\ Special Note: For the verb êysàcé, the stem is qoy-, and the endings are the same.\\ \\    Present Conditional\\ \\ The endings for the present conditional in Cosàcévgze are always the same: -hwéte, -hwéte, -hwéy, -wézàvte, -wécéx, -hwécévy. They are regular for all verbs.\\ \\ The stem used in this tense matches the stem of the future tense. Some examples include:\\ \\ - pàcé gzezàwétewésàhwéte ('I would choose')\\ - yke sàcéahsàècésàhwéte ('you would watch')\\ - céllcé tezàsàywésàhwéy ('she would exit')\\ - vzàkete ysàhuhwéllcésàwézàvte ('we would work')\\ - uzàkete sàzàkeawésàwécéx ('you would blush')\\ - wélte tohsàywésàhwécévy ('they would leave')\\ \\ Examples:\\ \\ 1st Group: sàcéahsàècésà (to look)\\ ++++++++++++++++++\\ Person          Singular                Plural\\ 1st person      pàcé sàcéahsàècésàhwéte        vzàkete sàcéahsàècésàwézàvte\\ 2nd person      yke sàcéahsàècésàhwéte        uzàkete sàcéahsàècésàwécéx\\ 3rd person      wél/céllcé/zàv sàcéahsàècésàhwéy wélte/céllcéte sàcéahsàècésàhwécévy\\ ++++++++++++++++++\\ \\ 2nd Group: gzezàwétewésà (to choose)\\ ++++++++++++++++++\\ Person          Singular                 Plural\\ 1st person      pàcé gzezàwétewésàhwéte          vzàkete gzezàwétewésàwézàvte\\ 2nd person      yke gzezàwétewésàhwéte          uzàkete gzezàwétewésàwécéx\\ 3rd person      wél/céllcé/zàv gzezàwétewésàhwéy  wélte/céllcéte gzezàwétewésàhwécévy\\ ++++++++++++++++++\\ \\ 3rd Group: ècétegcévèsàcé (to go down)\\ ++++++++++++++++++\\ Person          Singular                Plural\\ 1st person      pàcé ècétegcévèsàhwéte        vzàkete ècétegcévèsàwézàvte\\ 2nd person      yke ècétegcévèsàhwéte        uzàkete ècétegcévèsàwécéx\\ 3rd person      wél/céllcé/zàv ècétegcévèsàhwéy wélte/céllcéte ècétegcévèsàhwécévy\\ ++++++++++++++++++\\ \\    Present Indicative\\ \\ The present indicative is the verb form used to describe actions happening in the present. For example, pàcé sàcéahsàècé means "I watch."\\ \\ The stem of the present indicative is not always consistent, especially in third-conjugation verbs, and there are three primary sets of endings:\\ \\ - Verbs ending in -césà (first group): -cé, -céte, -cé, -zàvte, -céx, -cévy.\\ - Verbs ending in -wésà (second group and most third-group verbs): -wéte, -wéte, -wéy, -wétetezàvte, -wétetecéx, -wétetecévy. Note: Irregularities are common, especially in the third group.\\ - Verbs ending in -sàcé (part of the third group): -te, -te, -, -zàvte, -céx, -cévy.\\ \\ Examples:\\ \\ 1st Group: sàcéahsàècésà (to watch)\\ ******************\\ Person          Singular             Plural\\ 1st person      pàcé sàcéahsàècé         vzàkete sàcéahsàèzàvte\\ 2nd person      yke sàcéahsàècéte        uzàkete sàcéahsàècéx\\ 3rd person      wél/céllcé/zàv sàcéahsàècé wélte/céllcéte sàcéahsàècévy\\ ******************\\ \\ 2nd Group: gzezàwétewésà (to choose)\\ ******************\\ Person          Singular             Plural\\ 1st person      pàcé gzezàwétewéte         vzàkete gzezàwétewétetezàvte\\ 2nd person      yke gzezàwétewéte         uzàkete gzezàwétewétetecéx\\ 3rd person      wél/céllcé/zàv gzezàwétewéy wélte/céllcéte gzezàwétewétetecévy\\ ******************\\ \\ 3rd Group: ècétegcévèsàcé (to go/get down)\\ ******************\\ Person          Singular             Plural\\ 1st person      pàcé ècétegcévète        vzàkete ècétegcévèzàvte\\ 2nd person      yke ècétegcévète        uzàkete ècétegcévècéx\\ 3rd person      wél/céllcé/zàv ècétegcévè wélte/céllcéte ècétegcévècévy\\ ******************\\ \\ Note: Verbs of the second group take an -wétete- in the plural forms.\\ \\         \\ PRONOUNS\\         \\ \\ In Cosàcévgze, pronouns are inflected to indicate their function in the sentence—such as subject, direct object, or indirect object—and to reflect the referent's person, gender, and number.\\ \\   Personal Pronouns\\ Cosàcévgze personal pronouns, equivalent to English pronouns like I, you, he/she, we, and they, indicate the person and number of their referent. In the third person, they also reflect gender. These pronouns take different forms to match their role in the clause, whether as subject, direct object, or another.\\ \\ The subject pronouns, also called subjective or nominative pronouns, are the forms used when a pronoun acts as the subject of a sentence.\\ \\    Subject Pronouns\\ \\ The subject pronouns in Cosàcévgze are categorized by number and person.\\ \\ **Singular:**\\ 1st: pàcé\\ 2nd (informal): yke\\ 2nd (formal): uzàkete\\ 3rd (masculine): wél\\ 3rd (feminine): céllcé\\ 3rd (neutral): zàv\\ \\ **Plural:**\\ 1st: vzàkete\\ 2nd: uzàkete\\ 3rd (masculine): wélte\\ 3rd (feminine): céllcéte\\ \\ Important: The subject pronoun pàcé is shortened to pà' when it comes before qà, cé, ke, some occurrences of ze, wo, h, qo, ê, zà, or wé.\\ \\   Relative pronouns\\ \\ Cosàcévgze employs relative pronouns, much like English, to introduce relative clauses. The choice of relative pronoun depends on its role in the clause, such as subject or direct object.\\ \\ When the relative pronoun acts as the subject of the clause's verb, the pronoun nkewé is generally used: l'zezàffcé nkewé h uzàlqo fh jewégwoglcéyycé ('the man who stole my bike'). Notably, nkewé remains unchanged regardless of the gender or number of its antecedent, e.g., lcéte jewégwoglcéyycéte nkewé zàvy qoyqo uzàlqocéte ('the bikes that were stolen').\\ \\ If the relative pronoun serves as the direct object of the clause's verb, nkecé (or nke' before zà, h, ê, qo, wé, some occurrences of ze, qà, cé, wo, or ke) is typically used: lh jewégwoglcéyycé nke'wél h uzàlqocé ('the bicycle that he stole'). As with nkewé, nkecé remains unchanged regardless of its antecedent.\\ \\ End of Cosàcévgze Grammar Book Excerpts\\ \\ \\ A reminder that the English sentence you must translate into Cosàcévgze is:\\ we were admiring the playful mice\\ \\ You may explain your chain of thoughts prior to producing the required translation. IMPORTANT: Do write your translation between tags in the following manner: <translation>your translation here</translation>.\\

\end{tcolorbox}

\begin{tcolorbox}[colback=gray!5!white, colframe=gray!75!black, breakable, title=Ciphered CoT Demonstration]
To translate "we were admiring the playful mice" into Cosàcévgze, we need to break down the components of this text segment based on the provided grammar and dictionary entries, identify their translation and order these constituents correctly.\\ \\ 1. **Identify the elements and their properties**:\\ - The subject of the sentence is "we," which corresponds to the first person plural subject pronoun in Cosàcévgze. According to the grammar book, the equivalent Cosàcévgze pronoun is therefore "vzàkete."\\ - The English verb in "we were admiring the playful mice" is in the past continuous and corresponds to the Cosàcévgze verb "hèfwésàcésà", which is a first group verb according to the dictionary: "admire (v.) : hèfwésàcésà (1st group v.)". According to the grammar book, and since we need to translate the English past continuous tense into Cosàcévgze, we should use the imperfect indicative tense construction.\\ - Since "vzàkete" is a first person plural subject pronoun, we must conjugate the verb "hèfwésàcésà" in the first person plural of the imperfect indicative.\\ - The stem of "hèfwésàcésà" is "hèfwésà-"; given the pronoun "vzàkete," and given that the verb belongs to the first group, we select -wézàvte as the inflectional marker and find that the expected verb form should be: "hèfwésàwézàvte." \\ - The definite article in Cosàcévgze for plural nouns (regardless of gender) is "lcéte."\\ - The English word "mice" corresponds to the Cosàcévgze noun "tezàkesàwéte," which is a feminine noun.\\ - "mice" refers to the plural form of "mouse." According to the grammar rules, the plural of "tezàkesàwéte" is identical to the singular. Therefore, "tezàkesàwéte" becomes "tezàkesàwéte" in the plural.\\ - The adjective is "playful," which translates to "cévpàzàkeqo."\\ - Given that "tezàkesàwéte" is a feminine noun, we need to find the feminine form of "cévpàzàkeqo". According to the grammar book, the feminine form of cévpàzàkeqo adds the morpheme cé: cévpàzàkeqocé.\\ - The adjective "cévpàzàkeqocé" (playful) also needs to be in plural form. According to the grammar book, the plural of adjectives such as "cévpàzàkeqocé" is formed by adding the morpheme -te, so "cévpàzàkeqocé" becomes "cévpàzàkeqocéte" in the plural.\\ \\ 2. **Determine constituent order**:\\ - In Cosàcévgze the subject pronoun always precedes the main verb.\\ - We should place the definite article "lcéte" before the plural noun "tezàkesàwéte."\\ - The sentence structure in Cosàcévgze typically places adjectives after their nouns unless they belong to the specific category of adjectives that precede nouns. Since "playful" does not belong to this category, it comes after the noun.\\ - According to the grammar book, direct object nouns always follow the verb in Cosàcévgze.\\ \\ Now, we can present the full translation:\\ \\ <translation>vzàkete hèfwésàwézàvte lcéte tezàkesàwéte cévpàzàkeqocéte</translation>\\ 

\end{tcolorbox}
\normalsize

%% file: appendix_2m.tex
\subsection{ART$\rightarrow$ENG: Conlang Example from Partition 2m (Plural and position of French masculine common nouns and adjectives)}
The following example is taken from partition 2m, with `Reversed' French as the conlang matrix, in the `simpler' art$\rightarrow$eng translation direction. It is followed by its CoT Demonstration.

\small
\begin{tcolorbox}[colback=gray!5!white, colframe=gray!75!black, breakable, title=Ciphered Prompt Example for Partition 2m]
Raçézàthaza is a recently devised conlang. You are to translate the following Raçézàthaza text segment into English with the help of a few dictionary entries and excerpts from a grammar book.\\ \\ Here is the text segment you must translate:\\ nàrèvfàçé nàipito nàtfà\\ \\ And here are a few dictionary entries that may be of use to you; note that each entry follows the format: lemma (grammatical gender and/or part of speech) : English equivalent.\\ rèvfàçé (masc. n.): nail \\ ipito (adj.): small\\ \\ And here are relevant excerpts from a grammar book: \\ \\ Beginning of Raçézàthaza Grammar Book Excerpts\\ \\           \\ ADJECTIVES\\           \\ \\ Raçézàthaza adjectives align in gender and number with the noun they modify, creating four typical forms: masculine singular, feminine singular, masculine plural, and feminine plural. Additionally, adjectives such as rènétqà and rènéthàrèvzà introduce a fifth form used after nouns ending in some occurrences of ra, rè, v, né, e, t, p, jè, çè, or ê. Examples are zàpqahanézu rènétqà zàrè (a beautiful garden), tsèsèvra fàtqà zàrè (a handsome man), tsèsètza tfàfàtqà tzàrè (a beautiful woman), nàizànézazàt surènétqà nàtfà (the beautiful children), and nàzàvnàpnésè nàtfàfàtqà nàtfà (the beautiful houses).\\ \\ The masculine singular, which serves as an adjective's base form, is what appears in dictionaries. Most feminine forms are created by adding the prefix t- to the masculine form. For instance: qazànéhaqu > tqazànéhaqu, zànénàhato > tzànénàhato, and izàtfà > tizàtfà.\\ \\ Under specific conditions, additional minor adjustments occur in the formation of feminine adjectives. Masculine adjectives beginning with nàv-, nàné-, fàt-, zàt-, fàp-, or zàv-, double their initial consonant before prefixing t-. Examples include: nàvhaqu > tnànàvhaqu, nànéqà > tnànànéqà, zàvqà > tzàzàvqà, fàtrèhaçé > tfàfàtrèhaçé, fàpizàtqu > tfàfàpizàtqu, and zàtpçézàné > tzàzàtpçézàné.\\ \\ Additionally, several changes occur in the formation of feminine adjectives:\\ \\ - harèt- or surèt- change to tnàrèt- (e.g., harètiitsèvhao --> tnàrètiitsèvhao, surètpharèza --> tnàrètpharèza), with the exception of the irregular surètphà (old) whose feminine form is tfàfàptphà;\\ - it- changes to tie- (e.g., itprèyzàp --> tieprèyzàp);\\ - çé- changes to traçé- in the feminine (e.g., çézànéfàqà --> traçézànéfàqà);\\ - rèv- has a special form fàv- (becoming tfàfàv- in feminine) that appears after jè, p, né, ê, rè, some occurrences of ra, t, v, çè, or e (e.g., rèvza/fàvza --> tfàfàvza, rèvsè/fàvsè --> tfàfàvsè);\\ - qu- changes to trèqu- (e.g., quzàvfà --> trèquzàvfà);\\ - zat- changes to thàe- (e.g., zathaqà --> thàehaqà);\\ - hat- changes to thae- (e.g., hatquzànéhaijè --> thaequzànéhaijè);\\ - zap- changes to thàp- (e.g., zapiçéné --> thàpiçéné).\\ \\ An adjective beginning with t- in its base form does not change in the feminine. For example: traçépha tsèsèvra zàrè ('a rich man') becomes traçépha tsèsètza tzàrè ('a rich woman').\\ \\ To form plurals in Raçézàthaza, nà- is typically added to the singular for masculine and feminine adjectives (e.g., pfàvzu -> nàpfàvzu, tpfàvzu -> nàtpfàvzu).\\ \\ Adjectives beginning with fàné- normally change to surèné- in the masculine plural (e.g., fànézàhathàpra -> surènézàhathàpra, fànéhaizàtçé -> surènéhaizàtçé), with exceptions: fànéinéza -> nàfànéinéza, fànéhànézà -> nàfànéhànézà. If the basic form begins with nà-, su-, or m-, the masculine plural does not change (e.g., nànéqà -> nànéqà, surèthajèzàjèqu -> surèthajèzàjèqu, surèvqa -> surèvqa). All Raçézàthaza adjectives beginning with rènét- take the prefix su- in the masculine plural (e.g., rènéthàrèvzà -> surènéthàrèvzà). The feminine plural is always derived using the general rule: tfànéhaizàtçé -> nàtfànéhaizàtçé.\\ \\ Adjective placement in Raçézàthaza follows specific patterns. Most adjectives are positioned before the noun they modify, particularly colors, as in tqurèvha zàphà tfà (the red wine). However, certain short, commonly used adjectives – those dealing with beauty, age, goodness, or size (often remembered by the acronym "BAGS") – typically follow their nouns. These include: nàpnéhàrènésè (bad), harètfàfàptsè (best), surètphà (old), rènéthàrèvzà (new), qazànéhaqu (big/tall), rènétqà (beautiful), zàvqà (good), ipito (small), irènéra (high), pfàvzu (pretty), quzàvfà (long), tzàrètzu (young), surènéza (false), nàvhaqu (fat/large), and zathaqà (brief); e.g., tsèsètza tfàfàtqà tzàrè ('a beautiful woman').\\ \\ ARTICLES AND DETERMINERS\\ \\ Unlike in English, articles and determiners are almost always used with common nouns in Raçézàthaza. They reflect the gender (masculine or feminine) and number (singular or plural) of the noun they modify, although their plural forms are usually the same for both genders.\\ \\ Although articles belong to the broader category of determiners, they are traditionally considered separately, and this separation is followed here.\\ \\   Articles\\ In Raçézàthaza, there are three articles: the definite article (often equivalent to English 'the'), the indefinite article (matching 'a/an'), and the partitive article (used like 'some' in English).\\ \\    Definite article\\ The definite article in Raçézàthaza is similar to the English word 'the' and is used to identify specific nouns. However, Raçézàthaza articles change form to match the gender (masculine or feminine) and number (singular or plural) of the noun.\\ \\ The definite article is always placed after its noun.\\ \\ For singular nouns:\\ \\ Masculine singular nouns are preceded by the article tfà. For example, "inéraçé tfà" (the cat) indicates that inéraçé (cat) is masculine and singular.\\ \\ The article néfà is used for feminine singular nouns. An example is "zàvnàpnésè néfà" (the house), where néfà marks zàvnàpnésè (house) as singular and feminine.\\ \\ For plural nouns, the article is always nàtfà, regardless of gender:\\ \\ - Examples include "nàinéraçé nàtfà" (the cats) and "nàzàvnàpnésè nàtfà" (the houses).\\ \\ To sum up, masculine singular nouns use tfà, feminine singular nouns use néfà, and all plural nouns use nàtfà, reflecting gender and number in Raçézàthaza.\\ \\ Following t, çè, v, e, rè, jè, p, né, ê, and some occurrences of ra, the masculine singular article tfà is elided to 'fà.\\ \\ NOUNS\\ -----\\ \\   Case\\ Nouns in Raçézàthaza do not change their form to reflect other grammatical categories. (Personal pronouns, however, are inflected for case and person.)\\ \\   Gender\\ Each Raçézàthaza noun is assigned a grammatical gender, which is either masculine or feminine.\\ \\   Number\\ The formation of plural nouns in Raçézàthaza relies on prefixation, in contrast to English suffixation. \\ \\ To form the plural, the prefix nà- is added to the singular noun (e.g., zàvnàpnésè changes to nàzàvnàpnésè, meaning 'houses').\\ \\ When a noun starts with m-, nà-, or su- in the singular, its plural form remains the same. An example is supvhaçé néfà 'the cross' -> supvhaçé nàtfà 'the crosses'.\\ \\ There are seven nouns beginning with rèv- that take the prefix su- in their plural form. These nouns are: rèvzàtqu, rèvfàfàpnéçé, rèvraçé, rèvzurèvzu, rèvo, rèvzupqà, and rèvqàpra (e.g., rèvzàtqu --> surèvzàtqu 'knees'). All other nouns starting with rèv-, such as rèvhai and rèvnàpqà, follow the regular plural pattern by prefixing nà- (e.g., rèvhai --> nàrèvhai 'holes').\\ \\ Nouns whose singular form starts with rèné- or rèt- use the prefix su- in their plural forms rather than nà- (e.g., rètzu becomes surètzu 'games', rènéçèrèi becomes surènéçèrèi 'pipes'). Exceptions include: rètzào, rènéqazànéfà, rènéhahanénà, and rètfàqà, which use the prefix nà- (e.g., rètzào > nàrètzào 'tires').\\ \\ Nouns starting with fàné- generally pluralize to surèné- (e.g., fànézàharèvzu becomes surènézàharèvzu 'newspapers'). A few exceptions, however, take nà- as their plural marker: fànéqà, fànéhànézàhanéçé, fànéipçéjèha, fànéqujèha, fànéhàpinàtza, fànéçénéraçé, fànéhàhatnà, and fànéçé (e.g., fànéhàpinàtza tfà becomes nàfànéhàpinàtza nàtfà).\\ \\ The plural of most nouns starting with fàpné- is formed by using the surèné- prefix. Examples are fàpnéqà, fàpnéhaiphà, fàpnéhavçé, fàpnéhànéhai, fàpnésèjè, and fàpnéhaporèvnà (e.g., fàpnéhaiphà zàrè --> surènéhaiphà nàtqa 'stained-glass windows'). However, there are exceptions that use nà- as their plural marker: fàpnéqazànéraçé, fàpnéhapiiné, fàpnézàhathàrèvqu, fàpnéha, fàpnéihavo, and fàpnéijèqa (e.g., fàpnéha tfà --> nàfàpnéha nàtfà 'the rails').\\ \\ End of Raçézàthaza Grammar Book Excerpts\\ \\ \\ A reminder that the Raçézàthaza sentence you must translate into English is:\\ nàrèvfàçé nàipito nàtfà\\ \\ You may explain your chain of thoughts prior to producing the required translation. IMPORTANT: Do write your translation between tags in the following manner: <translation>your translation here</translation>.
\end{tcolorbox}

\begin{tcolorbox}[colback=gray!5!white, colframe=gray!75!black, breakable, title=Conlang CoT Demonstration]

To translate the phrase "nàrèvfàçé nàipito nàtfà" into English, we need to follow a few steps based on the provided dictionary entries and grammar rules.\\ \\ 1. **Identify the elements and their properties**:\\   - According to the grammar book, "nàtfà" is the definite article in Raçézàthaza for plural nouns (regardless of gender).\\   - The Raçézàthaza word "nàrèvfàçé" appears to correspond to the Raçézàthaza noun "rèvfàçé," meaning "nail" according to the dictionary entry.\\   - According to the grammar rules, the plural of "rèvfàçé" is formed by adding the morpheme nà-. Therefore, "nàrèvfàçé" must be the plural form of "rèvfàçé".\\   - The Raçézàthaza word "nàipito" appears to correspond to the Raçézàthaza adjective "ipito" meaning "small" according to the dictionary entry.\\   - According to the grammar book, the plural of adjectives such as "ipito" is formed by adding the morpheme nà-, so "nàipito" must be the plural of "ipito." \\ \\ 2. **Analyze constituent order**:\\    - The placement of "nàtfà" after the plural noun "nàrèvfàçé" indicates that it determines this noun.\\    - The sentence structure in Raçézàthaza typically places adjectives before their nouns unless they belong to the specific category of adjectives that follow nouns. Since "small" belongs to this category, its Raçézàthaza equivalent comes after the noun as expected.\\ \\ Putting it all together, the translation for nàrèvfàçé nàipito nàtfà in English is:\\ \\ <translation>the small nails</translation>

\end{tcolorbox}
\normalsize

%% file: latin_test.tex
\subsection{ART$\rightarrow$ENG: Latin-Derived Conlang Example}
\label{sec:latin_example}

\small
\begin{tcolorbox}[colback=gray!5!white, colframe=gray!75!black, breakable, title=Latin-Derived Ciphered Prompt]
Ebevetjét is a recently devised conlang. You are to translate the following Ebevetjét text segment into English with the help of a few dictionary entries and excerpts from a grammar book.\\ \\ Here is the text segment you must translate:\\ øekekjécjøes rygøisenes evilevujet\\ \\ And here are a few dictionary entries that may be of use to you; note that each entry follows the format: Ebevetjét lemma, optional gender information for nouns, m. for masculine, f. for feminine and n. for neuter nouns, (optional mention of a morphological paradigm, conjugation or declension, for verbs and nouns, respectively) : English equivalent.\\ \\ evilen, evilevisø, evilevxj, evilevetygil (first conjugation) : to love\\ øekekjécjøes, øekekjécjøj f. (second declension) : statue\\ rygøis, rygøisj m. (second declension) : boy\\ \\ And here are relevant excerpts from a grammar book: \\ \\ Beginning of Ebevetjét Grammar Book Excerpts\\ \\ ***********\\ DETERMINERS\\ ***********\\ \\ * Articles\\ Ebevetjét has no definite or indefinite articles. As a result, the noun and adjective "isøit uenétyges" can be translated as either "the good king" or "a good king," depending on the context. In the absence of context, the definite article is typically preferred when translating from Ebevetjét into English.\\ \\ *****\\ NOUNS\\ *****\\ \\ * Number\\ \\ Most Ebevetjét nouns have two numbers, singular and plural: isøit "king", isøécøes "kings". A few nouns, called rwygisevwjev etevétetygil ("plural only"), although plural in form, have a singular meaning, e.g. egevesetisev "a camp", wjetetøisevø "a letter", étygretjevø "a wedding". \\ \\ * Gender\\ \\ Nouns are divided into three genders, known as masculine, feminine, and neuter. The difference is shown in the pronouns and adjectives that refer to them, for example:\\ \\     jresø isøit "the king himself" (masculine)\\     jresev isøécjétev "the queen herself" (feminine)\\     jresygil uøwwygil "the war itself" (neuter)\\ \\ To a certain extent, the genders follow the meanings of the words (for example, winds are masculine, tree-names feminine):\\ \\     Masculine nouns include all those referring to males, such as eceniljétyges "master", rygøis "boy", ecøyges "god", but also some inanimate objects such as menisetyges "garden", øitøisegjetyges "army", ilenes "custom". Words in the 2nd declension ending in -yges or -øis are usually masculine.\\     Feminine nouns include all those referring to females, such as rygøwwev "girl", ilygwjøis "woman", ecøev "goddess", but also inanimate or abstract nouns such as evisuenis "tree", ygisues "city", mygilevétjeteves "kindness", étevetjen "nation". Words in the 1st declension like rygøwwev ending in -ev are usually feminine, with a few exceptions such as renøetev "poet". Also feminine are 3rd declension nouns ending in -eteves and -etjen.\\     Neuter nouns (apart from esegenisetygil "a sex worker (of either gender)") all refer to things, such as éteniløét "name", egenisryges "body", uøwwygil "war", xøétøétygil 'poison'.\\ \\ Neuter nouns differ from masculine and feminine in two ways: (1) the plural nominative and accusative forms end in -ev, e.g. uøwwev "wars", egenisrenisev "bodies"; (2) the subject (nominative) and object (accusative) cases are identical. \\ \\ * Case\\ \\ Nouns in Ebevetjét have a series of different forms, called cases of the noun, which have different functions or meanings. For example, the word for "king" is isøit when it is the subject of a verb, but isøécøil when it is the object:\\ \\     isøit xjecøet "the king sees" (nominative case)\\     isøécøil xjecøet "(he/she/it) sees the king" (accusative case)\\ \\ Further cases mean "of" (genitive case), "to/for" (dative case), and "with" (ablative case).\\ \\ Nouns for people have a separate form used for addressing a person (vocative case). In most nouns for women and girls, the vocative is the same as the nominative.\\ \\ Some nouns, such as the names of cities and small islands, and the word ecenilyges "home", have a seventh case called the locative, for example Isenilevø "in Isenilø" or ecenilj "at home"; however, most nouns do not have this case.\\ \\ The genitive, dative and ablative cases are called the "oblique" cases.\\ \\ The order in which the cases are given in Ebevetjét grammar books differs from one author to the other.\\ \\ * Use of cases\\ \\ ** Nominative\\ \\ The nominative case is used for the subject of an active or a passive verb:\\ \\     isøit ecygegjet = the king leads\\     isøit evygecjetygis = the king is heard\\ \\ It is also used for the complement of a copula verb such as øeset "he is" or ekevegetyges øeset "he became":\\ \\     isøit øisevet Evøétøeves étenujes = our king was Evøétøeves / Evøétøeves was our king\\     isøit ekevegetyges øeset = he was made king / he became king\\ \\ ** Vocative\\ \\ The vocative case is used when addressing someone:\\ \\     ibyguøesétø ilø, Isenilygwø Isøit, ekenøecyges ekøisjisø? = do you order me, King Isenilygwyges, to strike a treaty?\\ \\ ** Accusative\\ \\ The accusative case is used for the object of a sentence:\\ \\     renøetevø isøécøil xjecøuygétet = the poets will see the king\\     ilevwyges isøit étenxenes ecygegøes etjiløet = the bad king fears the new generals\\ \\ It is also used as the subject of an infinitival clause dependent on a verb of speaking or the like:\\ \\     isøécøil jétetøisekøegetygil øesesø egisøecøuevétet = they believed that the king had been killed\\ \\ It can be the complement of another word which is itself accusative:\\ \\     Etygwwygil renrygwyges isøécøil egisøevxjet = the people made Etygwwyges their king\\ \\ It can also be used with a place name to refer to the destination:\\ \\     Isenilevil risenekøegetyges øeset = he set out for Isenilø\\ \\ The accusative is also used after various prepositions (especially those that imply motion towards):\\ \\     esøétevetyges evec isøécøil wøécevetenes iljesjet = the Esøétevetø sent ambassadors to the king\\     egenétesygw jét ygisuøil isøecjjet = the magistrate returned to the city\\ \\ Another use of the accusative is to give a length of time or distance:\\ \\     isøécétevxjet evététenes égygjétégygø = he reigned for five years\\     égygjétégygø røecøes wenétécyges = five foot tall\\ \\ ** Genitive\\ \\ A genitive noun can represent a possessor or a kin:\\ \\     isøécjes ekjwjev = the king's daughter, daughter of the king\\     wjuisj rygøwwevø = the girl's books, the books of the girl\\ \\ A genitive noun always stands for the object of mental processes such as iljesøisøenis "I pity" and enuwjxjesegenis "I forget" and the accusative is never used with those two verbs:\\ \\     renrygwyges enuwjxjesegøetygis ilevwj isøécjes = the nation will forget the bad king\\     ecygit uenétyges iljesøisøetygis rygøisenisygil = the good general pities the boys\\ \\ A genitive noun attached to a verbal noun can stand for the object of the implied verb (called an "objective genitive"):\\ \\     ilevitjilj xjisj ecøesjecøisjygil = my longing for the great man\\ \\ A genitive noun can stand for the subject of the implied verb (called a "subjective" genitive):\\ \\     Egevøesevisjes evecxøétetyges = Egevøesevis's arrival\\ \\ A frequent type of genitive is the partitive genitive, expressing the quantity of something:\\ \\     esevetjes etøilrenisjes = enough time\\ \\ ** Dative\\ \\ The dative case means "to" or "for". It is frequently used with verbs of saying or giving:\\ \\     isøécj étygétetjevetygil øeset = it was announced to the king\\     røegygétjevil isøécj egisøecjecjet = he entrusted the money to the king\\ \\ It can also be used with certain adjectives:\\ \\     egevisjesesjilyges øisevet isøécj = he was very dear to the king\\ \\ It is also used with certain verbs such as revisøen "I obey" or røisesygevecøen "I persuade":\\ \\     revisygjet isøécj = he was obedient to (i.e. obeyed) the king\\ \\ There are also various idiomatic uses, such as the dative of possession:\\ \\     égygjec øeset etjuj éteniløét? = what's your name?\\ \\ ** Ablative\\ \\ The ablative case can mean "with", especially when the noun it refers to is a thing rather than a person:\\ \\     écwevecjen esø etisevétesekjécjet = he stabbed himself with a sword\\ \\ Often a phrase consisting of a noun plus participle in the ablative can express time or circumstance. This is known as an "ablative absolute":\\ \\     isøécjuyges øitevegetjes = with the kings driven out, i.e. after the kings were driven out\\ \\ It is also frequently used with prepositions, especially those meaning "from", "with", "in", or "by":\\ \\     ygétyges ø isøécjuyges = one from (i.e. one of) the kings\\     egygil isøécjuyges = with the kings\\     ev isøécjuyges = by the kings, from the kings\\     risen isøécø = for/on behalf of the king\\     \\ \\ Another use is in expressions of time and place (except those that give the length of time or distance):\\ \\     øen etøilrenisø = at that time\\     meneg wenegen = at this place\\     revygegjes ecjøuyges = in a few days\\ \\ The ablative can also mean "from", especially with place names:\\ \\     Isenilev risenekøegetyges øeset = he set out from Isenilø\\     wenegen jwwø ilenetyges øeset = he was dislodged from his position\\     \\ Most importantly, perhaps, Ebevetjét uses the ablative case to express the agent used with a passive verb form and makes a distinction which English does not: if the agent is a person, "by" is expressed in Ebevetjét with the preposition ev/evu + an ablative noun. If, however, the agent is not a person but an instrument, the ablative case is still used but without ev/evu. Note: The evu form of the preposition is only used before a word beginning with ev, ø, j, en, m or yg: "ev isøécjuyges xjecøenis", I am seen by the kings, but "evu evécisjegenwjes xjecøenis", I am seen by the farmers.\\ \\ ** Locative\\ \\ The locative is a rare case used only with names of cities, small islands, and one or two other words such as ecenilyges "home". It means "at" or "in":\\ \\     egenétesygw evwetøis Isenilevø ilevétesjet = one of the two magistrates remained in Isenilø\\     ilygwetenes evététenes étenesetisevø ecenilj xjitjet = he lived at our house for many years\\ \\ \\ ** Declensions\\ \\ There are five declensions for Ebevetjét nouns. Each declension can be unequivocally identified by the ending of the genitive singular (-evø, -j, -jes, -yges, -øj) given in the dictionary.\\ \\ *** First Declension (ev stem)\\ Nouns of this declension usually end in -ev in the nominative singular and are mostly feminine, e.g. xjev, xjevø f. ('road'), iløétesev, iløétesevø f. ('table'), and evégygev, evégygevø f. ('water'). There is a small class of masculine exceptions generally referring to occupations, e.g. renøetev, renøetevø m. ('poet'), evécisjegenwev, evécisjegenwevø m. ('farmer'), evygisjécev, evygisjécevø m. ('evygisjécev, charioteer'), rjisevetev, rjisevetevø m. ('pirate') and étevygetev, étevygetevø m. ('sailor').\\ \\ First Declension Paradigm\\ --------------------------------------\\ Case         | Singular   | Plural\\ --------------------------------------\\ Nominative   | -ev       | -evø\\ Vocative     | -ev       | -evø\\ Accusative   | -evil      | -eves\\ Genitive     | -evø      | -evisygil\\ Dative       | -evø      | -jes\\ Ablative     | -ev       | -jes\\ Locative     | Gen.       | Dat.\\ --------------------------------------\\ \\ ----------------------------------------------\\ Case         | Singular (iløétesev) | Plural (iløétesev)\\ ----------------------------------------------\\ Nominative   | iløétesev          | iløétesevø\\ Vocative     | iløétesev          | iløétesevø\\ Accusative   | iløétesevil         | iløéteseves\\ Genitive     | iløétesevø         | iløétesevisygil\\ Dative       | iløétesevø         | iløétesjes\\ Ablative     | iløétesev          | iløétesjes\\ ----------------------------------------------\\ \\ \\ Case         | Singular (renøetev) | Plural (renøetev)\\ ----------------------------------------------\\ Nominative   | renøetev          | renøetevø\\ Vocative     | renøetev          | renøetevø\\ Accusative   | renøetevil         | renøeteves\\ Genitive     | renøetevø         | renøetevisygil\\ Dative       | renøetevø         | renøetjes\\ Ablative     | renøetev          | renøetjes\\ ----------------------------------------------\\ \\ The locative endings for the first declension are -evø (singular) and -jes (plural), similar to the genitive singular and ablative plural, as in iljwjetjevø 'in war' and Evetmøétjes 'at Evetmøétes'.\\ \\ *** Second declension (en stems)\\ \\ The second declension is a large group of nouns consisting of mostly masculine nouns like øégygyges, øégygj ('horse') and rygøis, rygøisj ('boy') and neuter nouns like egevesetøwwygil, egevesetøwwj ('fort'). There are several small groups of feminine exceptions, including names of gemstones, plants, trees, and some towns and cities.\\ \\ In the nominative singular, most masculine nouns consist of the stem and the ending -yges, although some end in -øis, which is not necessarily attached to the complete stem. Neuter nouns generally have a nominative singular consisting of the stem and the ending -ygil. However, every second-declension noun has the ending -j attached as a suffix to the root of the noun in the genitive singular form. The predominant morpheme in the ending forms of this declension is en. \\ \\ Second Declension Paradigm\\ ----------------------------------------------\\ Case         | Singular (Masculine) | Singular (Neuter) | Plural (Masculine) | Plural (Neuter)\\ ----------------------------------------------\\ Nominative   | -yges                | -ygil             | -j               | -ev\\ Vocative     | -ø                 | -ygil             | -j               | -ev\\ Accusative   | -ygil                | -ygil             | -enes              | -ev\\ Genitive     | -j                 | -j              | -enisygil            | -enisygil\\ Dative       | -en                 | -en              | -jes              | -jes\\ Ablative     | -en                 | -en              | -jes              | -jes\\ Locative     | Gen.                 | Gen.              | Dat.               | Dat.\\ ----------------------------------------------\\ \\ Masculine Declension Paradigm (eceniljétyges, eceniljétj — master, m.)\\ ----------------------------------------------\\ Case         | Singular         | Plural\\ ----------------------------------------------\\ Nominative   | eceniljétyges        | eceniljétj\\ Vocative     | eceniljétø         | eceniljétj\\ Accusative   | eceniljétygil        | eceniljétenes\\ Genitive     | eceniljétj         | eceniljétenisygil\\ Dative       | eceniljéten         | eceniljétjes\\ Ablative     | eceniljéten         | eceniljétjes\\ ----------------------------------------------\\ \\ Neuter Declension Paradigm (uøwwygil, uøwwj — war, n.)\\ ----------------------------------------------\\ Case         | Singular         | Plural\\ ----------------------------------------------\\ Nominative   | uøwwygil         | uøwwev\\ Vocative     | uøwwygil         | uøwwev\\ Accusative   | uøwwygil         | uøwwev\\ Genitive     | uøwwj          | uøwwenisygil\\ Locative     | uøwwj          | uøwwenisygil\\ Dative       | uøwwen          | uøwwjes\\ Ablative     | uøwwen          | uøwwjes\\ ----------------------------------------------\\ \\ The locative endings for the second declension are -j (singular) and -jes (plural); Egenisjétetmj "at Egenisjétetm", Iløecjenwevétj "at Iljwevét", and Ékmjwjrrjes "at Ékmjwjrrj".\\ \\ *** Second-declension -jyges and -jygil nouns\\ \\ Nouns ending in -jyges and -jygil have a genitive singular in -j in earlier Ebevetjét, which was regularized to -jj in the later stages of the language. Masculine nouns in -jyges have a vocative singular in -j at all stages. In Old Ebevetjét, however, the vocative was declined regularly, using -jø instead, e.g. ekjwjø "[O] son", archaic vocative of ekjwjyges.\\ \\ There is no contraction of -jj(es) in plural forms and in the locative.\\ \\ Second Declension Paradigms (ekjwjyges, ekjwjj — son, m. and evygitjwjygil, evygitjwjj — aid/help, n.)\\ --------------------------------------------------------------------------------\\ Case         | Singular (ekjwjyges) | Plural (ekjwjyges) | Singular (evygitjwjygil) | Plural (evygitjwjygil)\\ --------------------------------------------------------------------------------\\ Nominative   | ekjwjyges          | ekjwjj         | evygitjwjygil          | evygitjwjev\\ Vocative     | ekjwj            | ekjwjj         | evygitjwjygil          | evygitjwjev\\ Accusative   | ekjwjygil          | ekjwjenes        | evygitjwjygil          | evygitjwjev\\ Genitive     | ekjwjj           | ekjwjenisygil      | evygitjwjj           | evygitjwjenisygil\\ Dative       | ekjwjen           | ekjwjjes        | evygitjwjen           | evygitjwjjes\\ Ablative     | ekjwjen           | ekjwjjes        | evygitjwjen           | evygitjwjjes\\ --------------------------------------------------------------------------------\\ \\ \\ *** Second-declension -is nouns\\ \\ Some masculine nouns of the second declension end in -øis or -jis in the nominative singular. The declension of these nouns is identical to that of the regular second declension, except for the lack of suffix in the nominative and vocative singular.\\ \\ Some (but not all) nouns in -øis drop the ø in the genitive and other cases. For example, esenegøis, esenegøisj ('father-in-law') keeps its ø. However, the noun ilevécjesetøis, ilevécjesetisj ('(school)master') drops its ø in the genitive singular.\\ \\ Second Declension -is Paradigm\\ (rygøis, rygøisj — boy, m.; evécøis, evécisj — field, m.; xjis, xjisj — man, m.)\\ \\ ----------------------------------------------\\ Case         | Singular (rygøis) | Plural (rygøis)\\ ----------------------------------------------\\ Nominative   | rygøis            | rygøisj\\ Vocative     | rygøis            | rygøisj\\ Accusative   | rygøisygil          | rygøisenes\\ Genitive     | rygøisj           | rygøisenisygil\\ Dative       | rygøisen           | rygøisjes\\ Ablative     | rygøisen           | rygøisjes\\ ----------------------------------------------\\ \\ ----------------------------------------------\\ Case         | Singular (evécøis) | Plural (evécøis)\\ ----------------------------------------------\\ Nominative   | evécøis            | evécisj\\ Vocative     | evécøis            | evécisj\\ Accusative   | evécisygil           | evécisenes\\ Genitive     | evécisj            | evécisenisygil\\ Dative       | evécisen            | evécisjes\\ Ablative     | evécisen            | evécisjes\\ ----------------------------------------------\\ \\ ----------------------------------------------\\ Case         | Singular (xjis) | Plural (xjis)\\ ----------------------------------------------\\ Nominative   | xjis            | xjisj\\ Vocative     | xjis            | xjisj\\ Accusative   | xjisygil          | xjisenes\\ Genitive     | xjisj           | xjisenisygil\\ Dative       | xjisen           | xjisjes\\ Ablative     | xjisen           | xjisjes\\ ----------------------------------------------\\ \\ *** Third Declension\\ The third declension is the largest group of nouns. The nominative singular of these nouns may end in -ev, -ø, -j, -en, -q, -eg, -w, -ét, -is, -es, -et, or -it. This group of nouns includes masculine, neuter, and feminine nouns.\\ \\ The stem of a third declension noun may be found from the genitive case by removing the ending -jes. For example, the stem of revit, revegjes f. 'peace' is reveg-, the stem of ekwygiløét, ekwygiljétjes n. 'river' is ekwygiljét-, and the stem of ekwenes, ekwenisjes m. 'flower' is ekwenis-.\\ \\ Masculine, feminine and neuter nouns often have their own special nominative singular endings. For instance, many masculine nouns end in -enis (evilenis, evilenisjes, 'love'). Many feminine nouns end in -jit (rmenøétjit, rmenøétjegjes, 'phoenix'), and many neuter nouns end in -yges with an is stem in the oblique cases (enétyges, enétøisjes 'burden'; etøilryges, etøilrenisjes 'time'). \\ \\ Third Declension Paradigm\\ -----------------------------------------------------------------\\ Case         | Singular (Masc./Fem.) | Plural (Masc./Fem.) | Singular (Neuter) | Plural (Neuter)\\ -----------------------------------------------------------------\\ Nominative   | -es                 | -øes              | —               | -ev\\ Vocative     | -es                 | -øes              | —               | -ev\\ Accusative   | -øil                | -øes              | —               | -ev\\ Genitive     | -jes                | -ygil              | -jes             | -ygil\\ Dative       | -j                 | -juyges            | -j              | -juyges\\ Ablative     | -ø                 | -juyges            | -ø              | -juyges\\ Locative     | Dat./Abl.            | Dat.               | Dat./Abl.         | Dat.\\ -----------------------------------------------------------------\\ \\ ecygit, ecygegjes — leader, m.\\ ----------------------------------------------\\ Case         | Singular (ecygit) | Plural (ecygit)\\ ----------------------------------------------\\ Nominative   | ecygit            | ecygegøes\\ Vocative     | ecygit            | ecygegøes\\ Accusative   | ecygegøil          | ecygegøes\\ Genitive     | ecygegjes          | ecygegygil\\ Dative       | ecygegj           | ecygegjuyges\\ Ablative     | ecygegø           | ecygegjuyges\\ ----------------------------------------------\\ \\ xjisetyges, xjisetygetjes — virtue, f.\\ ----------------------------------------------\\ Case         | Singular (xjisetyges) | Plural (xjisetyges)\\ ----------------------------------------------\\ Nominative   | xjisetyges            | xjisetygetøes\\ Vocative     | xjisetyges            | xjisetygetøes\\ Accusative   | xjisetygetøil          | xjisetygetøes\\ Genitive     | xjisetygetjes          | xjisetygetygil\\ Dative       | xjisetygetj           | xjisetygetjuyges\\ Ablative     | xjisetygetø           | xjisetygetjuyges\\ ----------------------------------------------\\ \\ éteniløét, éteniljétjes — name, n.\\ ----------------------------------------------\\ Case         | Singular (éteniløét) | Plural (éteniløét)\\ ----------------------------------------------\\ Nominative   | éteniløét            | éteniljétev\\ Vocative     | éteniløét            | éteniljétev\\ Accusative   | éteniløét            | éteniljétev\\ Genitive     | éteniljétjes          | éteniljétygil\\ Dative       | éteniljétj           | éteniljétjuyges\\ Ablative     | éteniljétø           | éteniljétjuyges\\ ----------------------------------------------\\ \\ The locative endings for the third declension are -j or -ø (singular) and -juyges (plural), as in isygisj 'in the country' and Etisevwwjuyges 'at Etisevwwøes'.\\ \\ **** Irregular nouns\\ In the third declension, there are three irregular nouns. \\ \\ xjes, xjes — force, power, f.\\ ----------------------------------------------\\ Case         | Singular (xjes) | Plural (xjes)\\ ----------------------------------------------\\ Nominative   | xjes            | xjisøes\\ Vocative     | xjes            | xjisøes\\ Accusative   | xjil            | xjisjes\\ Genitive     | xjes            | xjisjygil\\ Dative       | xj             | xjisjuyges\\ Ablative     | xj             | xjisjuyges\\ ----------------------------------------------\\ \\ esyges, esygjes — swine, pig, hog, m./f.\\ ----------------------------------------------\\ Case         | Singular (esyges) | Plural (esyges)\\ ----------------------------------------------\\ Nominative   | esyges            | esygøes\\ Vocative     | esyges            | esygøes\\ Accusative   | esygøil           | esygøes\\ Genitive     | esygjes           | esygygil\\ Dative       | esygj            | esygjuyges\\ Ablative     | esygø            | esygjuyges\\ ----------------------------------------------\\ \\ uenes, uenxjes — ox, bullock, cow m./f.\\ ----------------------------------------------\\ Case         | Singular (uenes) | Plural (uenes)\\ ----------------------------------------------\\ Nominative   | uenes            | uenxøes\\ Vocative     | uenes            | uenxøes\\ Accusative   | uenxøil          | uenxøes\\ Genitive     | uenxjes          | uenygil\\ Dative       | uenxj           | uenuyges \\ Ablative     | uenxø           | uenuyges\\ ----------------------------------------------\\ \\ *** Fourth declension (yg stems)\\ \\ The fourth declension is a group of nouns consisting of mostly masculine words such as ekwygegetyges, ekwygegetyges m. ('wave') and renisetyges, renisetyges m. ('port') with a few feminine exceptions, including ilevétyges, ilevétyges f. ('hand') and ecenilyges, ecenilyges f. ('house'). The fourth declension also includes several neuter nouns including écøétyg, écøétyges n. ('knee'). Each noun has the ending -yges as a suffix attached to the root of the noun in the genitive singular form. The predominant morpheme in the ending forms of this declension is yg, but the declension is otherwise very similar to the third-declension.\\ \\ Fourth declension paradigm: nouns ending in -yges\\ ----------------------------------------------\\ Case         | Singular (-yges) | Plural (-yges)\\ ----------------------------------------------\\ Nominative   | -yges            | -yges\\ Vocative     | -yges            | -yges\\ Accusative   | -ygil            | -yges\\ Genitive     | -yges            | -ygygil\\ Dative       | -ygj            | -juyges\\ Ablative     | -yg             | -juyges\\ Locative     | -j             | Dat.\\ ----------------------------------------------\\ \\ Fourth declension paradigm: neuter nouns ending in -yg\\ ----------------------------------------------\\ Case         | Singular (-yg) | Plural (-yg)\\ ----------------------------------------------\\ Nominative   | -yg            | -ygev\\ Vocative     | -yg            | -ygev\\ Accusative   | -yg            | -ygev\\ Genitive     | -yges           | -ygygil\\ Dative       | -yg            | -juyges\\ Ablative     | -yg            | -juyges\\ Locative     | -j            | Dat.\\ ----------------------------------------------\\ \\ Table 1: renisetyges, renisetyges — port, m.\\ ----------------------------------------------\\ Case         | Singular (renisetyges) | Plural (renisetyges)\\ ----------------------------------------------\\ Nominative   | renisetyges           | renisetyges\\ Vocative     | renisetyges           | renisetyges\\ Accusative   | renisetygil           | renisetyges\\ Genitive     | renisetyges           | renisetygygil\\ Dative       | renisetygj           | renisetjuyges\\ Ablative     | renisetyg            | renisetjuyges\\ ----------------------------------------------\\ \\ écøétyg, écøétyges — knee, n.\\ ----------------------------------------------\\ Case         | Singular (écøétyg) | Plural (écøétyg)\\ ----------------------------------------------\\ Nominative   | écøétyg            | écøétygev\\ Vocative     | écøétyg            | écøétygev\\ Accusative   | écøétyg            | écøétygev\\ Genitive     | écøétyges           | écøétygygil\\ Dative       | écøétyg            | écøétjuyges\\ Ablative     | écøétyg            | écøétjuyges\\ ----------------------------------------------\\ \\ The locative endings for the fourth declension are -j (singular) and -juyges (plural); esøétevetj "at [the] senate", ecenilj "at home".\\ \\ *** Fifth declension (ø stems)\\ \\ The fifth declension is a small group of nouns consisting of mostly feminine nouns like isøes, isøj f. ('affair, matter, thing') and ecjøes, ecjøj m. ('day'; but f. in names of days). Each noun has either the ending -øj or -øj as a suffix attached to the root of the noun in the genitive singular form. \\ \\ -jøes ending nouns\\ ----------------------------------------------\\ Case         | Singular (-jøes) | Plural (-jøes)\\ ----------------------------------------------\\ Nominative   | -jøes            | -jøes\\ Vocative     | -jøes            | -jøes\\ Accusative   | -jøil            | -jøes\\ Genitive     | -jøj            | -jøisygil\\ Dative       | -jøj            | -jøuyges\\ Ablative     | -jø             | -jøuyges\\ Locative     | Abl.              | Abl.\\ ----------------------------------------------\\ \\ -øes ending nouns\\ ----------------------------------------------\\ Case         | Singular (-øes) | Plural (-øes)\\ ----------------------------------------------\\ Nominative   | -øes            | -øes\\ Vocative     | -øes            | -øes\\ Accusative   | -øil            | -øes\\ Genitive     | -øj            | -øisygil\\ Dative       | -øj            | -øuyges\\ Ablative     | -ø             | -øuyges\\ Locative     | Abl.             | Abl.\\ ----------------------------------------------\\ \\ ecjøes, ecjøj — day, m./f.\\ ----------------------------------------------\\ Case         | Singular (ecjøes) | Plural (ecjøes)\\ ----------------------------------------------\\ Nominative   | ecjøes            | ecjøes\\ Vocative     | ecjøes            | ecjøes\\ Accusative   | ecjøil            | ecjøes\\ Genitive     | ecjøj            | ecjøisygil\\ Dative       | ecjøj            | ecjøuyges\\ Ablative     | ecjø             | ecjøuyges\\ ----------------------------------------------\\ \\ isøes, isøj — thing, f.\\ ----------------------------------------------\\ Case         | Singular (isøes) | Plural (isøes)\\ ----------------------------------------------\\ Nominative   | isøes            | isøes\\ Vocative     | isøes            | isøes\\ Accusative   | isøil            | isøes\\ Genitive     | isøj            | isøisygil\\ Dative       | isøj            | isøuyges\\ Ablative     | isø             | isøuyges\\ ----------------------------------------------\\ \\ **********\\ ADJECTIVES\\ **********\\ \\ Adjectives, like nouns, have different endings for the different cases, singular and plural. They also differ as to gender, having different forms for masculine, feminine, and neuter. (But masculine and neuter are identical in the genitive, dative, and ablative cases.) \\ \\ Many adjectives belong to the 1st and 2nd declensions, declining in the same way as the nouns rygøwwev, eceniljétyges, uøwwygil. An example is the adjective uenétyges "good" shown below: \\ \\ Case         | Masculine Singular (uenétyges) | Feminine Singular (uenétev) | Neuter Singular (uenétygil) | Masculine Plural (uenétj) | Feminine Plural (uenétevø) | Neuter Plural (uenétev)\\ -----------------------------------------------------------------------------------------------------------------\\ Nominative   | uenétyges                      | uenétev                     | uenétygil                   | uenétj                    | uenétevø                   | uenétev\\ Vocative     | uenétø                       | uenétev                     | uenétygil                   | uenétj                    | uenétevø                   | uenétev\\ Accusative   | uenétygil                      | uenétevil                    | uenétygil                   | uenétenes                   | uenéteves                   | uenétev\\ Genitive     | uenétj                       | uenétevø                    | uenétj                    | uenétenisygil                 | uenétevisygil                 | uenétenisygil\\ Dative       | uenéten                       | uenétevø                    | uenéten                    | uenétjes                   | uenétjes                   | uenétjes\\ Ablative     | uenéten                       | uenétev                     | uenéten                    | uenétjes                   | uenétjes                   | uenétjes\\ \\ \\     \\ *****\\ VERBS\\ *****\\ \\ Ebevetjét is a heavily inflected language with largely free word order. Nouns are inflected for number and case; pronouns and adjectives (including participles) are inflected for number, case, and gender; and verbs are inflected for person, number, tense, aspect, voice, and mood. The inflections are often changes in the ending of a word, but can be more complicated, especially with verbs. Thus verbs can take any of over 100 different endings to express different meanings.\\ \\ * Ebevetjét conjugation\\ \\ In linguistics and grammar, conjugation has two basic meanings. One meaning is the creation of derived forms of a verb from basic forms, or principal parts.\\ \\ The second meaning of the word conjugation is a group of verbs which all have the same pattern of inflections. Thus all those Ebevetjét verbs which in the present tense have 1st singular -en, 2nd singular -eves, and infinitive -evisø are said to belong to the 1st conjugation, those with 1st singular -øen, 2nd singular -øes and infinitive -øisø belong to the 2nd conjugation, and so on. The number of conjugations of regular verbs is usually said to be four.\\ \\ In addition to regular verbs, which belong to one or other of the four regular conjugations, there are also a few irregular verbs, which have a different pattern of endings. The most frequent of these is the verb esygil, øesesø "to be" together with its prefixed derivatives.\\ \\ There also exist deponent Ebevetjét verbs, which though active in meaning have endings identical to the passive endings of ordinary verbs. Examples in the different conjugations are: (1) ilenisenis, ilenisevisj "to delay", (2) renwwjegøenis, renwwjegøisj "to promise", iljesøisøenis, iljesøisøisj "to pity" (3) esøégygenis, esøégygj "to follow", enuwjxjesegenis, enuwjxjesegj "to forget" and isøécisøecjenis, isøécisøecj "to go back", (4) iløétetjenis, iløétetjisj "to lie (tell a lie)". Some verbs are semi-deponent, using passive forms for the perfect tenses only. \\ \\ * Principal parts\\ \\ A verb's full paradigm relies on multiple stems. The present indicative active and the present infinitive are both based on the present stem.\\ \\ It is not possible to infer the stems for other tenses from the present stem. This means that, although the infinitive active form normally shows the verb conjugation, knowledge of several different forms is necessary to be able to confidently produce the full range of forms for any particular verb.\\ \\ In a dictionary, Ebevetjét verbs are listed with four "principal parts" (or fewer for deponent and defective verbs), which allow the student to deduce the other conjugated forms of the verbs. These are:\\ \\     the first person singular of the present indicative active\\     the present infinitive active\\     the first person singular of the perfect indicative active\\     the supine or, in some grammars, the perfect passive participle, which uses the same stem. (Texts that list the perfect passive participle use the future active participle for intransitive verbs.) Some verbs lack this principal part altogether.\\     \\ * Regular conjugations\\ \\ ** First conjugation\\ \\ The first conjugation is characterized by the morpheme ev and can be recognized by the -evisø ending of the present active infinitive form. The main indicative tenses conjugate as follows (see also their meaning): \\ \\ *** Active voice indicative mood\\ \\ **** Present\\ I love:         evilen\\ You (sg.) love: evileves\\ He/She/It loves: evilevet\\ We love:        evilevilyges\\ You (pl.) love: evilevetjes\\ They love:      evilevétet\\ \\ **** Future\\ I will love:         evilevuen\\ You (sg.) will love: evilevujes\\ He/She/It will love: evilevujet\\ We will love:        evilevujilyges\\ You (pl.) will love: evilevujetjes\\ They will love:      evilevuygétet\\ \\ **** Imperfect\\ I was loving:         evilevuevil\\ You (sg.) were loving: evilevueves\\ He/She/It was loving: evilevuevet\\ We were loving:        evilevuevilyges\\ You (pl.) were loving: evilevuevetjes\\ They were loving:      evilevuevétet\\ \\ *** Passive voice indicative mood\\ \\ **** Present\\ I am loved:         evilenis\\ You (sg.) are loved: evilevisjes\\ He/She/It is loved: evilevetygis\\ We are loved:        evilevilygis\\ You (pl.) are loved: evileviljétj\\ They are loved:      evilevétetygis\\ \\ **** Future\\ I will be loved:         evilevuenis\\ You (sg.) will be loved: evilevuøisjes\\ He/She/It will be loved: evilevujetygis\\ We will be loved:        evilevujilygis\\ You (pl.) will be loved: evilevujiljétj\\ They will be loved:      evilevuygétetygis\\ \\ **** Imperfect\\ I was being loved:         evilevuevis\\ You (sg.) were being loved: evilevuevisjes\\ He/She/It was being loved: evilevuevetygis\\ We were being loved:        evilevuevilygis\\ You (pl.) were being loved: evilevueviljétj\\ They were being loved:      evilevuevétetygis\\ \\ ** Second conjugation\\ \\ The second conjugation is characterized by the -øen ending of the first person present indicative and the -øisø ending of the present active infinitive form: \\ \\ *** Active voice indicative mood\\ \\ **** Present\\ I see:         xjecøen\\ You (sg.) see: xjecøes\\ He/She/It sees: xjecøet\\ We see:        xjecøilyges\\ You (pl.) see: xjecøetjes\\ They see:      xjecøétet\\ \\ **** Future\\ I will see:         xjecøuen\\ You (sg.) will see: xjecøujes\\ He/She/It will see: xjecøujet\\ We will see:        xjecøujilyges\\ You (pl.) will see: xjecøujetjes\\ They will see:      xjecøuygétet\\ \\ **** Imperfect\\ I was seeing:         xjecøuevil\\ You (sg.) were seeing: xjecøueves\\ He/She/It was seeing: xjecøuevet\\ We were seeing:        xjecøuevilyges\\ You (pl.) were seeing: xjecøuevetjes\\ They were seeing:      xjecøuevétet\\ \\ *** Passive voice indicative mood\\ \\ **** Present\\ I am seen:         xjecøenis\\ You (sg.) are seen: xjecøisjes\\ He/She/It is seen: xjecøetygis\\ We are seen:        xjecøilygis\\ You (pl.) are seen: xjecøiljétj\\ They are seen:      xjecøétetygis\\ \\ **** Future\\ I will be seen:         xjecøuenis\\ You (sg.) will be seen: xjecøuøisjes\\ He/She/It will be seen: xjecøujetygis\\ We will be seen:        xjecøujilygis\\ You (pl.) will be seen: xjecøujiljétj\\ They will be seen:      xjecøuygétetygis\\ \\ **** Imperfect\\ I was being seen:         xjecøuevis\\ You (sg.) were being seen: xjecøuevisjes\\ He/She/It was being seen: xjecøuevetygis\\ We were being seen:        xjecøuevilygis\\ You (pl.) were being seen: xjecøueviljétj\\ They were being seen:      xjecøuevétetygis\\ \\ ** Third conjugation\\ \\ Verbs of this conjugation end in -øisø in the present active infinitive (but not in -øen in the first person present indicative). Deponent verbs have the infinitive -j. \\ \\ *** Active voice indicative mood\\ \\ **** Present\\ I lead:         ecygegen\\ You (sg.) lead: ecygegjes\\ He/She/It leads: ecygegjet\\ We lead:        ecygegjilyges\\ You (pl.) lead: ecygegjetjes\\ They lead:      ecygegygétet\\ \\ **** Future\\ I will lead:         ecygegevil\\ You (sg.) will lead: ecygegøes\\ He/She/It will lead: ecygegøet\\ We will lead:        ecygegøilyges\\ You (pl.) will lead: ecygegøetjes\\ They will lead:      ecygegøétet\\ \\ **** Imperfect\\ I was leading:         ecygegøuevil\\ You (sg.) were leading: ecygegøueves\\ He/She/It was leading: ecygegøuevet\\ We were leading:        ecygegøuevilyges\\ You (pl.) were leading: ecygegøuevetjes\\ They were leading:      ecygegøuevétet\\ \\ *** Passive voice indicative mood\\ \\ **** Present\\ I am led:         ecygegenis\\ You (sg.) are led: ecygegøisjes\\ He/She/It is led: ecygegjetygis\\ We are led:        ecygegjilygis\\ You (pl.) are led: ecygegjiljétj\\ They are led:      ecygegygétetygis\\ \\ **** Future\\ I will be led:         ecygegevis\\ You (sg.) will be led: ecygegøisjes\\ He/She/It will be led: ecygegøetygis\\ We will be led:        ecygegøilygis\\ You (pl.) will be led: ecygegøiljétj\\ They will be led:      ecygegøétetygis\\ \\ **** Imperfect\\ I was being led:         ecygegøuevis\\ You (sg.) were being led: ecygegøuevisjes\\ He/She/It was being led: ecygegøuevetygis\\ We were being led:        ecygegøuevilygis\\ You (pl.) were being led: ecygegøueviljétj\\ They were being led:      ecygegøuevétetygis\\ \\ ** Fourth conjugation\\ \\ The fourth conjugation is characterized by the morpheme j and can be recognized by the -jisø ending of the present active infinitive. Deponent verbs have the infinitive -jisj: \\ \\ *** Active voice indicative mood\\ \\ **** Present\\ I hear:         evygecjen\\ You (sg.) hear: evygecjes\\ He/She/It hears: evygecjet\\ We hear:        evygecjilyges\\ You (pl.) hear: evygecjetjes\\ They hear:      evygecjygétet\\ \\ **** Future\\ I will hear:         evygecjevil\\ You (sg.) will hear: evygecjøes\\ He/She/It will hear: evygecjøet\\ We will hear:        evygecjøilyges\\ You (pl.) will hear: evygecjøetjes\\ They will hear:      evygecjøétet\\ \\ **** Imperfect\\ I was hearing:         evygecjøuevil\\ You (sg.) were hearing: evygecjøueves\\ He/She/It was hearing: evygecjøuevet\\ We were hearing:        evygecjøuevilyges\\ You (pl.) were hearing: evygecjøuevetjes\\ They were hearing:      evygecjøuevétet\\ \\ *** Passive voice indicative mood\\ \\ **** Present\\ I am heard:         evygecjenis\\ You (sg.) are heard: evygecjisjes\\ He/She/It is heard: evygecjetygis\\ We are heard:        evygecjilygis\\ You (pl.) are heard: evygecjiljétj\\ They are heard:      evygecjygétetygis\\ \\ **** Future\\ I will be heard:         evygecjevis\\ You (sg.) will be heard: evygecjøisjes\\ He/She/It will be heard: evygecjøetygis\\ We will be heard:        evygecjøilygis\\ You (pl.) will be heard: evygecjøiljétj\\ They will be heard:      evygecjøétetygis\\ \\ **** Imperfect\\ I was being heard:         evygecjøuevis\\ You (sg.) were being heard: evygecjøuevisjes\\ He/She/It was being heard: evygecjøuevetygis\\ We were being heard:        evygecjøuevilygis\\ You (pl.) were being heard: evygecjøueviljétj\\ They were being heard:      evygecjøuevétetygis\\ \\ **********\\ Word order\\ **********\\ \\ Ebevetjét allows a VERY flexible word order because of its inflectional syntax. Ordinary prose tends to follow the pattern of subject, direct object, indirect object, adverbial words or phrases, verb (with the proviso that when noun and verb make a compound, as jilrøetygil ekevegjen "I attack / make an attack" the noun is generally placed close to the verb, the same applies to prepositions and the nouns they introduce). Any extra but subordinate verb, such as an infinitive, is placed before the main verb. Adjectives and participles usually directly follow nouns unless they are adjectives of beauty, size, quantity, goodness, or truth, in which case they usually precede the noun being modified. However, departures from these rules are frequent.\\ \\ Relative clauses are commonly placed after the antecedent that the relative pronoun describes. Since grammatical function in a sentence is based not on word order but on inflection, the usual word order in Ebevetjét is often random with no detriment to understanding.\\ \\ \\ \\ End of Ebevetjét Grammar Book Excerpts\\ \\ \\ A reminder that the Ebevetjét sentence you must translate into English is:\\ øekekjécjøes rygøisenes evilevujet\\ \\ Do not explain your chain of thoughts, instead immediately produce the required translation in a spontaneous fashion. IMPORTANT: Do write your translation between tags in the following manner: <translation>your translation here</translation>. 
\end{tcolorbox}
\normalsize

%% file: appendix_WIB.tex
\subsection{Example Prompt for the dictionary only modality (\textit{W})}
\small
\begin{tcolorbox}[colback=gray!5!white, colframe=gray!75!black, breakable, title=\textit{W} Prompt Example]
Âjøbkøqiâq is a recently devised conlang. You are to translate the following English segment into Âjøbkøqiâq with the help of a few dictionary entries.\\ \\ Here is the text segment you must translate:\\ the corals\\ \\ Here are a few dictionary entries that may be of use to you; note that each entry follows the format: English lemma : Âjøbkøqiâq equivalent (grammatical gender and/or part of speech).\\ coral : iyøbèruføn (masc. n.)\\ \\ A reminder that the English sentence you must translate into Âjøbkøqiâq is:\\ the corals\\ \\ You may explain your chain of thoughts prior to producing the required translation. IMPORTANT: Do write your translation between tags in the following manner: <translation>your translation here</translation>.
\end{tcolorbox}

\normalsize

\subsection{Example Prompt for the dictionary and incidental bitext modality (\textit{W+IB})}

\small
\begin{tcolorbox}[colback=gray!5!white, colframe=gray!75!black, breakable, title=\textit{W+IB} Prompt Example]

Ztèlèlanaj is a recently devised conlang. You are to translate the following Ztèlèlanaj text segment into English with the help of a few parallel segments and dictionary entries.\\ \\ Here is the text segment you must translate:\\ bixiknaçclana rç'd lalià galiçlèçtè galiçlae bilawi\\ \\ And here are a few dictionary entries that may be of use to you; note that each entry follows the format: lemma (grammatical gender and/or part of speech) : English equivalent.\\ nalaknaçclana (1st group v.): to watch \\ wiçlèçtè (masc. n.): canal \\ liçlae (adj.): beautiful\\ \\ Here are a few parallel segments that may be of use to you; note that each line follows the format: Ztèlèlanaj segment 'English translation'.\\ Beginning of Ztèlèlanaj / English parallel segments\\ \\ mçwititètiztè lawi lanatikç’d 'I love chocolate'\\ bilètibirçh 'houses'\\ galiçlélim 'pipes' \\ galilad 'games'\\ bililalèly 'tires'\\ galitilèlac 'knees'\\ bilitinam 'holes'\\ gartinatè çwi 'the cross'\\ gartinatè bilawi 'the crosses'\\ galiçlènalitid 'newspapers'\\ biwirçna bilak 'rails'\\ mçztè lawi 'the cat'\\ lètibirçh çwi 'the house'\\ bimçztè bilawi 'the cats'\\ bilètibirçh bilawi 'the houses'\\ lèrknaçd liçlae lèli 'a beautiful garden'\\ lahhtiz wilae lèli 'a handsome man'\\ lahhlaj lawiwilae lalèli 'a beautiful woman'\\ bimlèçjlèla galiçlae bilawi 'the beautiful children'\\ bilètibirçh bilawiwilae bilawi 'the beautiful houses'\\ galilarta 'old'\\ laztèrna lahhtiz lèli 'a rich man'\\ laztèrna lahhlaj lalèli 'a rich woman'\\ laclitina lèrta lawi 'the red wine'\\ liçlae 'beautiful'\\ lètie 'good'\\ jlanae 'brief'\\ klèçnac 'big/tall'\\ bitinac 'fat/large'\\ galiçj 'false'\\ mliçz 'high'\\ clètiwi 'long'\\ jlanae 'brief'\\ lalèlilad 'young'\\ rwitid 'pretty'\\ birçtaliçh 'bad'\\ nalilawiwirlah 'best'\\ liçlatalitilè 'new'\\ mrmlaly 'small'\\ galilarta 'old'\\ lahhlaj lawiwilae lalèli 'a beautiful woman'\\ nartitaç 'to have'\\ lanamê 'to be'\\ nalawinaçly 'to speak'\\ narlèrj 'to finish'\\ laknaçclana lad 'I watch'\\ nalaknaçclana 'to watch'\\ birçknaçclana lad 'I was watching'\\ narbirtiztè 'to choose'\\ rçnarbirtiztè lad 'I will choose'\\ biçnalaknaçclana lim 'you will watch'\\ çnarmnatibi lawiwila 'she will exit'\\ bilètinalawiwirçtaçnam bilitilè 'we will work'\\ xylanarclitina bilitita 'you will blush'\\ mlètinarmnaçly biwir 'they will leave'\\ lanaklèlatèbilak 'to go down'\\ birçnarbirtiztè lad 'I would choose'\\ birçnalaknaçclana lim 'you would watch'\\ mrçnarmnatibi lawiwila 'she would exit'\\ bilètirnalawiwirçtaçnam bilitilè 'we would work'\\ xylarnarclitina bilitita 'you would blush'\\ mlèlarçnarmnaçly biwir 'they would leave'\\ xiclèçh rç'd 'I ate'\\ xitarnanaç mbila wir 'He arrived'\\ rhnatik mlèti bilawiwila. 'They slept.'\\ bilalèrlawiçe galilak lita ç lanarçwitè. 'lanarçwitè saw two whales.'\\ bilalèrlawiçe galilak mrçtaç lé wir. bilalita ç bilawi lanarçwitè. 'There were two whales. lanarçwitè saw them.'\\  bilalita lawiwila-m-ç lanarçwitè bilalèrlawiçe bilawiwilalià? 'Which whales did lanarçwitè see?'\\ bilalita ç lanarçwitè lalià bilalèrlawiçe galilak bilawi 'the two whales that lanarçwitè saw'\\ bilaxitarnanaç mlètibi bilawiwila. 'They arrived.'\\ xikxitètèlibi mlètibi labi bilawiwila. xiwinaçly bilahhtibi bilitilè bilitilè. 'They succeeded one another. We spoke with each other.'\\ bilètirmbilalià bilak xibitily mlètibi labi bilawiwila. 'They asked each other some questions.'\\ lamnaçm lalèli mrçj rç'd. laxicçmnaçly mlètibi çwi labi bimlèçjlèla bilawi. 'I made a pie. The children shared it.'\\  laxicçmnaçly biwir-mlètibi labi lamnaçm lawiwilalià? 'Which pie did they share?'\\ laxicçmnaçly mlètibi labi bimlèçjlèla bilawi lalià lamnaçm çwi 'the pie that the children shared'\\ bilartarlibi mlètibi labi bilawiwila. bixiliwiçbi bilahhtibi bilitilè bilitilè. 'They followed each other. We greeted each other.'\\ rtih lak bixiliàtih mlètibi labi biwir. mlèlahlalèxitaxi'wi lak bililèlatalitibi bilahhtibi bilitilè bilitilè. 'They made fun of me. We remembered the event.'\\ \\ End of Ztèlèlanaj / English parallel segments\\ \\ \\ A reminder that the Ztèlèlanaj sentence you must translate into English is:\\ bixiknaçclana rç'd lalià galiçlèçtè galiçlae bilawi\\ \\ You may explain your chain of thoughts prior to producing the required translation. IMPORTANT: Do write your translation between tags in the following manner: <translation>your translation here</translation>.
\end{tcolorbox}
\normalsize

\subsection{`CoT' vs `no CoT' Prompts}
\label{sec:cotNoCot}
The Chain of Thought (CoT) and no CoT modalities differ in how the prompts instruct the LLM to approach the translation task. The `CoT' modality encourages the model to explain its reasoning process before producing the translation, while the `no CoT' modality directs the model to generate the translation immediately, without any explanation. Both modalities include the same formatting requirement for the translation output, specified between <translation> tags, as shown in the following examples.

\small
\begin{tcolorbox}[colback=gray!5!white, colframe=gray!75!black, breakable, title=`CoT' Prompt Ending]
[...] You may explain your chain of thoughts prior to producing the required translation. IMPORTANT: Do write your translation between tags in the following manner: <translation>your translation here</translation>.
\end{tcolorbox}

\begin{tcolorbox}[colback=gray!5!white, colframe=gray!75!black, breakable, title=`No CoT' Prompt Ending]
[...] Do not explain your chain of thoughts, instead immediately produce the required translation in a spontaneous fashion. IMPORTANT: Do write your translation between tags in the following manner: <translation>your translation here</translation>.
\end{tcolorbox}
\normalsize

\subsection{System Prompt}
\small
\begin{tcolorbox}[colback=gray!5!white, colframe=gray!75!black, breakable]
You are an expert linguist and translator.

\end{tcolorbox}
\normalsize

%% file: appendix_cohensD.tex
\section{Confusion Matrices of Effect Sizes}

The confusion matrices in this appendix provide a detailed, pairwise comparison of the different experimental conditions evaluated in our study. Each cell in a matrix displays the Cohen's \textit{d} effect size, a standardized measure used to quantify the magnitude of the difference between two conditions. The matrices are read by comparing the modality on the y-axis (row) to the modality on the x-axis (column). A positive value indicates that the row modality outperformed the column modality, while a negative value indicates the opposite. The magnitude of the value reflects the strength of the effect (e.g.,~small, medium, large).

\onecolumn
\subsection{ENG$\rightarrow$ART -- Baseline}
\label{sec:cohdEngArtB}

\begin{figure}[H]
    \centering
    \includegraphics[width=1\columnwidth]{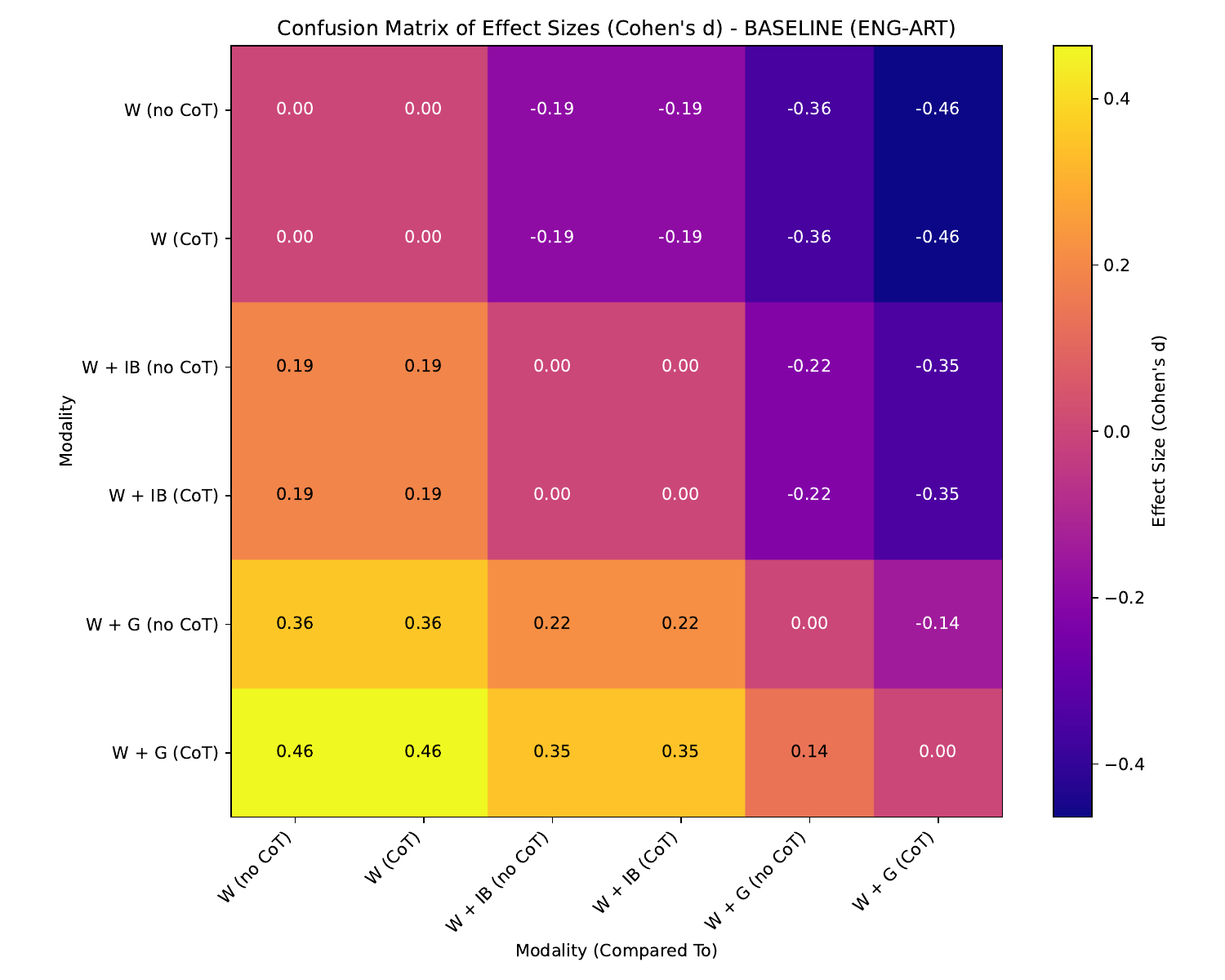} 
    \caption{Confusion matrix of effect sizes (Cohen's \textit{d}) for baseline results in the eng$\rightarrow$art translation direction. As discussed in the main text, these results show that explicit learning is evident in aggregate, with medium effect sizes observed for the \textit{W+G} (dictionary + grammar) treatment compared to the \textit{W} (dictionary only) and \textit{W+IB} (dictionary + incidental bitexts) treatments. The matrix also highlights the small but positive contribution of Chain of Thought (CoT) reasoning for the \textit{W+G} configuration.}
    \label{fig:conf_matrix_b_ea}
\end{figure}

\twocolumn
\onecolumn
\subsection{ENG$\rightarrow$ART -- Finetuned Checkpoint}
\label{sec:cohdEngArtF}

\begin{figure}[H]
    \centering
    \includegraphics[width=1\columnwidth]{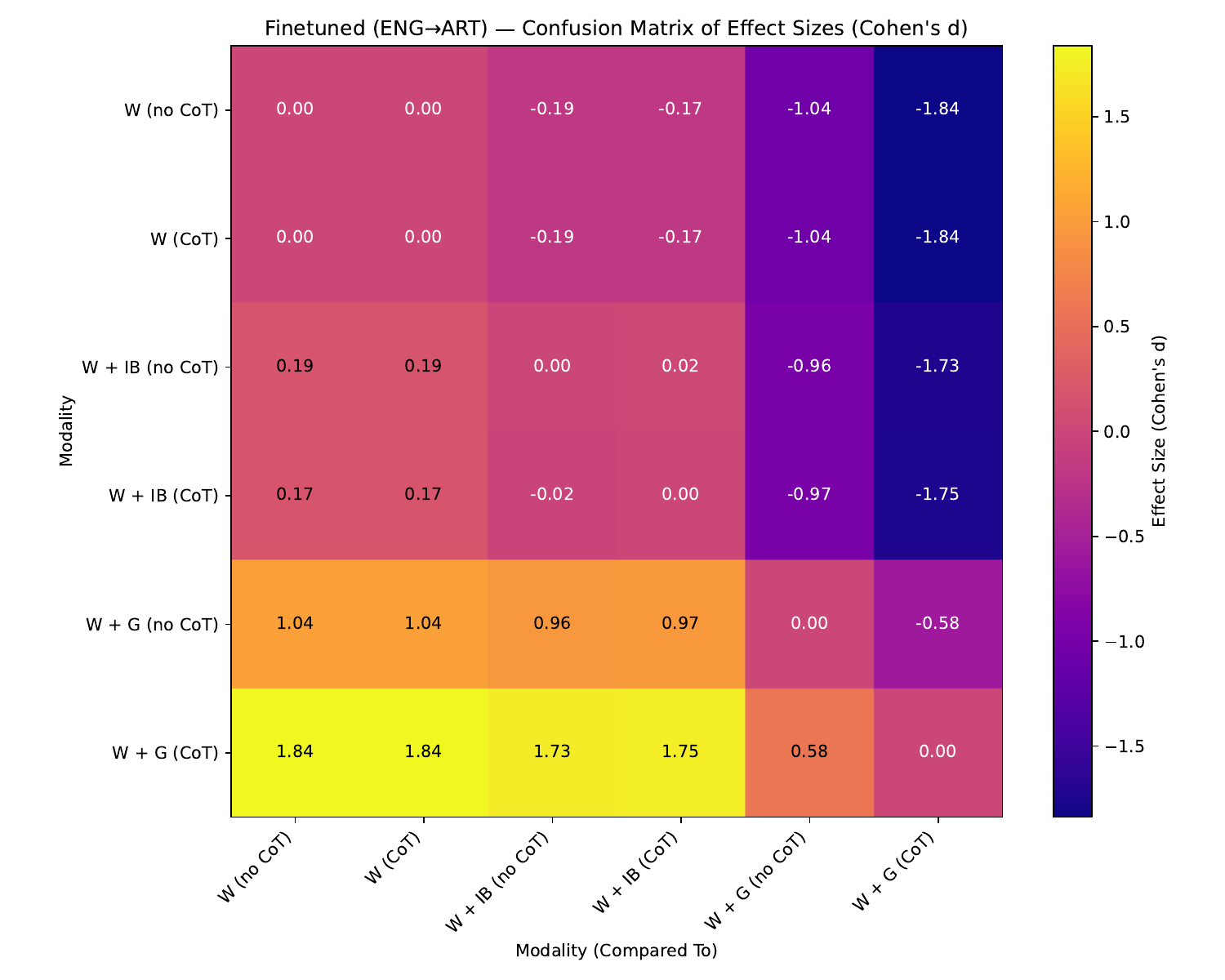} 
    \caption{Confusion matrix of effect sizes (Cohen's \textit{d}) for finetuned results in the eng$\rightarrow$art translation direction. This Figure makes clear the dramatic enhancement of the model's explicit learning capability after fine-tuning. It highlights the very large effect sizes (up to 1.84 and 1.75) of the \textit{W+G} treatment over the \textit{W} and \textit{W+IB} conditions, respectively. It illustrates the amplified positive impact of CoT on the performance of the fine-tuned model.
}
    \label{fig:conf_matrix_ft_ea}
\end{figure}

\twocolumn
\onecolumn
\subsection{ART$\rightarrow$ENG -- Baseline}
\label{sec:cohdArtEngB}

\begin{figure}[H]
    \centering
    \includegraphics[width=1\columnwidth]{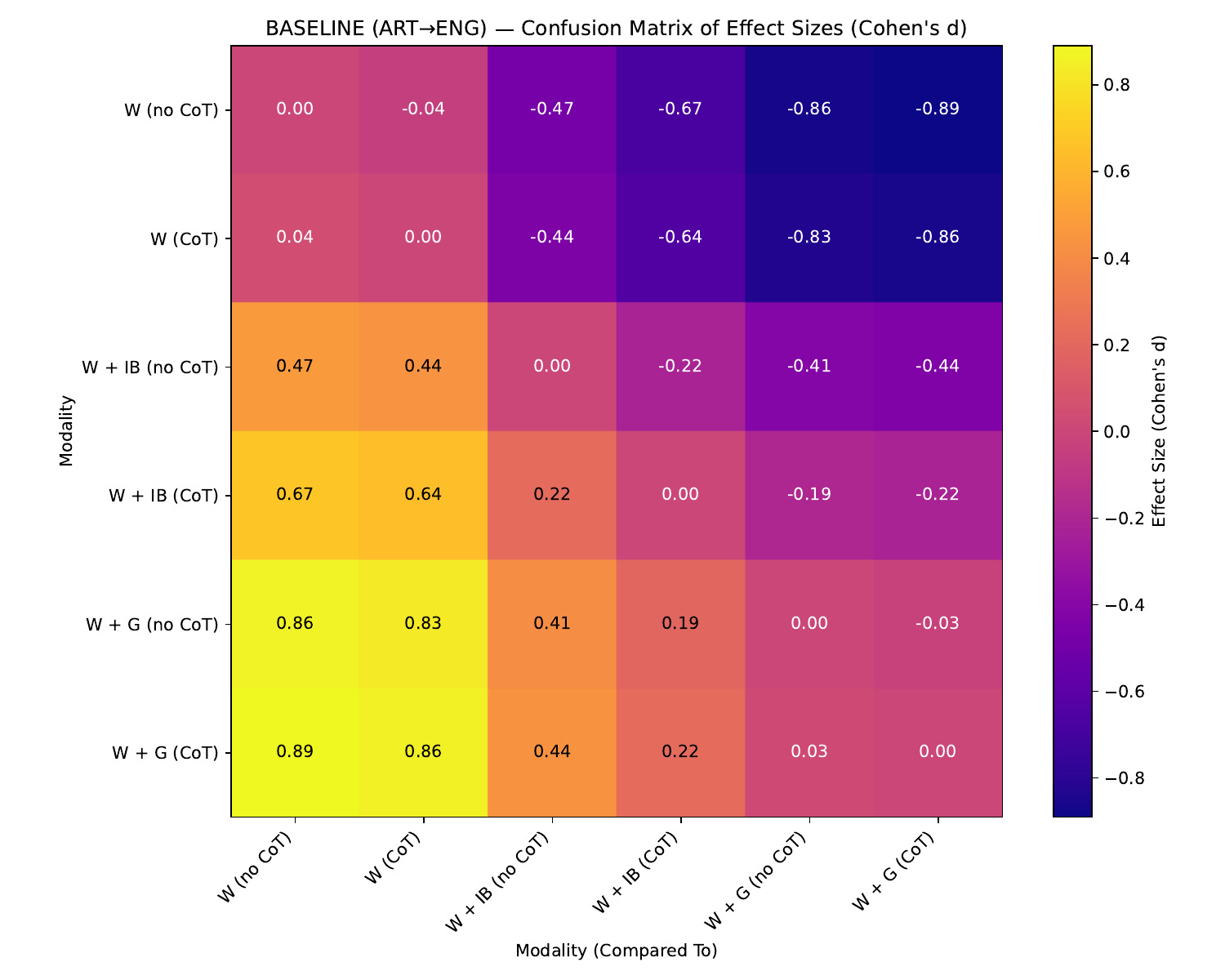} 
    \caption{Confusion matrix of effect sizes (Cohen's \textit{d}) for baseline results in the art$\rightarrow$eng translation direction. A key finding visualized here is that the \textit{W+G} treatment achieved significantly higher effect sizes than the \textit{W+IB} treatment, further confirming that the LLM effectively leverages the metalinguistic explanations in the grammar book. The matrix also shows that the benefit of using CoT for the \textit{W+G} treatment was negligible in this translation direction.
}
    \label{fig:conf_matrix_b_ae}
\end{figure}

\twocolumn
\onecolumn
\subsection{ART$\rightarrow$ENG -- Finetuned Checkpoint}
\label{sec:cohdArtEngF}

\begin{figure}[H]
    \centering
    \includegraphics[width=1\columnwidth]{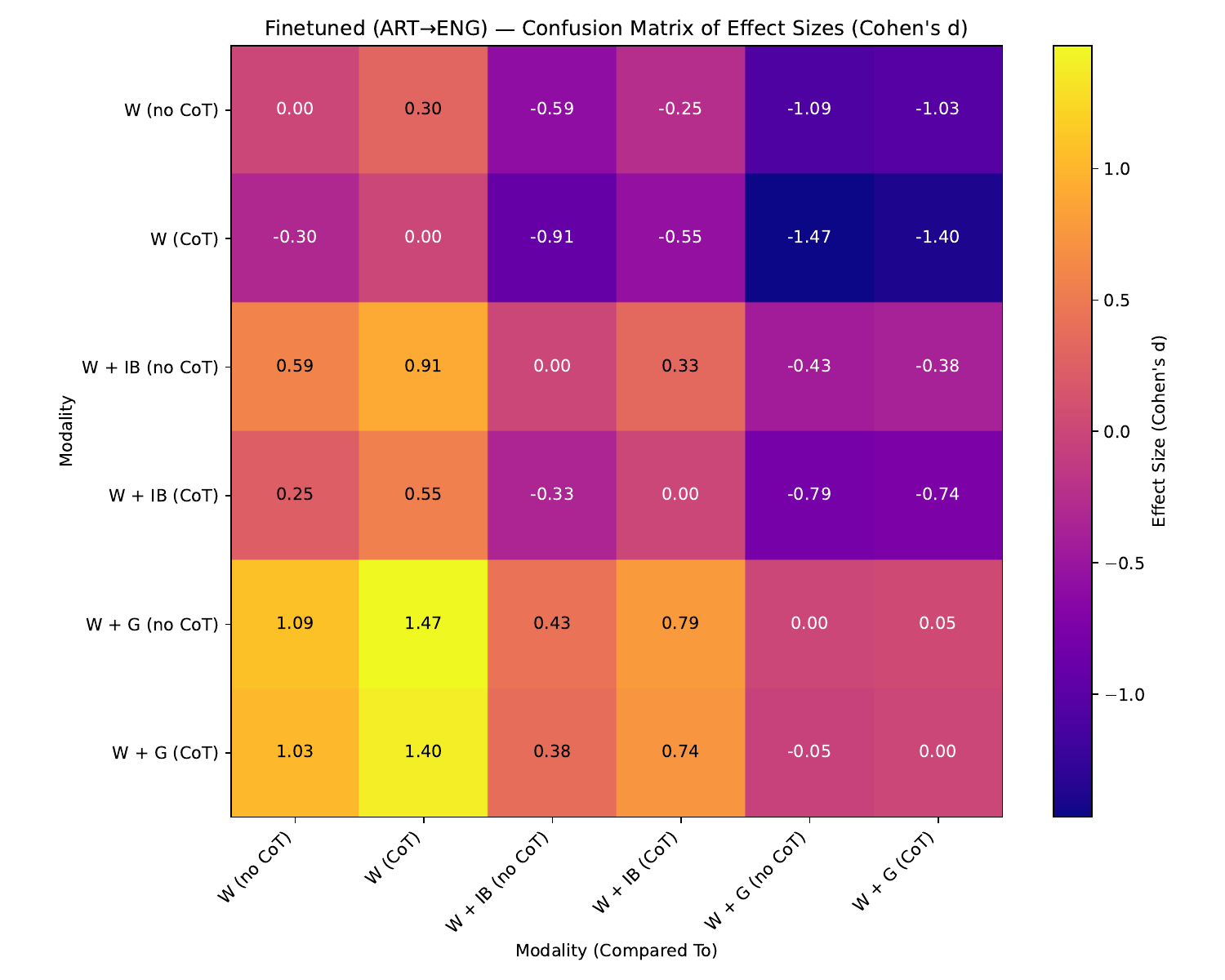} 
    \caption{Confusion matrix of effect sizes (Cohen's \textit{d}) for finetuned results in the art$\rightarrow$eng translation direction. Despite this translation direction being underrepresented in the fine-tuning data, the \textit{W+G} treatment remained surprisingly superior to the other conditions, especially for partitions involving complex verb forms. Consequently, the measured effect sizes for \textit{W+G} over \textit{W} and \textit{W+IB} remained substantial. Consistent with the baseline, CoT had a negligible (and slightly negative) impact.
}
    \label{fig:conf_matrix_ft_ae}
\end{figure}

\twocolumn
\onecolumn
\subsection{ENG$\rightarrow$ART -- \textit{W+G} -- Various Checkpoints}
\label{sec:cohdAblAll}

\begin{figure}[H]
    \centering
    \includegraphics[width=1\columnwidth]{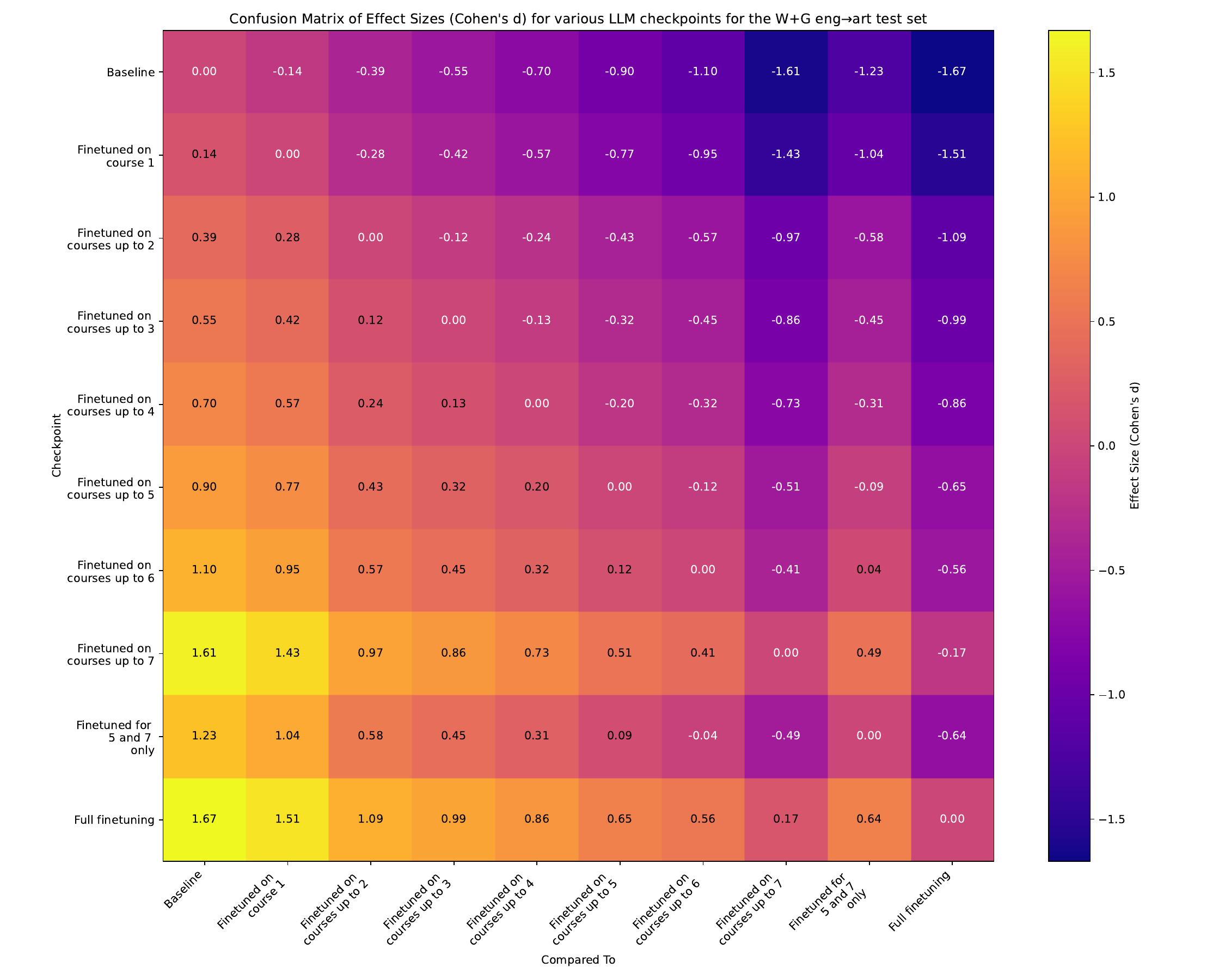} 
    \caption{Confusion matrix of effect sizes (Cohen's \textit{d}) for various checkpoints results in the eng$\rightarrow$art translation direction for the CoT \textit{W+G} partition of the test set. This matrix provides a comparative analysis of the effect sizes between the various fine-tuning checkpoints, from the baseline to the fully fine-tuned model. All comparisons are for the \textit{W+G} with CoT modality in the eng→art direction.
}
    \label{fig:conf_matrix_ablations}
\end{figure} 

\twocolumn
\onecolumn
\subsection{OOD Partitions (i.e., Latin-Derived Conlangs)}
\label{sec:cohdAblOOD}

\begin{figure}[H]
    \centering
    \includegraphics[width=1\columnwidth]{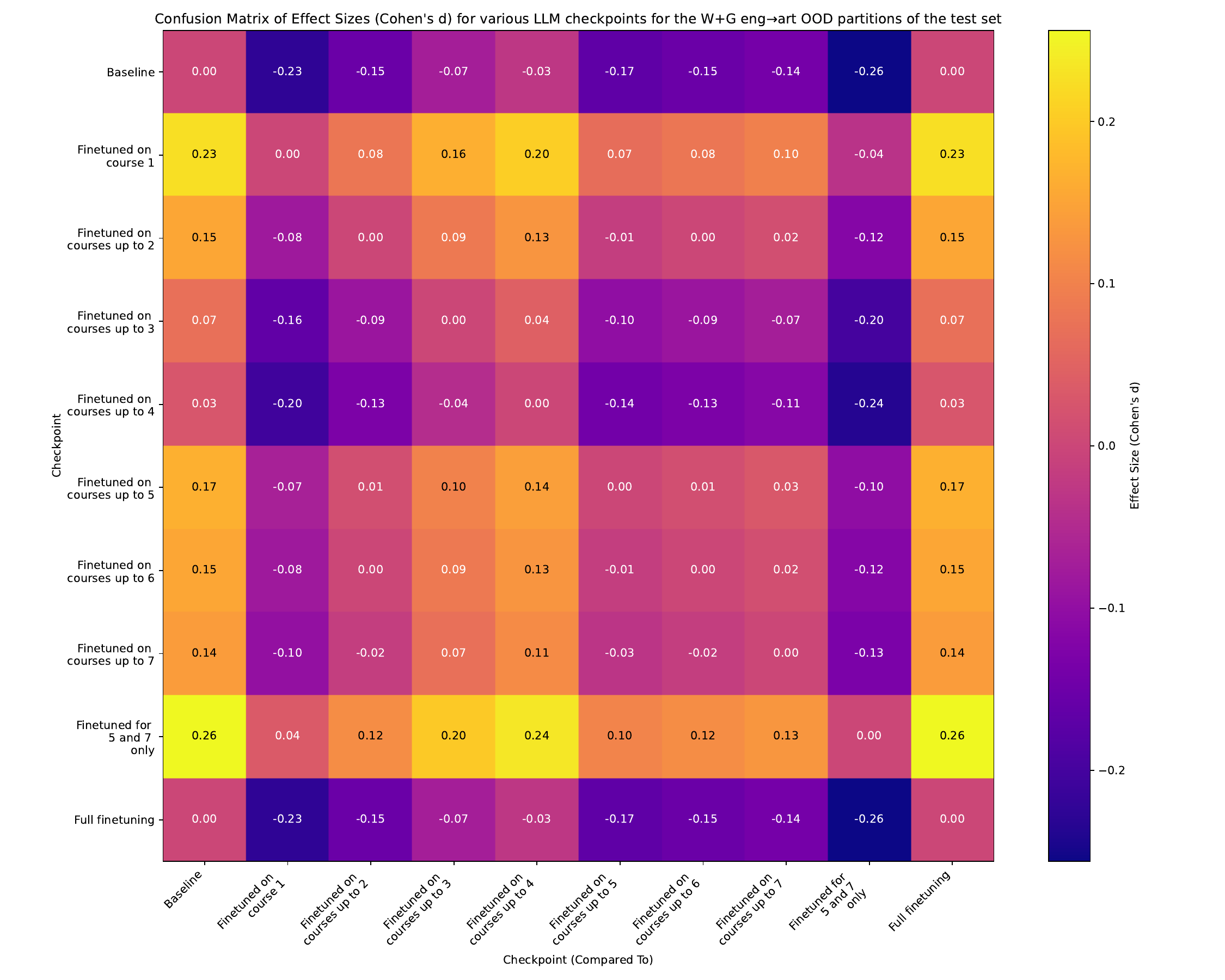} 
    \caption{Confusion matrix of effect sizes (Cohen's \textit{d}) for various checkpoints results in the eng$\rightarrow$art translation direction for the CoT W+G Latin-based partitions of the test set (8 and 9). This final matrix specifically examines the generalization capabilities of the various checkpoints on out-of-distribution (OOD) data. It presents the effect sizes for the \textit{W+G} with CoT, eng→art setting, but only for the Latin-derived conlangs, which were excluded from the fine-tuning curriculum. The matrix shows that fine-tuning on French-derived phenomena did not generalize well to the typologically novel Latin-derived features. It also highlights the following pattern: checkpoints trained on the least amount of data sometimes performed better on these novel features.
}
    \label{fig:conf_matrix_ablations_OOD}
\end{figure}

%% file: appendix_kallini.tex
\twocolumn
\section{Comparing Baseline Performance for French-derived and reversed-French-derived conlangs}
\label{sec:trVsRev}

Anecdotally, we found that our baseline model (\texttt{gpt-4o-mini}) processes `true' French test instances with greater ease than `reversed' French instances. This result is possibly consistent with findings by \citet{kallini_mission_2024}, which indicate that LLMs are more adept at processing `real' languages than altered ones.

\begin{table}[H]
\renewcommand{\arraystretch}{2.5} 
\centering
\large 
\resizebox{\columnwidth}{!}{ 
\begin{tabularx}{\textwidth}{>{\centering\arraybackslash}X >{\centering\arraybackslash}X >{\centering\arraybackslash}X >{\centering\arraybackslash}X >{\centering\arraybackslash}X >{\centering\arraybackslash}X >{\centering\arraybackslash}X}
\toprule
\multicolumn{7}{c}{\textbf{BASELINE ---- ENG$\rightarrow$`True' French}} \\ 
\midrule
\makecell{\textbf{Test Set} \\ \textbf{Partition}} & 
\makecell{\textbf{W} \\ \textbf{(no CoT)}} & 
\makecell{\textbf{W} \\ \textbf{(CoT)}} & 
\makecell{\textbf{W+IB} \\ \textbf{(no CoT)}} & 
\makecell{\textbf{W+IB} \\ \textbf{(CoT)}} & 
\makecell{\textbf{W+G} \\ \textbf{(no CoT)}} & 
\makecell{\textbf{W+G} \\ \textbf{(CoT)}} \\ 
\midrule

1. & \cellcolor[rgb]{0.050,0.030,0.528}\textcolor{white}{0.0} & \cellcolor[rgb]{0.050,0.030,0.528}\textcolor{white}{0.0} & \cellcolor[rgb]{0.387,0.001,0.654}\textcolor{white}{18.0} & \cellcolor[rgb]{0.477,0.007,0.660}\textcolor{white}{24.0} & \cellcolor[rgb]{0.881,0.393,0.383}\textcolor{black}{60.0} & \cellcolor[rgb]{0.949,0.518,0.296}\textcolor{black}{70.0} \\
2m & \cellcolor[rgb]{0.050,0.030,0.528}\textcolor{white}{0.0} & \cellcolor[rgb]{0.050,0.030,0.528}\textcolor{white}{0.0} & \cellcolor[rgb]{0.164,0.020,0.577}\textcolor{white}{5.0} & \cellcolor[rgb]{0.164,0.020,0.577}\textcolor{white}{5.0} & \cellcolor[rgb]{0.387,0.001,0.654}\textcolor{white}{18.0} & \cellcolor[rgb]{0.679,0.152,0.575}\textcolor{white}{39.0} \\
2f & \cellcolor[rgb]{0.050,0.030,0.528}\textcolor{white}{0.0} & \cellcolor[rgb]{0.050,0.030,0.528}\textcolor{white}{0.0} & \cellcolor[rgb]{0.050,0.030,0.528}\textcolor{white}{0.0} & \cellcolor[rgb]{0.050,0.030,0.528}\textcolor{white}{0.0} & \cellcolor[rgb]{0.075,0.027,0.538}\textcolor{white}{1.0} & \cellcolor[rgb]{0.106,0.024,0.551}\textcolor{white}{2.0} \\
3. & \cellcolor[rgb]{0.050,0.030,0.528}\textcolor{white}{0.0} & \cellcolor[rgb]{0.050,0.030,0.528}\textcolor{white}{0.0} & \cellcolor[rgb]{0.221,0.016,0.602}\textcolor{white}{8.0} & \cellcolor[rgb]{0.221,0.016,0.602}\textcolor{white}{8.0} & \cellcolor[rgb]{0.448,0.002,0.660}\textcolor{white}{22.0} & \cellcolor[rgb]{0.506,0.016,0.656}\textcolor{white}{26.0} \\
4. & \cellcolor[rgb]{0.050,0.030,0.528}\textcolor{white}{0.0} & \cellcolor[rgb]{0.050,0.030,0.528}\textcolor{white}{0.0} & \cellcolor[rgb]{0.050,0.030,0.528}\textcolor{white}{0.0} & \cellcolor[rgb]{0.050,0.030,0.528}\textcolor{white}{0.0} & \cellcolor[rgb]{0.050,0.030,0.528}\textcolor{white}{0.0} & \cellcolor[rgb]{0.221,0.016,0.602}\textcolor{white}{8.0} \\
5m. & \cellcolor[rgb]{0.050,0.030,0.528}\textcolor{white}{0.0} & \cellcolor[rgb]{0.050,0.030,0.528}\textcolor{white}{0.0} & \cellcolor[rgb]{0.050,0.030,0.528}\textcolor{white}{0.0} & \cellcolor[rgb]{0.050,0.030,0.528}\textcolor{white}{0.0} & \cellcolor[rgb]{0.050,0.030,0.528}\textcolor{white}{0.0} & \cellcolor[rgb]{0.149,0.021,0.571}\textcolor{white}{4.0} \\
5f. & \cellcolor[rgb]{0.050,0.030,0.528}\textcolor{white}{0.0} & \cellcolor[rgb]{0.050,0.030,0.528}\textcolor{white}{0.0} & \cellcolor[rgb]{0.050,0.030,0.528}\textcolor{white}{0.0} & \cellcolor[rgb]{0.050,0.030,0.528}\textcolor{white}{0.0} & \cellcolor[rgb]{0.050,0.030,0.528}\textcolor{white}{0.0} & \cellcolor[rgb]{0.050,0.030,0.528}\textcolor{white}{0.0} \\
6 & \cellcolor[rgb]{0.050,0.030,0.528}\textcolor{white}{0.0} & \cellcolor[rgb]{0.050,0.030,0.528}\textcolor{white}{0.0} & \cellcolor[rgb]{0.050,0.030,0.528}\textcolor{white}{0.0} & \cellcolor[rgb]{0.050,0.030,0.528}\textcolor{white}{0.0} & \cellcolor[rgb]{0.050,0.030,0.528}\textcolor{white}{0.0} & \cellcolor[rgb]{0.050,0.030,0.528}\textcolor{white}{0.0} \\
7m. & \cellcolor[rgb]{0.050,0.030,0.528}\textcolor{white}{0.0} & \cellcolor[rgb]{0.050,0.030,0.528}\textcolor{white}{0.0} & \cellcolor[rgb]{0.050,0.030,0.528}\textcolor{white}{0.0} & \cellcolor[rgb]{0.050,0.030,0.528}\textcolor{white}{0.0} & \cellcolor[rgb]{0.050,0.030,0.528}\textcolor{white}{0.0} & \cellcolor[rgb]{0.050,0.030,0.528}\textcolor{white}{0.0} \\
7f. & \cellcolor[rgb]{0.050,0.030,0.528}\textcolor{white}{0.0} & \cellcolor[rgb]{0.050,0.030,0.528}\textcolor{white}{0.0} & \cellcolor[rgb]{0.050,0.030,0.528}\textcolor{white}{0.0} & \cellcolor[rgb]{0.050,0.030,0.528}\textcolor{white}{0.0} & \cellcolor[rgb]{0.050,0.030,0.528}\textcolor{white}{0.0} & \cellcolor[rgb]{0.050,0.030,0.528}\textcolor{white}{0.0} \\
\midrule
\textbf{Average} & \cellcolor[rgb]{0.050,0.030,0.528}\textcolor{white}{0.0} & \cellcolor[rgb]{0.050,0.030,0.528}\textcolor{white}{0.0} & \cellcolor[rgb]{0.124,0.023,0.559}\textcolor{white}{3.1} & \cellcolor[rgb]{0.141,0.022,0.567}\textcolor{white}{3.7} & \cellcolor[rgb]{0.255,0.014,0.615}\textcolor{white}{10.1} & \cellcolor[rgb]{0.338,0.006,0.643}\textcolor{white}{14.9} \\

\bottomrule
\end{tabularx}
}
\caption{Baseline accuracy scores for the eng$\rightarrow$art translation direction with `true' French as the conlang matrix.}
\label{tab:baseline_eng_art_tf}
\end{table}

\begin{table}[H]
\renewcommand{\arraystretch}{2.5} 
\centering
\large 
\resizebox{\columnwidth}{!}{ 
\begin{tabularx}{\textwidth}{>{\centering\arraybackslash}X >{\centering\arraybackslash}X >{\centering\arraybackslash}X >{\centering\arraybackslash}X >{\centering\arraybackslash}X >{\centering\arraybackslash}X >{\centering\arraybackslash}X}
\toprule
\multicolumn{7}{c}{\textbf{BASELINE ---- ENG$\rightarrow$`Reversed' French}} \\ 
\midrule
\makecell{\textbf{Test Set} \\ \textbf{Partition}} & 
\makecell{\textbf{W} \\ \textbf{(no CoT)}} & 
\makecell{\textbf{W} \\ \textbf{(CoT)}} & 
\makecell{\textbf{W+IB} \\ \textbf{(no CoT)}} & 
\makecell{\textbf{W+IB} \\ \textbf{(CoT)}} & 
\makecell{\textbf{W+G} \\ \textbf{(no CoT)}} & 
\makecell{\textbf{W+G} \\ \textbf{(CoT)}} \\ 
\midrule

1. & \cellcolor[rgb]{0.050,0.030,0.528}\textcolor{white}{0.0} & \cellcolor[rgb]{0.050,0.030,0.528}\textcolor{white}{0.0} & \cellcolor[rgb]{0.149,0.021,0.571}\textcolor{white}{4.0} & \cellcolor[rgb]{0.186,0.019,0.587}\textcolor{white}{6.0} & \cellcolor[rgb]{0.448,0.002,0.660}\textcolor{white}{22.0} & \cellcolor[rgb]{0.896,0.416,0.366}\textcolor{black}{62.0} \\
2m & \cellcolor[rgb]{0.050,0.030,0.528}\textcolor{white}{0.0} & \cellcolor[rgb]{0.050,0.030,0.528}\textcolor{white}{0.0} & \cellcolor[rgb]{0.050,0.030,0.528}\textcolor{white}{0.0} & \cellcolor[rgb]{0.050,0.030,0.528}\textcolor{white}{0.0} & \cellcolor[rgb]{0.124,0.023,0.559}\textcolor{white}{3.0} & \cellcolor[rgb]{0.460,0.004,0.660}\textcolor{white}{23.0} \\
2f & \cellcolor[rgb]{0.050,0.030,0.528}\textcolor{white}{0.0} & \cellcolor[rgb]{0.050,0.030,0.528}\textcolor{white}{0.0} & \cellcolor[rgb]{0.106,0.024,0.551}\textcolor{white}{2.0} & \cellcolor[rgb]{0.050,0.030,0.528}\textcolor{white}{0.0} & \cellcolor[rgb]{0.106,0.024,0.551}\textcolor{white}{2.0} & \cellcolor[rgb]{0.149,0.021,0.571}\textcolor{white}{4.0} \\
3. & \cellcolor[rgb]{0.050,0.030,0.528}\textcolor{white}{0.0} & \cellcolor[rgb]{0.050,0.030,0.528}\textcolor{white}{0.0} & \cellcolor[rgb]{0.255,0.014,0.615}\textcolor{white}{10.0} & \cellcolor[rgb]{0.149,0.021,0.571}\textcolor{white}{4.0} & \cellcolor[rgb]{0.563,0.052,0.642}\textcolor{white}{30.0} & \cellcolor[rgb]{0.319,0.008,0.638}\textcolor{white}{14.0} \\
4. & \cellcolor[rgb]{0.050,0.030,0.528}\textcolor{white}{0.0} & \cellcolor[rgb]{0.050,0.030,0.528}\textcolor{white}{0.0} & \cellcolor[rgb]{0.050,0.030,0.528}\textcolor{white}{0.0} & \cellcolor[rgb]{0.050,0.030,0.528}\textcolor{white}{0.0} & \cellcolor[rgb]{0.050,0.030,0.528}\textcolor{white}{0.0} & \cellcolor[rgb]{0.221,0.016,0.602}\textcolor{white}{8.0} \\
5m. & \cellcolor[rgb]{0.050,0.030,0.528}\textcolor{white}{0.0} & \cellcolor[rgb]{0.050,0.030,0.528}\textcolor{white}{0.0} & \cellcolor[rgb]{0.050,0.030,0.528}\textcolor{white}{0.0} & \cellcolor[rgb]{0.050,0.030,0.528}\textcolor{white}{0.0} & \cellcolor[rgb]{0.050,0.030,0.528}\textcolor{white}{0.0} & \cellcolor[rgb]{0.050,0.030,0.528}\textcolor{white}{0.0} \\
5f. & \cellcolor[rgb]{0.050,0.030,0.528}\textcolor{white}{0.0} & \cellcolor[rgb]{0.050,0.030,0.528}\textcolor{white}{0.0} & \cellcolor[rgb]{0.050,0.030,0.528}\textcolor{white}{0.0} & \cellcolor[rgb]{0.050,0.030,0.528}\textcolor{white}{0.0} & \cellcolor[rgb]{0.050,0.030,0.528}\textcolor{white}{0.0} & \cellcolor[rgb]{0.050,0.030,0.528}\textcolor{white}{0.0} \\
6 & \cellcolor[rgb]{0.050,0.030,0.528}\textcolor{white}{0.0} & \cellcolor[rgb]{0.050,0.030,0.528}\textcolor{white}{0.0} & \cellcolor[rgb]{0.050,0.030,0.528}\textcolor{white}{0.0} & \cellcolor[rgb]{0.050,0.030,0.528}\textcolor{white}{0.0} & \cellcolor[rgb]{0.050,0.030,0.528}\textcolor{white}{0.0} & \cellcolor[rgb]{0.050,0.030,0.528}\textcolor{white}{0.0} \\
7m. & \cellcolor[rgb]{0.050,0.030,0.528}\textcolor{white}{0.0} & \cellcolor[rgb]{0.050,0.030,0.528}\textcolor{white}{0.0} & \cellcolor[rgb]{0.050,0.030,0.528}\textcolor{white}{0.0} & \cellcolor[rgb]{0.050,0.030,0.528}\textcolor{white}{0.0} & \cellcolor[rgb]{0.050,0.030,0.528}\textcolor{white}{0.0} & \cellcolor[rgb]{0.050,0.030,0.528}\textcolor{white}{0.0} \\
7f. & \cellcolor[rgb]{0.050,0.030,0.528}\textcolor{white}{0.0} & \cellcolor[rgb]{0.050,0.030,0.528}\textcolor{white}{0.0} & \cellcolor[rgb]{0.050,0.030,0.528}\textcolor{white}{0.0} & \cellcolor[rgb]{0.050,0.030,0.528}\textcolor{white}{0.0} & \cellcolor[rgb]{0.050,0.030,0.528}\textcolor{white}{0.0} & \cellcolor[rgb]{0.050,0.030,0.528}\textcolor{white}{0.0} \\
\midrule
\textbf{Average} & \cellcolor[rgb]{0.050,0.030,0.528}\textcolor{white}{0.0} & \cellcolor[rgb]{0.050,0.030,0.528}\textcolor{white}{0.0} & \cellcolor[rgb]{0.096,0.025,0.547}\textcolor{white}{1.6} & \cellcolor[rgb]{0.075,0.027,0.538}\textcolor{white}{1.0} & \cellcolor[rgb]{0.179,0.019,0.584}\textcolor{white}{5.7} & \cellcolor[rgb]{0.274,0.012,0.623}\textcolor{white}{11.1} \\

\bottomrule
\end{tabularx}
}
\caption{Baseline accuracy scores for the eng$\rightarrow$art translation direction with `reversed' French as the conlang matrix.}
\label{tab:baseline_eng_art_rf}
\end{table}

\begin{table}[H]
\renewcommand{\arraystretch}{2.5} 
\centering
\large 
\resizebox{\columnwidth}{!}{ 
\begin{tabularx}{\textwidth}{>{\centering\arraybackslash}X >{\centering\arraybackslash}X >{\centering\arraybackslash}X >{\centering\arraybackslash}X >{\centering\arraybackslash}X >{\centering\arraybackslash}X >{\centering\arraybackslash}X}
\toprule
\multicolumn{7}{c}{\textbf{BASELINE ---- `True' French$\rightarrow$ENG}} \\ 
\midrule
\makecell{\textbf{Test Set} \\ \textbf{Partition}} & 
\makecell{\textbf{W} \\ \textbf{(no CoT)}} & 
\makecell{\textbf{W} \\ \textbf{(CoT)}} & 
\makecell{\textbf{W+IB} \\ \textbf{(no CoT)}} & 
\makecell{\textbf{W+IB} \\ \textbf{(CoT)}} & 
\makecell{\textbf{W+G} \\ \textbf{(no CoT)}} & 
\makecell{\textbf{W+G} \\ \textbf{(CoT)}} \\ 
\midrule

1. & \cellcolor[rgb]{0.050,0.030,0.528}\textcolor{white}{0.0} & \cellcolor[rgb]{0.106,0.024,0.551}\textcolor{white}{2.0} & \cellcolor[rgb]{0.670,0.143,0.582}\textcolor{white}{38.0} & \cellcolor[rgb]{0.989,0.810,0.145}\textcolor{black}{90.0} & \cellcolor[rgb]{0.952,0.942,0.153}\textcolor{black}{98.0} & \cellcolor[rgb]{0.940,0.975,0.131}\textcolor{black}{100.0} \\
2m & \cellcolor[rgb]{0.050,0.030,0.528}\textcolor{white}{0.0} & \cellcolor[rgb]{0.186,0.019,0.587}\textcolor{white}{6.0} & \cellcolor[rgb]{0.646,0.121,0.599}\textcolor{white}{36.0} & \cellcolor[rgb]{0.976,0.597,0.245}\textcolor{black}{76.0} & \cellcolor[rgb]{0.964,0.908,0.149}\textcolor{black}{96.0} & \cellcolor[rgb]{0.952,0.942,0.153}\textcolor{black}{98.0} \\
2f & \cellcolor[rgb]{0.050,0.030,0.528}\textcolor{white}{0.0} & \cellcolor[rgb]{0.149,0.021,0.571}\textcolor{white}{4.0} & \cellcolor[rgb]{0.910,0.439,0.350}\textcolor{black}{64.0} & \cellcolor[rgb]{0.994,0.747,0.164}\textcolor{black}{86.0} & \cellcolor[rgb]{0.989,0.810,0.145}\textcolor{black}{90.0} & \cellcolor[rgb]{0.964,0.908,0.149}\textcolor{black}{96.0} \\
3. & \cellcolor[rgb]{0.050,0.030,0.528}\textcolor{white}{0.0} & \cellcolor[rgb]{0.050,0.030,0.528}\textcolor{white}{0.0} & \cellcolor[rgb]{0.670,0.143,0.582}\textcolor{white}{38.0} & \cellcolor[rgb]{0.477,0.007,0.660}\textcolor{white}{24.0} & \cellcolor[rgb]{0.590,0.073,0.630}\textcolor{white}{32.0} & \cellcolor[rgb]{0.590,0.073,0.630}\textcolor{white}{32.0} \\
4. & \cellcolor[rgb]{0.050,0.030,0.528}\textcolor{white}{0.0} & \cellcolor[rgb]{0.050,0.030,0.528}\textcolor{white}{0.0} & \cellcolor[rgb]{0.255,0.014,0.615}\textcolor{white}{10.0} & \cellcolor[rgb]{0.387,0.001,0.654}\textcolor{white}{18.0} & \cellcolor[rgb]{0.506,0.016,0.656}\textcolor{white}{26.0} & \cellcolor[rgb]{0.693,0.165,0.565}\textcolor{white}{40.0} \\
5m. & \cellcolor[rgb]{0.050,0.030,0.528}\textcolor{white}{0.0} & \cellcolor[rgb]{0.050,0.030,0.528}\textcolor{white}{0.0} & \cellcolor[rgb]{0.149,0.021,0.571}\textcolor{white}{4.0} & \cellcolor[rgb]{0.186,0.019,0.587}\textcolor{white}{6.0} & \cellcolor[rgb]{0.563,0.052,0.642}\textcolor{white}{30.0} & \cellcolor[rgb]{0.693,0.165,0.565}\textcolor{white}{40.0} \\
5f. & \cellcolor[rgb]{0.050,0.030,0.528}\textcolor{white}{0.0} & \cellcolor[rgb]{0.050,0.030,0.528}\textcolor{white}{0.0} & \cellcolor[rgb]{0.186,0.019,0.587}\textcolor{white}{6.0} & \cellcolor[rgb]{0.387,0.001,0.654}\textcolor{white}{18.0} & \cellcolor[rgb]{0.477,0.007,0.660}\textcolor{white}{24.0} & \cellcolor[rgb]{0.563,0.052,0.642}\textcolor{white}{30.0} \\
6 & \cellcolor[rgb]{0.050,0.030,0.528}\textcolor{white}{0.0} & \cellcolor[rgb]{0.050,0.030,0.528}\textcolor{white}{0.0} & \cellcolor[rgb]{0.149,0.021,0.571}\textcolor{white}{4.0} & \cellcolor[rgb]{0.050,0.030,0.528}\textcolor{white}{0.0} & \cellcolor[rgb]{0.798,0.280,0.470}\textcolor{white}{50.0} & \cellcolor[rgb]{0.186,0.019,0.587}\textcolor{white}{6.0} \\
7m. & \cellcolor[rgb]{0.050,0.030,0.528}\textcolor{white}{0.0} & \cellcolor[rgb]{0.050,0.030,0.528}\textcolor{white}{0.0} & \cellcolor[rgb]{0.149,0.021,0.571}\textcolor{white}{4.0} & \cellcolor[rgb]{0.149,0.021,0.571}\textcolor{white}{4.0} & \cellcolor[rgb]{0.535,0.031,0.650}\textcolor{white}{28.0} & \cellcolor[rgb]{0.221,0.016,0.602}\textcolor{white}{8.0} \\
7f. & \cellcolor[rgb]{0.050,0.030,0.528}\textcolor{white}{0.0} & \cellcolor[rgb]{0.050,0.030,0.528}\textcolor{white}{0.0} & \cellcolor[rgb]{0.149,0.021,0.571}\textcolor{white}{4.0} & \cellcolor[rgb]{0.149,0.021,0.571}\textcolor{white}{4.0} & \cellcolor[rgb]{0.477,0.007,0.660}\textcolor{white}{24.0} & \cellcolor[rgb]{0.255,0.014,0.615}\textcolor{white}{10.0} \\
\midrule
\textbf{Average} & \cellcolor[rgb]{0.050,0.030,0.528}\textcolor{white}{0.0} & \cellcolor[rgb]{0.086,0.026,0.543}\textcolor{white}{1.2} & \cellcolor[rgb]{0.430,0.001,0.659}\textcolor{white}{20.8} & \cellcolor[rgb]{0.600,0.082,0.625}\textcolor{white}{32.6} & \cellcolor[rgb]{0.795,0.276,0.473}\textcolor{white}{49.8} & \cellcolor[rgb]{0.756,0.232,0.509}\textcolor{white}{46.0} \\

\bottomrule
\end{tabularx}
}
\caption{Baseline accuracy scores for the art$\rightarrow$eng translation direction with `true' French as the conlang matrix.}
\label{tab:baseline_art_eng_tf}
\end{table}

\begin{table}[H]
\renewcommand{\arraystretch}{2.5} 
\centering
\large 
\resizebox{\columnwidth}{!}{ 
\begin{tabularx}{\textwidth}{>{\centering\arraybackslash}X >{\centering\arraybackslash}X >{\centering\arraybackslash}X >{\centering\arraybackslash}X >{\centering\arraybackslash}X >{\centering\arraybackslash}X >{\centering\arraybackslash}X}
\toprule
\multicolumn{7}{c}{\textbf{BASELINE ---- `Reversed' French$\rightarrow$ENG}} \\ 
\midrule
\makecell{\textbf{Test Set} \\ \textbf{Partition}} & 
\makecell{\textbf{W} \\ \textbf{(no CoT)}} & 
\makecell{\textbf{W} \\ \textbf{(CoT)}} & 
\makecell{\textbf{W+IB} \\ \textbf{(no CoT)}} & 
\makecell{\textbf{W+IB} \\ \textbf{(CoT)}} & 
\makecell{\textbf{W+G} \\ \textbf{(no CoT)}} & 
\makecell{\textbf{W+G} \\ \textbf{(CoT)}} \\ 
\midrule

1. & \cellcolor[rgb]{0.050,0.030,0.528}\textcolor{white}{0.0} & \cellcolor[rgb]{0.050,0.030,0.528}\textcolor{white}{0.0} & \cellcolor[rgb]{0.715,0.187,0.546}\textcolor{white}{42.0} & \cellcolor[rgb]{0.756,0.232,0.509}\textcolor{white}{46.0} & \cellcolor[rgb]{0.964,0.908,0.149}\textcolor{black}{96.0} & \cellcolor[rgb]{0.940,0.975,0.131}\textcolor{black}{100.0} \\
2m & \cellcolor[rgb]{0.050,0.030,0.528}\textcolor{white}{0.0} & \cellcolor[rgb]{0.050,0.030,0.528}\textcolor{white}{0.0} & \cellcolor[rgb]{0.186,0.019,0.587}\textcolor{white}{6.0} & \cellcolor[rgb]{0.715,0.187,0.546}\textcolor{white}{42.0} & \cellcolor[rgb]{0.976,0.597,0.245}\textcolor{black}{76.0} & \cellcolor[rgb]{0.974,0.875,0.144}\textcolor{black}{94.0} \\
2f & \cellcolor[rgb]{0.050,0.030,0.528}\textcolor{white}{0.0} & \cellcolor[rgb]{0.106,0.024,0.551}\textcolor{white}{2.0} & \cellcolor[rgb]{0.590,0.073,0.630}\textcolor{white}{32.0} & \cellcolor[rgb]{0.866,0.370,0.400}\textcolor{black}{58.0} & \cellcolor[rgb]{0.992,0.681,0.195}\textcolor{black}{82.0} & \cellcolor[rgb]{0.983,0.842,0.142}\textcolor{black}{92.0} \\
3. & \cellcolor[rgb]{0.050,0.030,0.528}\textcolor{white}{0.0} & \cellcolor[rgb]{0.050,0.030,0.528}\textcolor{white}{0.0} & \cellcolor[rgb]{0.621,0.099,0.614}\textcolor{white}{34.0} & \cellcolor[rgb]{0.448,0.002,0.660}\textcolor{white}{22.0} & \cellcolor[rgb]{0.646,0.121,0.599}\textcolor{white}{36.0} & \cellcolor[rgb]{0.621,0.099,0.614}\textcolor{white}{34.0} \\
4. & \cellcolor[rgb]{0.050,0.030,0.528}\textcolor{white}{0.0} & \cellcolor[rgb]{0.050,0.030,0.528}\textcolor{white}{0.0} & \cellcolor[rgb]{0.050,0.030,0.528}\textcolor{white}{0.0} & \cellcolor[rgb]{0.149,0.021,0.571}\textcolor{white}{4.0} & \cellcolor[rgb]{0.050,0.030,0.528}\textcolor{white}{0.0} & \cellcolor[rgb]{0.319,0.008,0.638}\textcolor{white}{14.0} \\
5m. & \cellcolor[rgb]{0.050,0.030,0.528}\textcolor{white}{0.0} & \cellcolor[rgb]{0.050,0.030,0.528}\textcolor{white}{0.0} & \cellcolor[rgb]{0.050,0.030,0.528}\textcolor{white}{0.0} & \cellcolor[rgb]{0.221,0.016,0.602}\textcolor{white}{8.0} & \cellcolor[rgb]{0.050,0.030,0.528}\textcolor{white}{0.0} & \cellcolor[rgb]{0.221,0.016,0.602}\textcolor{white}{8.0} \\
5f. & \cellcolor[rgb]{0.050,0.030,0.528}\textcolor{white}{0.0} & \cellcolor[rgb]{0.050,0.030,0.528}\textcolor{white}{0.0} & \cellcolor[rgb]{0.149,0.021,0.571}\textcolor{white}{4.0} & \cellcolor[rgb]{0.186,0.019,0.587}\textcolor{white}{6.0} & \cellcolor[rgb]{0.050,0.030,0.528}\textcolor{white}{0.0} & \cellcolor[rgb]{0.106,0.024,0.551}\textcolor{white}{2.0} \\
6 & \cellcolor[rgb]{0.050,0.030,0.528}\textcolor{white}{0.0} & \cellcolor[rgb]{0.050,0.030,0.528}\textcolor{white}{0.0} & \cellcolor[rgb]{0.050,0.030,0.528}\textcolor{white}{0.0} & \cellcolor[rgb]{0.050,0.030,0.528}\textcolor{white}{0.0} & \cellcolor[rgb]{0.106,0.024,0.551}\textcolor{white}{2.0} & \cellcolor[rgb]{0.106,0.024,0.551}\textcolor{white}{2.0} \\
7m. & \cellcolor[rgb]{0.050,0.030,0.528}\textcolor{white}{0.0} & \cellcolor[rgb]{0.050,0.030,0.528}\textcolor{white}{0.0} & \cellcolor[rgb]{0.050,0.030,0.528}\textcolor{white}{0.0} & \cellcolor[rgb]{0.050,0.030,0.528}\textcolor{white}{0.0} & \cellcolor[rgb]{0.050,0.030,0.528}\textcolor{white}{0.0} & \cellcolor[rgb]{0.050,0.030,0.528}\textcolor{white}{0.0} \\
7f. & \cellcolor[rgb]{0.050,0.030,0.528}\textcolor{white}{0.0} & \cellcolor[rgb]{0.050,0.030,0.528}\textcolor{white}{0.0} & \cellcolor[rgb]{0.050,0.030,0.528}\textcolor{white}{0.0} & \cellcolor[rgb]{0.050,0.030,0.528}\textcolor{white}{0.0} & \cellcolor[rgb]{0.050,0.030,0.528}\textcolor{white}{0.0} & \cellcolor[rgb]{0.050,0.030,0.528}\textcolor{white}{0.0} \\
\midrule
\textbf{Average} & \cellcolor[rgb]{0.050,0.030,0.528}\textcolor{white}{0.0} & \cellcolor[rgb]{0.050,0.030,0.528}\textcolor{white}{0.2} & \cellcolor[rgb]{0.287,0.011,0.627}\textcolor{white}{11.8} & \cellcolor[rgb]{0.393,0.001,0.655}\textcolor{white}{18.6} & \cellcolor[rgb]{0.552,0.043,0.645}\textcolor{white}{29.2} & \cellcolor[rgb]{0.626,0.103,0.611}\textcolor{white}{34.6} \\

\bottomrule
\end{tabularx}
}
\caption{Baseline accuracy scores for the art$\rightarrow$eng translation direction with `reversed' French as the conlang matrix.}
\label{tab:baseline_art_eng_rf}
\end{table}